\documentclass[journal,10pt]{IEEEtran}

\usepackage{amssymb}
\usepackage{graphicx,psfrag} % for pdf, jpeg, bitmapped graphics files
\usepackage{epsfig} % for postscript and .eps graphics files
\usepackage{esdiff} %package for derivative 
\usepackage{amsmath,amssymb} % assumes amsmath package installed
\usepackage{wasysym}
\usepackage{xfrac}   % for writting fraction in better way. Not essential
\usepackage{subfigure}  % for making a row or column of subfigures
\usepackage{pst-sigsys} % for makinf signal processing block diagrams
\usepackage{algorithm, algorithmic}
\usepackage{cite}
\usepackage{mathtools}
\usepackage{url}
\usepackage{bbm}
\usepackage{xr}
\usepackage{setspace}
\usepackage{color}
\usepackage{etoolbox}
\makeatletter
\patchcmd{\@makecaption}
  {\scshape}
  {}
  {}
  {}
\makeatother

\numberwithin{thm}{section}

\numberwithin{lem}{section}
\newtheorem{propos}{Proposition}

\numberwithin{defn}{section}

\ifCLASSINFOpdf
\else
\fi
\begin{document}
\title{Quantization-Aware Phase Retrieval}% from Quantized Measurements Using a Consistency Criterion}
\author{Subhadip~Mukherjee
        and~Chandra~Sekhar~Seelamantula, {\it Senior Member, IEEE}% <-this % stops a space
\thanks{\scriptsize The authors are with the Department
of Electrical Engineering, Indian Institute of Science, Bangalore--560012, India. Phone: +91 80 22932695; Fax: +91 80 23600444; Emails: subhadipm@iisc.ac.in, chandra.sekhar@ieee.org. 
}
}

\markboth{}
{Shell \MakeLowercase{\textit{et al.}}: Bare Demo of IEEEtran.cls for Journals}
\maketitle

\begin{abstract}
We address the problem of phase retrieval (PR) from quantized measurements. The goal is to reconstruct a signal from quadratic measurements encoded with a finite precision, which is indeed the case in many practical applications. We develop a rank-1 projection algorithm that recovers the signal subject to ensuring consistency with the measurement, that is, the recovered signal when encoded must yield the same set of measurements that one started with. The rank-1 projection stems from the idea of \textbf{\textit{lifting}}, originally proposed in the context of \textbf{\textit{PhaseLift}}. The consistency criterion is enforced using a one-sided quadratic cost. We also determine the probability with which different vectors lead to the same set of quantized measurements, which makes it impossible to  resolve them. Naturally, this probability depends on how correlated such vectors are, and how coarsely/finely the measurements get quantized. The proposed algorithm is also capable of incorporating a sparsity constraint on the signal. An analysis of the cost function reveals that it is bounded, both above and below, by functions that are dependent on how well correlated the estimate is with the ground truth. We also derive the Cram\'er-Rao lower bound (CRB) on the achievable reconstruction accuracy. A comparison with the state-of-the-art algorithms shows that the proposed algorithm has a higher reconstruction accuracy and is about 2 to 3 dB away from the CRB. The edge, in terms of the reconstruction signal-to-noise ratio, over the competing algorithms is higher (about 5 to 6 dB) when the quantization is coarse. 
\end{abstract}
\section{Introduction}
\IEEEPARstart{T}{he} objective of phase retrieval (PR) is to estimate a signal $\boldsymbol x^*\in \mathbb{R}^n$ from intensity measurements given as
\begin{equation}
b_i = \left|\boldsymbol a_i^\top \boldsymbol x^*\right|^2, i=1:m,
\label{pr_int_model}
\end{equation}
where $\left\{\boldsymbol a_i \right\}_{i=1}^{m}\in \mathbb{R}^n \,\,(\text{or\,\,}\mathbb{C}^n)$ are \textit{known} sampling vectors. For a complex vector $\boldsymbol a$, the notation $\boldsymbol a^\top$ denotes its Hermitian transpose. In addition, the measurements may get corrupted by noise. The measurement model considered in \eqref{pr_int_model} arises in a number of imaging applications, such as X-ray crystallography \cite{crystallography_chspr}, holography \cite{holography_chspr}, electron microscopy \cite{microscopy_chspr}, etc. For example, the diffraction patterns of objects to be imaged using X-ray crystallography closely approximate their Fourier transforms. The sensors can only record the intensities of the complex wave-field; and the phase, which contains critical structural information about the object, is not directly measured. Thus, it becomes imperative to recover the phase, starting from the Fourier magnitude/intensity measurements, in order to reconstruct the object accurately. The fundamental objective of PR is to solve this otherwise ill-posed inverse problem by taking into account prior information about the underlying signal, such as non-negativity, compact support, sparsity, etc. One could also resolve the phase ambiguity by considering oversampled measurements exceeding the signal dimension ($m>n$). In the special case where the vectors $\boldsymbol a_i$ correspond to the discrete Fourier transform (DFT) basis vectors, the PR problem reduces to classical Fourier PR, wherein one seeks to reconstruct a signal starting from its Fourier magnitude or intensity. The generalized setting, which involves projections onto random sampling vectors is the one that we shall consider. Before giving a formal statement of the problem considered in this paper, we provide a concise review of the existing PR literature to put our contributions into perspective.       
\subsection{A Survey of Phase Retrieval Literature}
The PR problem has its origin in optics and astronomy. The initial contributions were due to Fienup \cite{fienup_main1,fienup_main2}, and Gerchberg and Saxton \cite{gs_pr}, who proposed \textit{iterative error-reduction algorithms} that bounce estimates back and forth between the object and the measurement domains, and incorporate respective priors. The most widely used priors in the signal domain are causality, non-negativity, compact support, sparsity, etc. Fienup's algorithm has been the most popular technique for PR in the optics community, and works reasonably well for a wide class of imaging problems. A comprehensive overview of the Fienup algorithm and several of its variants can be found in \cite{bauschke_pr} and the references therein. A notable variant of the Fienup algorithm was developed by Quatieri et al. \cite{quatieri_pr} for reconstructing minimum-phase signals from their DFT magnitude measurements, wherein one iteratively imposes the causality constraint in the signal domain, and combines the measured magnitude spectrum with the current estimate of phase in the frequency domain. Apart from the iterative algorithms, there exist non-iterative techniques \cite{yegna_pr}, which rely on the Hilbert transform relationship between the log-magnitude and the phase of the Fourier transform of minimum-phase signals, in order to reconstruct them from magnitude-/phase-only measurements. The two-dimensional (2-D) counterpart of such results and exact reconstruction guarantees were proposed in \cite{css_holography} in the context of digital holography. We recently developed a non-iterative algorithm \cite{param2D_icip,param2D_tip} to solve the PR problem for a class of 2-D parametric models, by extending the concept of minimum-phase signals in 1-D. An exact PR methodology for signals lying in shift-invariant spaces was developed in \cite{pr_shift_invar}. We also recently constructed generalized minimum-phase signals and developed corresponding 2-D Hilbert integral equations \cite{css_2d_hilbert}.\\ 
\indent Moravec et al. addressed the problem of PR within the realm of sparsity and magnitude-only compressive measurements \cite{moravec_cpr}. The \textit{compressive PR} problem received considerable attention because of its wide applicability. Signals encountered in a number of applications indeed admit a sparse representation in an appropriately chosen basis. Yu and Vetterli proposed a sparse spectral factorization technique \cite{yu_vettreli_ssf}, and established uniqueness guarantees, where the objective was to recover a sparse signal from its autocorrelation sequence. A greedy local search-based algorithm for sparse PR, referred to as GESPAR, was proposed by Schechtman et al. \cite{gespar}. The scalability and accuracy of GESPAR has been established in \cite{gespar} by extensive simulations in a variety of practical settings. Netrapalli et al. \cite{netrapalli_altminpr} developed an analytical convergence guarantee for the well-known alternating minimization (Alt. Min.) framework for PR, with and without the constraint of sparsity, referred to as \textit{AltMinPhase} and \textit{SparseAltMinPhase}, respectively. Their work is the first one in the literature to establish the correctness of Alt. Min. for PR, subject to the so-called \textit{spectral initialization} \cite{netrapalli_altminpr}. Vaswani et al. \cite{low_rank_PR_vaswani} recently proposed an Alt. Min. technique for low-rank PR, where the task is to recover a low-rank matrix from the quadratic measurements corresponding to projections with each of its columns. Other notable contributions for sparsity regularized PR include dictionary learning for PR (DOLPHIn) \cite{dolphin}, cone programming \cite{tfocs}, compressive PR via generalized message passing \cite{schniter_gamp}, simulated annealing for sparse Boolean signals \cite{bspr_peng}, majorization-minimization for recovery from under-sampled measurements \cite{palomar_mm}, etc. Fogel et al. \cite{fogel_pr_imaging} have recently shown that incorporating signal priors, such as sparsity and positivity lead to a significant speed-up of iterative reconstruction techniques.\\ 
\indent A distinctive contribution in PR is the \textit{PhaseLift} framework pioneered by Cand\`es et al. \cite{PL1_candes,PL2_candes}. PhaseLift relies on the idea of \textit{lifting} a vector $\boldsymbol{x}^*$ to a matrix $\boldsymbol{X}^*=\boldsymbol{x}^* \boldsymbol{x}^{*\top}$ such that the quadratic measurements of $\boldsymbol{x}^*$ get converted to an equivalent set of linear measurements of $\boldsymbol{X}^*$. Reconstruction is achieved by solving a tractable semi-definite program (SDP). One encounters two scenarios within this framework: (i) the absence of a signal prior, together with a set of oversampled measurements; or (ii) reconstruction subject to a signal prior such as sparsity. Ohlsson et al.'s \textit{compressive PR with lifting} (CPRL) \cite{cpr_pl} technique falls in the second category, where sparsity is enforced via an $\ell_1$ penalty. Schechtman et al. \cite{qcs_eldar} also developed a sparse PR technique in the context of sub-wavelength imaging with partially incoherent light by employing the idea of PhaseLift, wherein the sparsity constraint on the underlying image is imposed via a \textit{log-det} penalty. The PhaseLift technique requires spectral decomposition of an $n\times n$ matrix within every iteration, where $n$ is the dimension of the signal to be recovered, and, therefore, its complexity per iteration scales as $\mathcal{O}\left(n^3\right)$. However, since one needs to compute only the top few eigenvectors of a symmetric matrix, one could employ \textit{power iterations} \cite[Chapter 7]{power_it} to significantly speed-up the computation. Gradient-descent approaches for PR without lifting include the \textit{Wirtinger flow} (WF) method \cite{candes_wf_it} and its \textit{truncated} version (TWF) \cite{candes_twf}. These algorithms lead to stable reconstruction, and have convergence guarantees provided that the starting point is accurate, which is typically achieved by using spectral initialization. The WF and TWF algorithms are also scalable with respect to the signal dimension. Waldspurger et al. developed the \textit{PhaseCut} technique \cite{phasecut}, where the PR problem is formulated as a non-convex quadratic program and solved using a provable block-coordinate-descent approach. Although the number of variables in the resulting SDP in \textit{PhaseCut} is larger than that in PhaseLift, it has been shown in \cite{phasecut} that the proposed algorithm for PhaseCut has a per-iteration complexity comparable with that of the iterative error-reduction type algorithms.\\
\indent The issue of noise robustness of several PR algorithms has been considered, but the effect of quantization, which is ubiquitous in practical acquisition systems, has not been addressed in the PR literature. In contrast, a significant amount of research has gone into developing algorithms for \textit{compressive sensing} (CS) reconstruction with quantized measurements starting from the work of Zymnis et al. \cite{zymnis_qcs}, who developed two reconstruction algorithms for quantized CS based on $\ell_1$-regularized maximum likelihood (ML) and least-squares estimation. Laska et al. \cite{laska1} studied the effect of quantization and saturation on linear compressive measurements and a bit-precision analysis for CS was carried out by Ardestanizadeh et al. \cite{qcs_bit_precision}. The extreme case of binary/one-bit quantization, wherein one retains only the sign of the measurements, has been investigated in detail in the context of CS \cite{binary_imaging1,binary_imaging2,baraniuk1,plan1,msp}. However, the issue of measurement quantization and its ramifications on the reconstruction algorithms have not been considered in the PR literature. Our attempt in this paper is to fill this void.
\subsection{Our Contributions}
We introduce the problem of QPR, that is phase retrieval from quantized measurements (Section~\ref{sec_problem_formulation_qpr}) and address issues related to distinguishability (Section~\ref{effect_of_quant_sec}) of two distinct signals in terms of their quantized quadratic measurements. Issues related to noise robustness and quantizer design are also part of this development. Subsequently, the optimization framework for QPR is developed by combining the principles of \textit{consistent recovery} and \textit{lifting}, and we propose two gradient-based iterative projection algorithms for signal  reconstruction (Section \ref{formal_opt_lift_const_qpr_sec}). Our algorithms are amenable to incorporating the sparsity prior as well, which boosts the reconstruction signal-to-noise ratio (SNR). The \textit{descent property} of the projected gradient approach for QPR is established in Section~\ref{pgd_descent_proof_qpr} of the supplementary material. An analysis of the cost function shows that it can be bounded both above and below, in a probabilistic setting, as a function of the distance between the ground-truth and the estimate, and parameters that measure the precision of the quantizer (Section~\ref{analysiscost}, supplementary material). Since no explicit quantization-aware PR algorithms exist in the literature, we consider the quantization noise to be additive as far as the implementation of the state-of-the-art PR algorithms (such as \textit{PhaseLift}, \textit{truncated Wirtinger flow} (TWF), \textit{AltMinPhase}, \textit{GESPAR}, \textit{compressive PR with lifting} (CPRL)) are considered, and make performance comparisons for the QPR problem first without a sparsity prior (Section~\ref{QPR_results_sec_dense}) and subsequently with the sparsity prior incorporated (Section~\ref{SQPR_results_sec}). For benchmarking the performance, we derive the Cram\'er-Rao bounds (CRB) assuming white Gaussian noise contamination prior to quantization (Section~\ref{crb_derivation_kbit}, supplementary material). The comparisons show that the proposed algorithm achieves a reconstruction mean-squared error (MSE) that is within $2$--$3$ dB of the CRB and about $5$ dB better than the best performing technique from the state-of-the-art (Section \ref{np_crb_sec_qpr_ch}). Our recent work on PR from binary measurements \cite{css_binaryPR} follows as a special case of the formalism developed in this paper. 
\section{Measurement Model for QPR}
\label{sec_problem_formulation_qpr}
Recall that the objective of PR is to reconstruct a signal $\boldsymbol x^* \in \mathbb{R}^n$ from $m$ intensity measurements of the form \eqref{pr_int_model}, which encompasses the special case of signal reconstruction from DFT intensity measurements. In this paper, we consider the scenario where the intensity measurements $b_i$ are acquired with a finite precision, that is, they are quantized using a finite codebook $\mathcal{S}=\left\{s_1,s_2,\cdots,s_k \right\}$, containing $k$ distinct symbols. The acquired measurements take the form 
\begin{equation}
y_i=\mathcal{Q}\left(\left| \boldsymbol a_i^\top \boldsymbol x^*  \right|^2\right), \text{\,\,}i = 1 : m,
\label{QPR_measurement_eq}
\end{equation}
where the encoding map $\mathcal{Q}:\mathbb{R}^+ \rightarrow \mathcal{S}$ is determined using a set of thresholds $0 = \tau_0 < \tau_1 <\cdots < \tau_k = +\infty$, such that $\mathcal{Q}(u)=s_j$, whenever $\tau_{j-1}\leq u < \tau_{j}$, for $j=1:k$. A codebook of size $k$ requires $\left \lceil\log_2 k \right \rceil$ bits for quantization. The measurement model \eqref{QPR_measurement_eq} is more practical than \eqref{pr_int_model} as any real-world measurement device would have a finite bit precision. In principle, for the consistent recovery framework pursued in this work, the encoding symbols could be arbitrary as long as the association between $[\tau_{j-1}, \tau_j)$ and $s_j$ is known. However, for the sake of simplicity, and for making comparisons with the state-of-the-art techniques that use the encoded measurement values, we assume that 
\begin{equation}
\tau_{j-1} < s_j < \tau_j,\text{\,}j=1:k,
\label{encoding}
\end{equation}
meaning that the encoding symbol for an interval is chosen to be a point falling inside that interval. Since the state-of-the-art PR algorithms aim to minimize an appropriate loss function in the measurement domain, selecting the encoding symbols $\left\{s_j\right\}_{j=1}^{k}$ from the corresponding intervals $(\tau_{j-1},\tau_{j})$ is a reasonable choice. As we shall see, the encoding symbols $s_j$ have no bearing on the reconstruction performance of the proposed approach. However, it would affect the performance of the state-of-the-art techniques.  
\subsection{The Principle of Consistent Recovery}
\indent Since quantization is a non-invertible mapping, it would not be possible to determine $\boldsymbol x^*$ exactly. In other words, there could be two candidate signals $\hat{\boldsymbol x}_1$ and $\hat{\boldsymbol x}_2$ such that their quantized intensity measurements match exactly and one needs to design an optimization objective that does not distinguish between them. If one uses the conventional squared-error loss $J_{\text{sq}}(\hat{\boldsymbol x})=\sum_{i=1}^{m}\left(y_i-\left| \boldsymbol a_i^\top \hat{\boldsymbol x}  \right|^2\right)^2$, wherein the intensity measurements of $\hat{\boldsymbol x}_1$ and $\hat{\boldsymbol x}_2$ are compared with the acquired measurement $\boldsymbol y$, one might get different values of the error corresponding to $\hat{\boldsymbol x}_1$ and $\hat{\boldsymbol x}_2$, although the measurement process does not distinguish between them. Therefore, instead of minimizing the traditional squared-error loss, we seek a solution $\hat{\boldsymbol x}$ that is {\it consistent} with the measurements. The idea of consistency has earlier been used in the context of reconstruction from quantized CS samples \cite{baraniuk1}. To elucidate the idea of \textit{consistent recovery} in the PR context, let 
\begin{equation}
\mathcal{H}_j = \left\{\boldsymbol a_i, 1\leq i \leq m \text{\,}\rvert \text{\,} y_i=s_j \right\}, j=1:k,
\label{separating_hyperplanes}
\end{equation}     
denote the collection of sampling signals such that the corresponding measurements are encoded as $s_j$. A solution $\hat{\boldsymbol x}$ is said to be \textit{consistent} with the measurements if
\begin{equation}
\tau_{j-1}\leq  \left| \boldsymbol a_i^\top \hat{\boldsymbol x}\right|^2< \tau_j, \text{\,\,whenever\,\,} \boldsymbol a_i \in \mathcal{H}_j.
\label{consistency}
\end{equation}  
Essentially, the consistency condition in \eqref{consistency} ensures that the reconstructed vector, when passed through the same acquisition process, matches the measurements that one started with. One can impose the constraint of sparsity depending on the application. Accordingly, we have two problem statements as specified below.
\begin{enumerate}
\item Quantized PR (QPR): Find $\boldsymbol x$ such that $\tau_{j-1}\leq \left| \boldsymbol a_i^\top \boldsymbol x  \right|^2<\tau_j, \text{\,\,whenever\,\,} \boldsymbol a_i \in \mathcal{H}_j.$
\item Sparse QPR (SQPR): Find $\boldsymbol x$ satisfying $\left\|\boldsymbol x\right\|_0\leq s$ such that $\tau_{j-1}\leq \left| \boldsymbol a_i^\top \boldsymbol x  \right|^2<\tau_j, \text{\,\,whenever\,\,} \boldsymbol a_i \in \mathcal{H}_j.$
%\item Binary PR (BPR): Find $\boldsymbol x$ such that $y_i\left(\left| \boldsymbol a_i^\top \boldsymbol x  \right|^2-\tau\right)>0$, for $i=1:m$.
%\item Sparse BPR (SBPR): Find $\boldsymbol x$ with $\left\|\boldsymbol x\right\|_0\leq s$ such that $y_i\left(\left| \boldsymbol a_i^\top \boldsymbol x  \right|^2-\tau\right)>0$, for $i=1:m$.
\end{enumerate}
\subsection{Performance Metrics}
\indent If $\hat{\boldsymbol x}\in \mathbb{R}^n$ is a consistent solution to any of the PR problems posed above, so is $-\hat{\boldsymbol x}$. In order to factor out the effect of the global sign, an appropriate metric to quantify the accuracy of reconstruction {\it vis-\`a-vis} the ground truth $\boldsymbol x^*$ would be the global-sign-invariant reconstruction SNR \cite{PL2_candes} defined as\footnote{For a complex ground-truth signal $\boldsymbol x^*$, the global-phase-invariant reconstruction SNR can be defined as $\underset{0\leq \phi \leq 2\pi}{\max}\frac{\left\| \boldsymbol x^* \right\|_2^2}{\left\| e^{\mathrm{j}\phi}\hat{\boldsymbol x}-\boldsymbol x^* \right\|_2^2}$.}:
\begin{equation}
\text{Reconstruction SNR}=\underset{\alpha\in\{-1,+1\}}{\max}\frac{\left\| \boldsymbol x^* \right\|_2^2}{\left\| \alpha\hat{\boldsymbol x}-\boldsymbol x^* \right\|_2^2}.
\label{mse_def2}
\end{equation}
The MSE of reconstruction is defined as the reciprocal of the SNR metric. The second metric that would be relevant in the context of QPR is the \textit{consistency} (denoted as $\Upsilon$) of the reconstruction $\hat{\boldsymbol x}$ with the measurements, defined as
\begin{equation}
\Upsilon=\frac{1}{m}\sum_{i=1}^{m}\mathbbm{1}\left\{\mathcal{Q}\left( \left| \boldsymbol a_i^\top \boldsymbol x^*\right|^2 \right)=\mathcal{Q}\left( \left| \boldsymbol a_i^\top \hat{\boldsymbol x}\right|^2 \right) \right\},
\label{consistency_def_eq}
\end{equation}
where $\mathbbm{1}\left(\mathcal{E}\right)$ is the indicator of the event $\mathcal{E}$. Essentially, the consistency metric $\Upsilon$ quantifies the fraction of measurements correctly explained by $\hat{\boldsymbol x}$. Naturally, $0 \leq \Upsilon \leq 1$, and $\Upsilon = 1$ is the best that one could hope to achieve.
\section{Effect of Quantization on the Measurements}
\label{effect_of_quant_sec}
We next address the issues of distinguishability, noise-robustness of the quantized measurements, and the design of the quantizer. Akin to \cite{candes_twf}, one could consider two sampling models: (i) the real model, where $\boldsymbol a_i\sim\mathcal{N}\left(\boldsymbol 0,\boldsymbol I_n\right)$; and (ii) the complex model, where the test vectors $\boldsymbol a_i$ can be expressed as $\boldsymbol a_i=\boldsymbol{a}_{i_{\text{re}}}+\mathrm{j}\boldsymbol{a}_{i_{\text{im}}}$, $\mathrm{j}=\sqrt{-1}$, with $\boldsymbol{a}_{i_{\text{re}}}$ and $\boldsymbol{a}_{i_{\text{im}}}$ drawn independently drawn from $\mathcal{N}\left(\boldsymbol 0,\boldsymbol I_n\right)$. The subscripts re and im denote the real and imaginary parts, respectively, whereas $\boldsymbol I_n$ denotes the $n\times n$ identity matrix. The signal $\boldsymbol x^*$ is assumed to have unit norm in both models, without loss of generality. We consider the real signal model throughout our development and the analysis carried out in this section is only applicable to the real model. The notation $\mathbb{U}^n$ denotes the unit sphere in $\mathbb{R}^n$. In the following proposition, we derive the distribution of the full-precision quadratic measurement.  
\begin{propos}
The quadratic measurement $b=\left| \boldsymbol a^\top \boldsymbol x^* \right|^2$ is $\chi^2$ distributed with one degree of freedom for any $\boldsymbol x^* \in \mathbb{U}^n$ and $\boldsymbol a\sim\mathcal{N}\left(\boldsymbol 0,\boldsymbol I_n\right)$, that is $b \sim \chi^2_1$. 
\label{meas_dist_prop}
\end{propos}
\noindent\textit{Proof}: Observe that 
\begin{equation*}
\left| \boldsymbol a^\top \boldsymbol x^* \right|^2=\boldsymbol a^\top \boldsymbol x^* \boldsymbol x^{*\top} \boldsymbol a=\boldsymbol a^\top \boldsymbol U \boldsymbol \Lambda \boldsymbol U^\top \boldsymbol a=\tilde{a}_1^2, 
\end{equation*}
where $\boldsymbol \Lambda=\text{diag}\left(\underbrace{\boldsymbol x^{*\top} \boldsymbol x^*}_{=1},0,\cdots,0\right)$, and $\tilde{\boldsymbol a}=\boldsymbol U^\top\boldsymbol a$ follows the same distribution as $\boldsymbol a$, since $\boldsymbol U$ is orthonormal. The proposition is now a direct consequence of the fact that $\tilde{a}_1^2$ is the square of a $\mathcal{N}\left(0,1\right)$ random variable. \hfill $\blacksquare$\\
The cumulative distribution function (c.d.f.) of $b=\left| \boldsymbol a^\top \boldsymbol x^* \right|^2$ is given by 
\begin{equation}
\mathcal{F}_b\left(b\right)=\frac{\gamma\left(\frac{b}{2},\frac{1}{2}\right)}{\sqrt{\pi}},
\label{def_measurement_cdf}
\end{equation}
where $\gamma\left(w,\alpha\right)=\displaystyle\int_{0}^{w}t^{\alpha -1}e^{-t}\mathrm{d}t$ denotes the lower incomplete gamma function.
%%%%%%%%%%%%%%%%%%%%%%%%%%%%%%%%%%%%%%%%%%%%%
\subsection{Distinguishability of the Quantized Measurements}
\label{Discriminability_sec}
Let $\boldsymbol x_1$ and $\boldsymbol x_2$ be two linearly independent signals in $\mathbb{U}^n$, meaning that $\boldsymbol x_1 \neq \pm \boldsymbol x_2$. We analyze the probability $P_e$ of $\boldsymbol x_1$ and $\boldsymbol x_2$ being mapped to the same set of quantized measurements, which renders them indiscernible. For any reconstruction algorithm to succeed, it is necessary to collect a sufficient number of measurements to keep $P_e$ low. To begin with, we place an upper-bound on the error probability $P_e$ in the case where only one measurement is acquired. Let $b_1=\left|\boldsymbol a^\top \boldsymbol x_1\right|^2$ and $b_2=\left|\boldsymbol a^\top \boldsymbol x_2\right|^2$. It would be impossible for any algorithm to differentiate $\boldsymbol x_1$ from $\boldsymbol x_2$ from their quantized measurements if 
\begin{equation}
\mathcal{Q}\left(b_1\right)=\mathcal{Q}\left(b_2\right). 
\label{error_event_eq}
\end{equation}
Assuming noise-free measurements, there are two events that could possibly lead to \eqref{error_event_eq}: (i) where both $b_1$ and $b_2$ fall within the same quantization bin; and (ii) where both $b_1$ and $b_2$ are saturated, that is, $b_1,b_2\geq \tau_{k-1}$. The probability of the first event is upper-bounded by that of the event $\mathcal{E}_1=\left\{\left|b_1-b_2 \right|<\delta\right\}$, where 
\begin{equation}
\delta=\underset{1\leq j \leq k-1}{\max}\text{\,\,}\left(\tau_j-\tau_{j-1}\right),
\label{delta_def_eq}
\end{equation}
is the the \textit{precision} of the quantizer, indicating that $b_1$ and $b_2$ are closer apart than the precision of the quantizer. The probability of the second event is denoted as $\mathcal{E}_2=\left\{b_1,b_2\geq \tau_{k-1}\right\}$, which arises in the case of measurement \textit{saturation}. The probability $P_e$ satisfies  $P_e\leq P\left(\mathcal{E}_1\right)+P\left(\mathcal{E}_2\right)$. In order to place an upper-bound on $P_e$, we bound the error probabilities $P\left(\mathcal{E}_1\right)$ and $P\left(\mathcal{E}_2\right)$ separately.
\subsubsection{An Upper Bound on $P\left(\mathcal{E}_1\right)$}
Consider the separation between the quadratic measurements of $\boldsymbol x_1$ and $\boldsymbol x_2$, given by 
\begin{eqnarray}
\psi&=&\left|b_1-b_2\right| = \left| \left|\boldsymbol a^\top \boldsymbol x_1\right|^2 - \left|\boldsymbol a^\top \boldsymbol x_2\right|^2\right|\nonumber\\ &=& |\boldsymbol a^\top \underbrace{\left(\boldsymbol x_1 \boldsymbol x_1^\top - \boldsymbol x_2 \boldsymbol x_2^\top\right)}_{\boldsymbol V} \boldsymbol a |.
\label{gammaeq}
\end{eqnarray}
An important property of $\boldsymbol V=\boldsymbol x_1 \boldsymbol x_1^\top - \boldsymbol x_2 \boldsymbol x_2^\top$ is established in Proposition \ref{error_prob_prop1}, eventually leading to a bound on $P\left(\mathcal{E}_1\right)$. 
\begin{propos}
\label{error_prob_prop1}
Let $\boldsymbol x_1$ and $\boldsymbol x_2$ be two linearly independent vectors in $\mathbb{U}^n$. The matrix $\boldsymbol V = \boldsymbol x_1 \boldsymbol x_1^\top - \boldsymbol x_2 \boldsymbol x_2^\top$ has two nonzero eigenvalues of equal magnitude $\nu_1$, such that $\nu_1^2=1-\rho^2$, where $\rho=\left|\boldsymbol x_1^\top \boldsymbol x_2\right|$ is the coefficient of correlation. 
\end{propos}
\noindent\textit{Proof:} Observe that $\boldsymbol V$ has two nonzero eigenvalues, since, for any vector $\boldsymbol u$ in the orthogonal complement of $\text{span}\left\{ \boldsymbol x_1,\boldsymbol x_2 \right\}$, we have $\boldsymbol V \boldsymbol u=\boldsymbol 0$. Therefore, the eigenvectors corresponding to the nonzero eigenvalues must be of the form $\left(\alpha \boldsymbol x_1+\beta \boldsymbol x_2\right)$, for scalars $\alpha$ and $\beta$ that are not simultaneously equal to zero. If the corresponding eigenvalue is $\nu_1$, then we have 
\begin{equation}
\left(\boldsymbol x_1 \boldsymbol x_1^\top - \boldsymbol x_2 \boldsymbol x_2^\top\right)\left(\alpha \boldsymbol x_1+\beta \boldsymbol x_2\right)=\nu_1\left(\alpha \boldsymbol x_1+\beta \boldsymbol x_2\right).
\label{eval_evec_eq}
\end{equation}
Since $\boldsymbol x_1$ and $\boldsymbol x_2$ are linearly independent and of unit norm, by comparing terms in \eqref{eval_evec_eq}, we get that
\begin{eqnarray*}
\alpha+\beta \left(\boldsymbol x_1^\top \boldsymbol x_2\right)=\nu_1\alpha, \text{\,\,and\,\,}\alpha\left(\boldsymbol x_2^\top \boldsymbol x_1\right)+\beta=-\nu_1\beta.
\label{eval_evec_eq1}
\end{eqnarray*}
Writing $\alpha=-\frac{\beta \left(\boldsymbol x_1^\top \boldsymbol x_2\right)}{1-\nu_1}$ and substituting in $\alpha\left(\boldsymbol x_2^\top \boldsymbol x_1\right)+\beta=-\nu_1\beta$, and using the fact that $\left(\boldsymbol x_1^\top \boldsymbol x_2\right)\left(\boldsymbol x_2^\top \boldsymbol x_1\right)=\rho^2$, we have
\begin{eqnarray*}
\beta\left(1+\nu_1-\frac{\rho^2}{1-\nu_1}  \right)=0,
\label{eval_evec_eq2}
\end{eqnarray*}
leading to $\nu_1^2=1-\rho^2$, since $\beta \neq 0$.\hfill $\blacksquare$\\
Therefore, considering the separation $\psi$ in \eqref{gammaeq} and invoking the eigenvalue decomposition (EVD) $\boldsymbol V=\boldsymbol R \boldsymbol D \boldsymbol R^\top$, we get
\begin{eqnarray}
\psi &=& |\boldsymbol a^\top \boldsymbol V \boldsymbol a| = \left|\boldsymbol a^\top  \boldsymbol R \boldsymbol D \boldsymbol R^\top \boldsymbol a \right| = |\tilde{\boldsymbol a}^\top \boldsymbol D \tilde{\boldsymbol a}|, 
\label{gamma_eval_eq}
\end{eqnarray}
where $\tilde{\boldsymbol a} = \boldsymbol R^\top \boldsymbol a$. The top two diagonal entries of $\boldsymbol D$ have identical magnitudes $\nu_1 = \sqrt{1-\rho^2}$, and the remaining diagonal entries are zero. Taking this property into account, the separation may be bounded from below as
\begin{eqnarray}
\psi \geq \sqrt{1-\rho^2} | \tilde a_1^2 - \tilde a_2^2|,
\label{psi_ineq1_qpr_ch}
\end{eqnarray} 
where $\tilde a_1$ and $\tilde a_2$ are respectively the first and second entries of $\tilde{\boldsymbol a}$, which is a Gaussian random vector since $\boldsymbol R$ is orthonormal and $\boldsymbol a \sim \mathcal{N}\left(\boldsymbol 0, \boldsymbol I_n\right)$. Hence, the probability of $\mathcal{E}_1$ can be upper-bounded as follows:
\begin{eqnarray}
P\left(\mathcal{E}_1\right) &=& P(\psi \leq \delta) \stackrel{\text{(a)}}{\leq} P(\sqrt{1-\rho^2} | \tilde a_1^2 - \tilde a_2^2| \leq \delta),\nonumber\\
\Rightarrow P\left(\mathcal{E}_1\right)&\leq& P\left(\left|\tilde{a}_1^2  - \tilde{a}_2^2\right| < \frac{\delta}{\sqrt{1-\rho^2}}\right),
\label{pe1_bound_final1}
\end{eqnarray}
where the inequality (a) in \eqref{pe1_bound_final1} follows from the fact that the event $\left\{\psi \leq \delta\right\}$ implies that $\left\{\sqrt{1-\rho^2} | \tilde a_1^2 - \tilde a_2^2| \leq \delta\right\}$ happens, as a consequence of \eqref{psi_ineq1_qpr_ch}. The random variables $\tilde{a}_1^2$ and $\tilde{a}_2^2$ are independent and follow the $\chi^2_1$ distribution (cf. Proposition~\ref{meas_dist_prop}). The moment generating function (m.g.f.) of the $\chi^2_1$ random variable is 
\begin{equation*}
M_{\chi^2_1}(t)=(1-2t)^{-1/2}, \text{\,\,for\,\,} t < \frac{1}{2}. 
\end{equation*}
Let $u=\tilde{a}_1^2 - \tilde{a}_2^2$, which is the difference between two such independent random variables. The m.g.f. of $u$ turns out to be 
\begin{equation}
M_{u}(t)=\left(1-4t^2\right)^{-\frac{1}{2}}=\left(\frac{\frac{1}{4}}{\frac{1}{4}-t^2}\right)^{\frac{1}{2}}, |t|<\frac{1}{2}.
\label{mgf_eq}
\end{equation}
Comparing \eqref{mgf_eq} with the m.g.f. of the \textit{variance-gamma} distribution \cite{vargamma_ref} with parameters $\mu_0$, $\alpha$, $\beta$, and $\lambda_0$, given by
\begin{equation}
M(t)=e^{\mu_o t}\left(  \frac{\alpha^2-\beta^2}{\alpha^2-\left(\beta+t\right)^2}\right)^{\lambda_0}, 
\label{mgf_var_gamma}
\end{equation}
we obtain an equivalence of \eqref{mgf_var_gamma} and \eqref{mgf_eq} with $\mu_0=0$, $\alpha=\frac{1}{2}$, $\beta=0$, and $\lambda_0=\frac{1}{2}$. The probability density function (p.d.f.) corresponding to the m.g.f. in \eqref{mgf_eq} is given by
\begin{equation}
f_{u}\left(u\right)=\frac{1}{2\pi}K_0\left( \frac{|u|}{2} \right), 
\label{pdf_var_gamma}
\end{equation}
where $K_0$ is the modified Bessel-function of the second kind and zeroth order. Since the p.d.f. in \eqref{pdf_var_gamma} is symmetric, the upper-bound on $P\left(\mathcal{E}_1\right)$ in \eqref{pe1_bound_final1} reduces to
\begin{equation}
P\left(\mathcal{E}_1\right)\leq \underbrace{\frac{2}{\pi}\int_{0}^{\frac{\delta}{\sqrt{1-\rho^2}}} K_0\left( u \right) \mathrm{d}u}_{g(\delta')},
\label{pe1_bound_final_vargamma_1}
\end{equation}
where $\delta' = {\frac{\delta}{\sqrt{1-\rho^2}}}$. In order to obtain a more readily interpretable upper bound, we approximate the integral in \eqref{pe1_bound_final_vargamma_1} as $\hat{g}(\delta')=1-\exp\left(-1.6\delta'\right)$. Figure~\ref{measurements_required_figure}(a) shows that $\hat{g}$ is a reliable and accurate approximation to $g(u)$. Consequently, 
\begin{equation}
P\left(\mathcal{E}_1\right)\leq 1-\exp\left(-1.6{\frac{\delta}{\sqrt{1-\rho^2}}}\right).
\label{pe1bound}
\end{equation}
\subsubsection{Computing an Upper Bound on $P\left(\mathcal{E}_2\right)$} Since $\mathcal{E}_2\stackrel{\Delta}{=}\left\{b_1, b_2\geq \tau_{k-1}\right\}$, which corresponds to {\it both} $b_1$ and $b_2$ going into saturation, where $b_1$ and $b_2$ are not independent. The event $\mathcal{E}_2$ is clearly a subset of the event $\left\{b_1 \geq \tau_{k-1}\right\}$, and using Proposition \ref{meas_dist_prop}, the probability of $\mathcal{E}_2$ can be bounded as
\begin{equation}
P\left(\mathcal{E}_2\right)\leq P\left(b_1 \geq \tau_{k-1}\right)=1-\frac{\gamma\left(\frac{\tau_{k-1}}{2},\frac{1}{2}\right)}{\sqrt{\pi}}.
\label{bound_on_pe2}
\end{equation}
Finally, we combine the upper bounds on the error events $\mathcal{E}_1$ and $\mathcal{E}_2$ to obtain a bound on $P_e$, which dictates the minimum number of measurements required to ensure that two linearly independent signals can be discerned from their quantized intensity measurements.
\subsubsection{An Upper Bound on $P_e$}
\indent Combining \eqref{pe1bound} and \eqref{bound_on_pe2} results in an upper bound on $P_e$, given by
\begin{equation}
P_e \leq P_e^{\max} = 2-\frac{\gamma\left(\frac{\tau_{k-1}}{2},\frac{1}{2}\right)}{\sqrt{\pi}}-\exp\left(-\frac{1.6\delta}{\sqrt{1-\rho^2}}\right),
\label{final_bound_on_pe}
\end{equation}
which, we recall, is the probability with which the measurements would be indistinguishable. The upper bound, in turn, must be less than unity for it to be meaningful. The quantizer parameters $\delta$ and $\tau_{k-1}$ may be chosen accordingly.
\begin{figure}[t]
\centering
\subfigure[$g(u)$ and $\hat{g}(u)$ versus $u$]{
\includegraphics[width=1.5in]{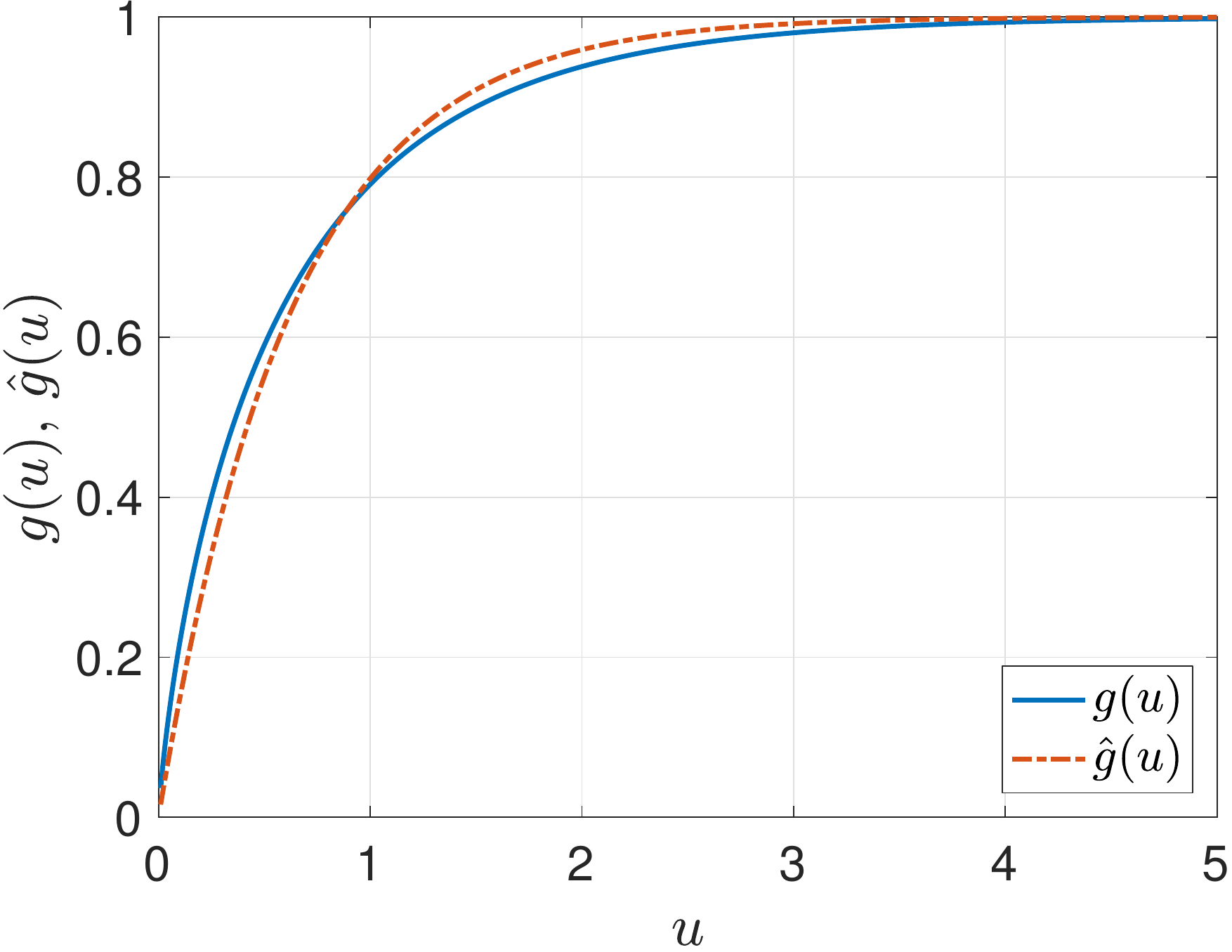}}
\subfigure[$m_{\min}$ versus $\delta$, for different $\rho$]{
\includegraphics[width=1.5in]{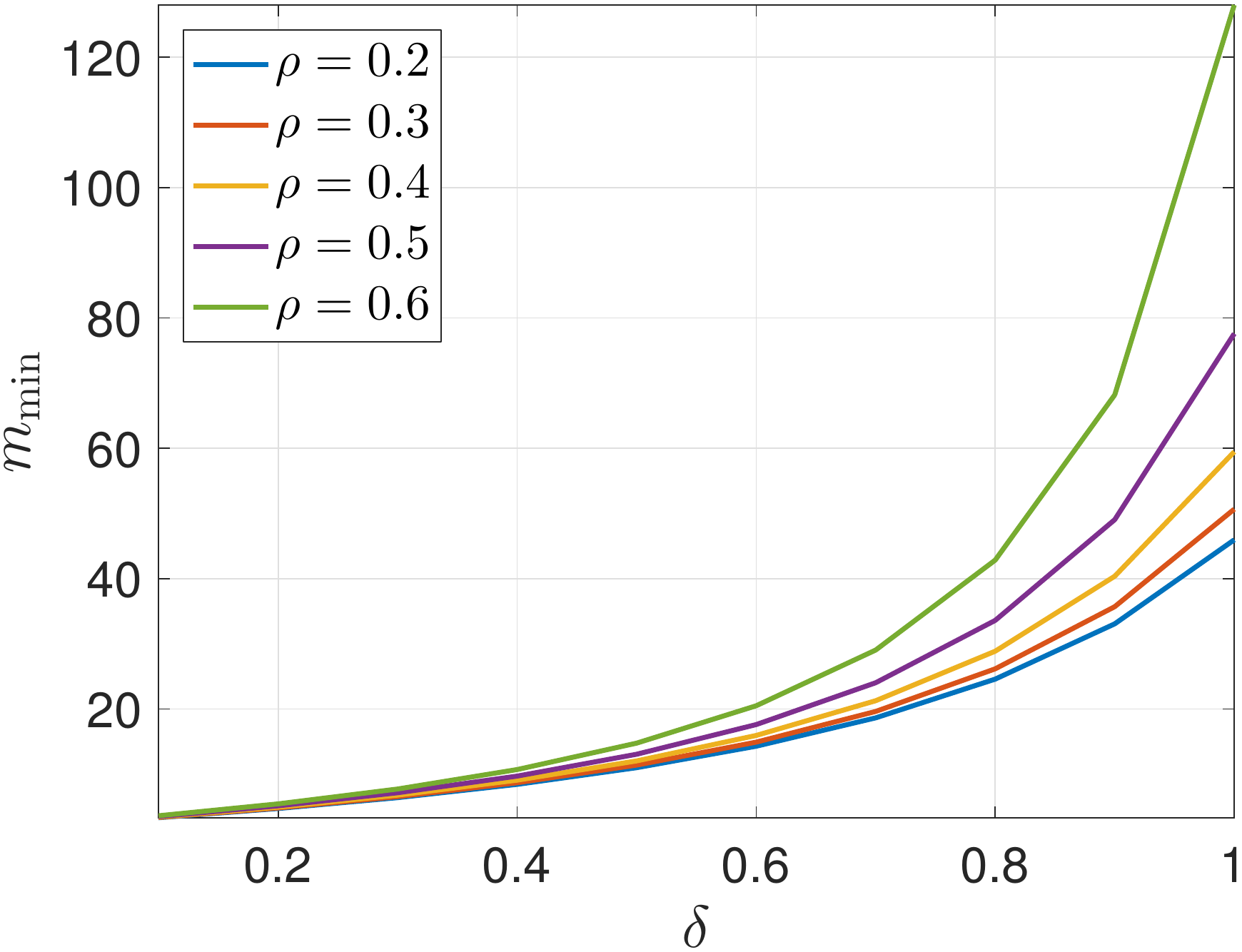}}
\caption{\small{\small (a) $g(u)$ and its approximation $\hat{g}(u)$; and (b) Minimum number of measurements $m_{\min}$ required to maintain the distinguishability of quantized measurements with high probability as a function of quantizer precision $\delta$, for different values of the correlation $\rho$.}}
\label{measurements_required_figure}
\end{figure}
%%%%%%%%%%%%%%%%%%%%%%%%%%%%%%%%%%%%%%%%%%%%%
\subsubsection{A Numerical Example}
\indent Consider $m$ independent measurements corresponding to $\boldsymbol x_1$ and $\boldsymbol x_2$. The overall probability that the measurements will be indistinguishable is upper-bounded as $P_e^{\text{overall}} \leq (P_e^{\max})^m.$ If we desire to keep this value below a certain $\epsilon$, the minimal number of measurements would be $m_{\min} = \log \epsilon/ \log P_e^{\max}$. Consider the case where $\rho=\left|\boldsymbol x_1^\top \boldsymbol x_2\right|=0.6$ and a four-bit quantizer ($k=16$), with equiprobable intervals, meaning that, $P\left(b\leq \tau_j\right)=\frac{j}{k}$, for $j = 1:k$. Such a quantizer ensures that roughly the same number of measurements fall in each interval, for large enough number of measurements $m$. The illustrative values chosen for $\rho$ and $k$ are indicative of a fair degree of correlation between $\boldsymbol x_1$ and $\boldsymbol x_2$ and coarse quantization. The values of $\delta$ and $\tau_{k-1}$ for this particular choice turn out to be $\delta=1.1162$ and $\tau_{k-1}=3.4698$. The corresponding upper bound on $P_e$ in \eqref{final_bound_on_pe} evaluates to $P_e^{\max} = 0.9552$. For $m$ independent measurements, we have $P_e^{\text{overall}} \leq 0.9552^m$. If we desire to achieve $P_e^{\text{overall}}\leq 0.01$, we must have $m\geq 101$.\\
\indent The minimum number of measurements $m_{\min}$ required to maintain $P_e^{\text{overall}}\leq 0.01$ is shown in Figure~\ref{measurements_required_figure}(b) as a function of $\delta$, for various values of $\rho$. The value of $\tau_{k-1}$ is kept fixed at $\tau_{k-1}=2.7056$, which corresponds to $P\left(b\leq \tau_{k-1}\right)=0.9$, thereby guaranteeing that the probability of measurement saturation is no more than $0.1$. For a fixed $\rho$, the number of required measurements $m_{\min}$ increases with increasing $\delta$ (coarse quantization). Moreover, for a given quantizer precision $\delta$, one needs to collect more measurements in order to maintain distinguishability as the correlation $\rho$ increases. The upper bound on $P_e$ in \eqref{final_bound_on_pe} is not tight and the minimum number of measurements required, as given by the preceding analysis, is independent of the reconstruction algorithm, and is necessary, but not sufficient.
%%%%%%%%%%%%%%%%%%%%%%%%%%%%%%%%%%%%%%%%%%%%%
\begin{figure}[t]
\centering
\subfigure[c.d.f. of $b=\left| \boldsymbol a^\top \boldsymbol x^* \right|^2$]{
\includegraphics[width=1.5in]{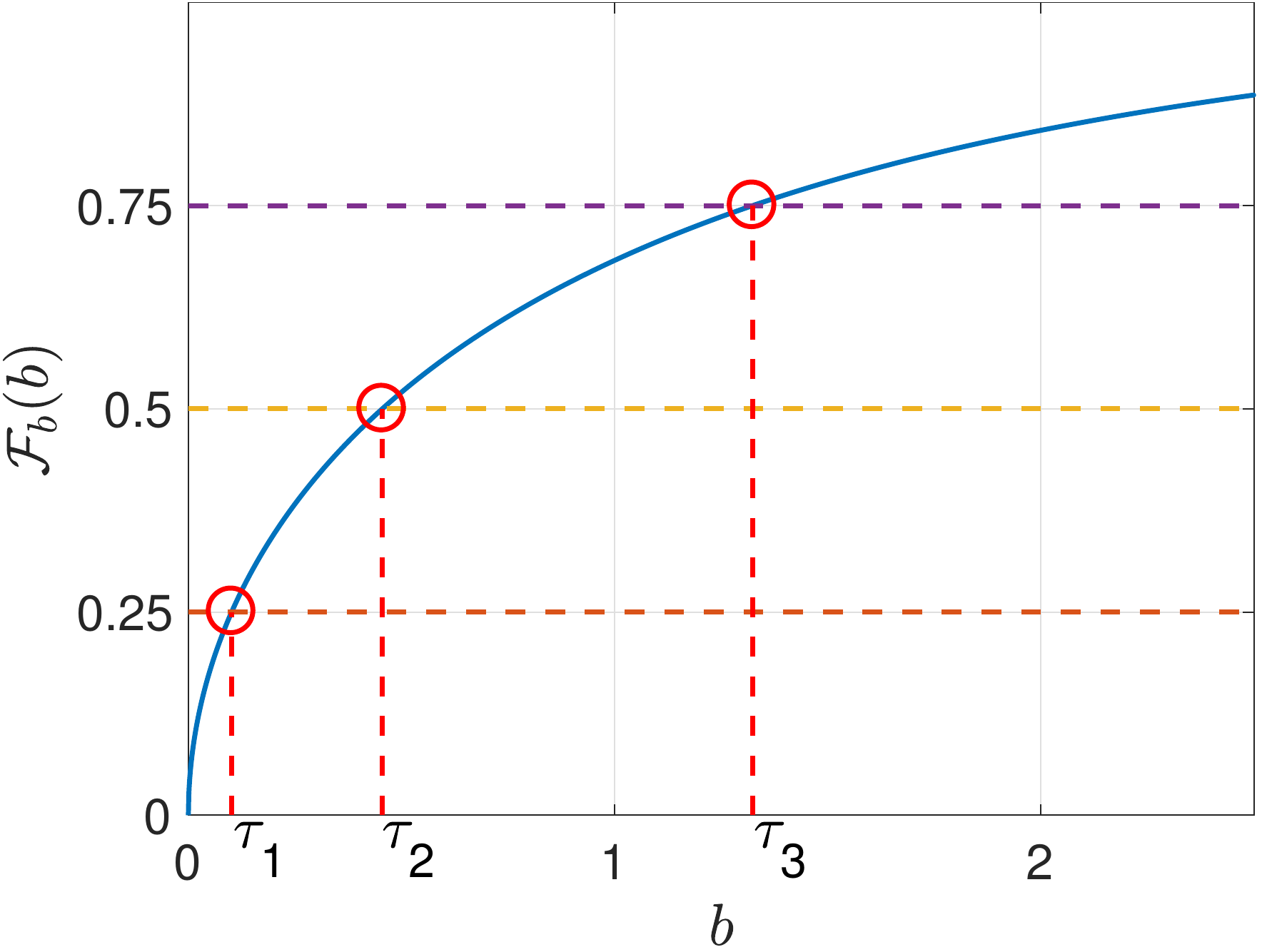}}
\subfigure[Quantization SNR profile]{
\includegraphics[width=1.5in]{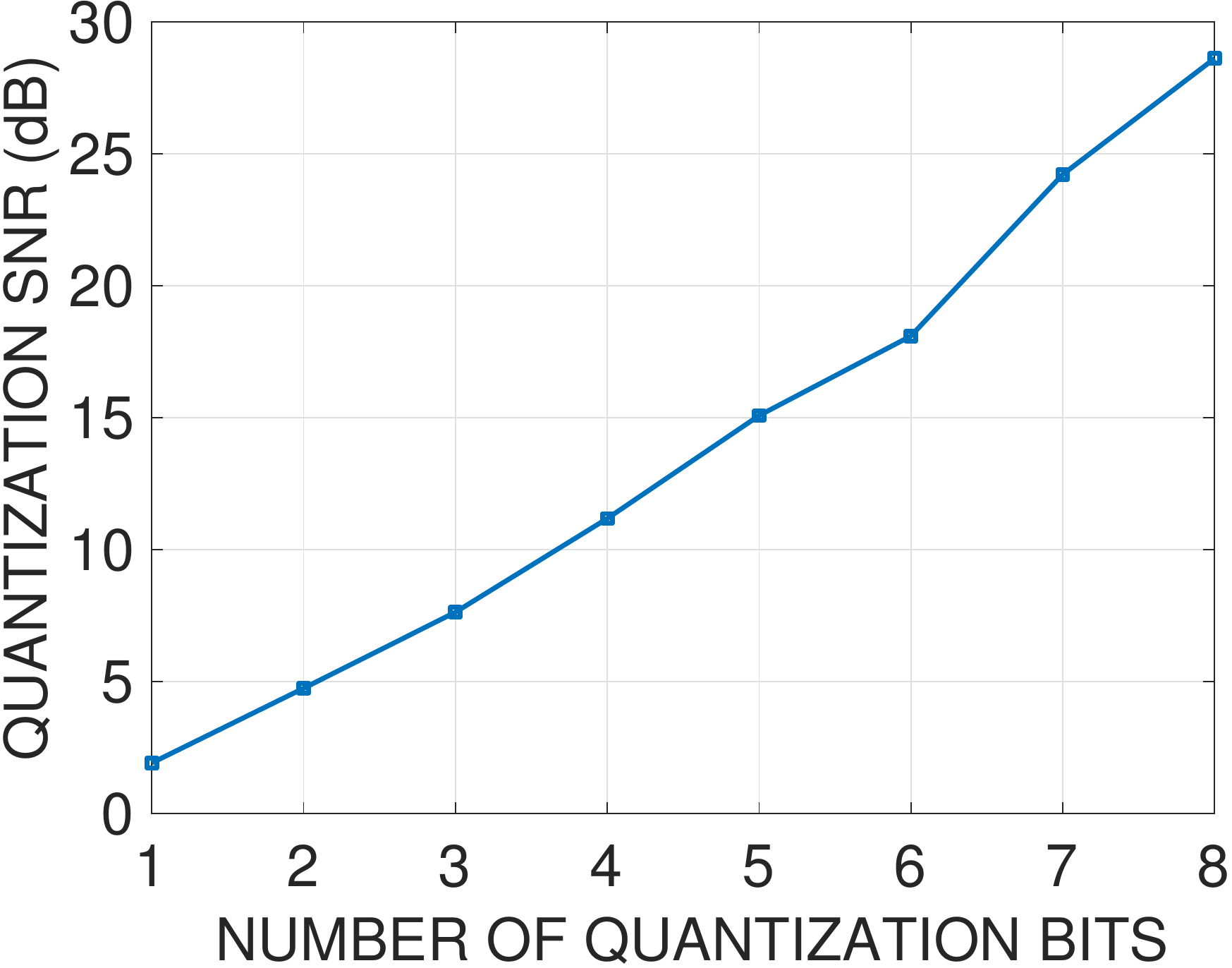}}
\caption{\small{An illustration of quantizer design and quantization SNR: (a) c.d.f. of the quadratic measurement $b=\left| \boldsymbol a^\top \boldsymbol x^* \right|^2$ and the thresholds of a four-level quantizer with equiprobable intervals; and (b) Average quantization SNR as a function of $k$.}}
\label{Femp_real_figure}
\end{figure}
%%%%%%%%%%%%%%%%%%%%%%%%%%%%%%%%%%%%%%%%%%%%%
\subsection{The Issue of Quantizer Design}
\label{quant_design_sec}
As shown in Section~\ref{effect_of_quant_sec}, a quadratic measurement of the form $b=\left| \boldsymbol a^\top \boldsymbol x^*\right|^2$ of a signal $\boldsymbol x^* \in \mathbb{U}^n$ follows the c.d.f. $\mathcal{F}_b\left(b\right)=\frac{\gamma\left(\frac{b}{2},\frac{1}{2}\right)}{\sqrt{\pi}}$, plotted in Figure \ref{Femp_real_figure}(a). The quantizer may be designed using the Lloyd-Max algorithm \cite{jayant_noll}, which jointly optimizes for the thresholds and the codebook, such that the quantization SNR, defined as
\begin{equation}
\text{SNR}_{\text{quant}}=\frac{\sum_{i=1}^{m}b_i^2}{\sum_{i=1}^{m}\left(b_i-y_i\right)^2},
\label{input_SNR_eq}
\end{equation}  
is maximized, where $b_i$ and $y_i$ are as defined in \eqref{pr_int_model} and \eqref{QPR_measurement_eq}, respectively. As emphasized in Section \ref{sec_problem_formulation_qpr}, the proposed formalism based on consistent recovery is independent of the choice of encoding symbols. This aspect will also become clear as we develop the optimization framework for QPR in Section \ref{formal_opt_lift_const_qpr_sec}. Consequently, there is no guarantee that maximization of $\text{SNR}_{\text{quant}}$ would lead to a superior phase retrieval performance. In the present context, it would be more appropriate to optimize the thresholds such that the expected estimation error, given by
\begin{equation}
\left(\tau^*_1,\cdots,\tau^*_{k-1}\right)=\arg\underset{\tau_1,\cdots,\tau_{k-1}}{\min}\text{\,\,}\mathbb{E}\left\|\hat{\boldsymbol x}\left(\tau_1,\cdots,\tau_{k-1}\right)-\boldsymbol x^*   \right\|_2^2,
\label{threshold_opt}
\end{equation}
where the expectation is calculated over the distribution of the sampling vectors $\boldsymbol a_i$, is minimized.~Unfortunately, solving \eqref{threshold_opt} iteratively or in closed-form is mathematically intractable. Therefore, we adopt an approach similar to that proposed by Zymnis et al. \cite{zymnis_qcs} in the context of quantized CS, wherein the quantizer is designed to have equiprobable intervals (along the lines of \textit{companding} \cite{bstj_paper}). The design of such a quantizer is facilitated by the knowledge of the c.d.f. as illustrated in Figure \ref{Femp_real_figure}(a) for $k=4$ levels. The thresholds are marked as $\tau_i, i = 0:4$ with $\tau_0=0$ and $\tau_4=+\infty$. For the purpose of illustration, we show in Figure \ref{Femp_real_figure}(b), $\text{SNR}_{\text{quant}}$ as a function of the number of bits $\log_2k$ for equiprobable-interval quantizers. The encoding symbols for the levels $j = 1:k-1$ are taken as the midpoints of the corresponding intervals, whereas the encoding symbol for the $k^{\text{th}}$ level is set to $s_k=\tau_{k-1}+\frac{\delta}{2}$, where $\delta$ is as defined in \eqref{delta_def_eq}. We observe from Figure \ref{Femp_real_figure}(b) that $\text{SNR}_{\text{quant}}$ tends to increase almost linearly as the number of bits increases, and attains a value of approximately $29$ dB when the number of quantization bits is $8$. However, for coarse quantization, $\text{SNR}_{\text{quant}}$ is about $10$ dB or lower, which is far too low for the existing PR algorithms, considering that one models the quantization noise as additive. This underscores the need for developing a quantization-aware PR algorithm.
\begin{figure}[t]
\centering
\subfigure{
\includegraphics[width=2.2in]{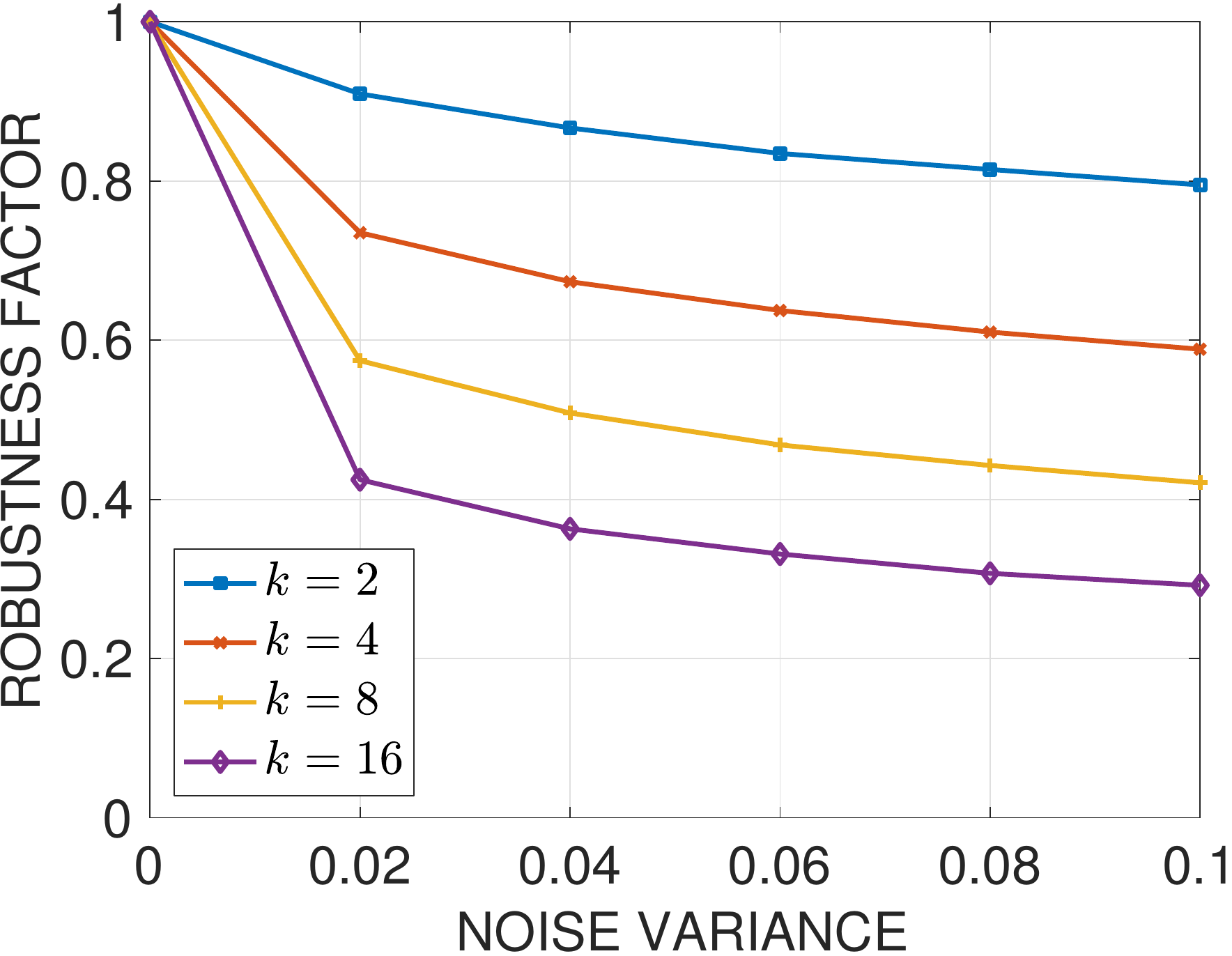}}
\caption{\small{Noise-robustness of quantized measurements versus noise variance, for various number of quantization levels $k$. As the quantization becomes coarser, the measurements become less susceptible to noise.}}
\label{QPR_noise_robustness_figure}
\end{figure}
%%%%%%%%%%%%%%%%%%%%%%%%%%%%%%%%%%%%%%%%%%%%%
\subsection{Noise Robustness of Quantized Measurements}
\label{noise_perf-sec}
In practice, due to noise, the acquired measurements take the form 
\begin{equation}
y_i=\mathcal{Q}\left(\left|   \boldsymbol a_i^\top \boldsymbol x^*   \right|^2+\xi_i\right), i=1:m,
\label{QPR_measurement_eq_noisy}
\end{equation}
where $\left\{\xi_i\right\}_{i=1}^{m}$ are independent and identically distributed (i.i.d.) additive noise samples. The quantization process is inherently noise-robust as long as the perturbation due to noise does not alter the output symbol. We shall illustrate this inherent robustness using Monte-Carlo simulations.\\
\indent Let $\xi\sim\mathcal{N}(0,\sigma_{\xi}^2)$. We define the robustness factor $p_r$ corresponding to a representative quadratic measurement as
\begin{equation}
p_r = P\left(\mathcal{Q}\left(b\right)=\mathcal{Q}\left(b+\xi\right)\right), \text{\,\,where\,\,}b=\left| \boldsymbol a^\top \boldsymbol x^*\right|^2,
\label{QPR_robustness_factor}
\end{equation}
which measures the probability that the noise does not alter the quantized measurement. Recall that the intensity measurement $b$ follows the $\chi^2_1$ distribution. Since it is cumbersome to obtain an analytical expression for $p_r$, for the purpose of illustration, we adopt a Monte-Carlo approach to estimate $p_r$ as 
\begin{equation}
\hat{p}_r = \frac{1}{N_{\text{trial}}}\sum_{t=1}^{N_{\text{trial}}}\mathbbm{1}\left\{\mathcal{Q}\left(b_t\right)=\mathcal{Q}\left(b_t+\xi_t\right)\right\},
\label{QPR_robustness_factor_approx}
\end{equation}
where $\mathbbm{1}$ denotes the indicator function, and $\left\{b_t\right\}$ and $\left\{\xi_t\right\}$ are drawn independently from the $\chi^2_1$ and $\mathcal{N}(0,\sigma_{\xi}^2)$ distributions, respectively. The number of trials $N_{\text{trial}}$ is taken as $1000$. The variation of $\hat{p}_r$ as a function of $\sigma_{\xi}^2$, for different number of quantization levels $k$, is shown in Figure~\ref{QPR_noise_robustness_figure}. An equiprobable interval quantizer is considered. We observe from Figure~\ref{QPR_noise_robustness_figure} that the measurements get increasingly robust to noise as the quantization gets coarser. As one would expect, $p_r$ drops monotonically as the noise variance $\sigma_{\xi}^2$ increases. For binary measurements ($k=2$), the noise does not alter the measurements with probability approximately $0.80$, even for relatively high noise variance of $\sigma_{\xi}^2=0.1$. The trade-off, however, is that such a coarse quantization will compromise distinguishability of the measurements.
\section{The Optimization Framework and Reconstruction Algorithms for QPR}
\label{formal_opt_lift_const_qpr_sec}
\subsection{The QPR Optimization Framework}
\indent We combine the requirement of \textit{consistent recovery} with the principle of \textit{lifting} \cite{PL1_candes,PL2_candes} and formulate an appropriate cost function. The central idea behind lifting is to write the quadratic expression $\left| \boldsymbol a_i^\top \boldsymbol x  \right|^2$ as 
\begin{equation*}
\left| \boldsymbol a_i^\top \boldsymbol x  \right|^2=\text{Tr}\left(\boldsymbol A_i \boldsymbol X\right),
\label{lifting_eq}
\end{equation*}   
where $\boldsymbol X = \boldsymbol x \boldsymbol x^\top$, $\boldsymbol A_i = \boldsymbol a_i \boldsymbol a_i^\top$, and $\text{Tr}(\cdot)$ denotes the trace operator. The two representations are equivalent, but the advantage of the lifted version is that it enables one to express the quadratic measurements in 1-D as a set of linear measurements in 2-D. We assume that $m_j$ measurements are encoded with the symbol $s_j$ and denote the sampling vectors in $\mathcal{H}_j$ as $\{\boldsymbol a_{i}^j\}_{i=1}^{m_j}$, for $j=1:k$. Since the matrix $\boldsymbol X$ is a rank-1 and positive semi-definite (PSD) matrix by construction, it would be imperative to enforce these conditions. Effectively, we seek ${\boldsymbol X}$ such that ${\boldsymbol X}\succeq 0$, $\text{rank}\left({\boldsymbol X}\right)=1$, and
\begin{equation}
\tau_{j-1}\leq \text{Tr}\left(\boldsymbol A_{i}^{j}{\boldsymbol X}\right)\leq\tau_j; \text{\,\,}i=1: m_j,\text{\,\,} j=1: k, 
\label{recovery_eq1}
\end{equation}
where $\boldsymbol A_{i}^{j}=\boldsymbol a_{i}^j(\boldsymbol a_{i}^j)^\top$. We incorporate the inequality constraints in \eqref{recovery_eq1} arising out of the consistency criterion in the optimization objective by using the one-sided quadratic function $f:\mathbb{R}\rightarrow \mathbb{R}$: 
\begin{equation*}
\label{def_cost}
 f(u) = 
\begin{cases}
\frac{1}{2}u^2, & \text{\,\,if\,\,}u\leq 0,\text{\,and}\\
0, & \text{\,\,otherwise}.
\end{cases}
\end{equation*}
Therefore, the QPR problem may be formulated as
\begin{equation}
\hat{\boldsymbol X}=\arg\underset{\boldsymbol X \succeq 0}{\min}\text{\,} F(\boldsymbol{X}) \text{\,\,\,subject to\,\,\,} \text{rank}\left({\boldsymbol X}\right)=1,
\label{uc_recovery_sdp}
\end{equation}
where the optimization objective $F\left(\boldsymbol X \right)$ is given by
\small
\begin{equation}
F\left(\boldsymbol X \right)=\sum_{j=1}^{k}\sum_{i=1}^{m_j}f\left(\tau_j-\text{Tr}\left(\boldsymbol A_{i}^j \boldsymbol X\right)\right)+f\left(\text{Tr}\left(\boldsymbol A_{i}^j \boldsymbol X\right)-\tau_{j-1}\right).
\label{F_def_eq}
\end{equation}
\normalsize
Although the QPR objective function $F\left(\boldsymbol X \right)$ is convex in $\boldsymbol X$, the minimization posed in \eqref{uc_recovery_sdp} is non-convex, because of the rank constraint.
\subsection{Reconstruction Algorithms for QPR}
We develop two projected gradient-based algorithms to solve \eqref{uc_recovery_sdp}, wherein one retains the best rank-1 approximation of the estimate obtained following a gradient-based update. The algorithms could be terminated whenever the measurement consistency requirement in \eqref{recovery_eq1} is met or when a maximum number of iterations $N_{\text{iter}}$ have elapsed. The proposed algorithmic framework is amenable to accommodating constraints such as positivity, sparsity, etc., which are relevant in many practical imaging modalities. The sparsity prior may be incorporated by hard-thresholding the estimate obtained subsequent to applying the rank-1 constraint, akin to the approach we developed in \cite{css_sparsePR}.
\subsubsection{Projected Gradient-Descent (PGD) for Quantized PR}
\indent The first algorithm employs a simple projected gradient-descent technique, wherein one iteratively computes an update of the form
\begin{equation}
{\boldsymbol X}^{t+1}=\mathcal{P}_{\text{rank}-1}\left({\boldsymbol X}^{t}-\eta^t  \boldsymbol G^t\right),
\label{pgd_algo}
\end{equation}
starting with an initial estimate ${\boldsymbol X}^0$, where $\eta^t$ is the step-size parameter, and $\boldsymbol G^t=\left. \nabla F\left(\boldsymbol X\right)\right|_{\boldsymbol X=\boldsymbol X^t}$ is the gradient matrix of $F\left(\boldsymbol X\right)$ evaluated at $\boldsymbol X^t$. We shall refer to \eqref{pgd_algo} as the {\it QPR update}. The rank-1 projection operator $\mathcal{P}_{\text{rank}-1}$ applied on a symmetric matrix $\boldsymbol Y$ of size $n\times n$ is defined as follows:
\begin{equation*}
\mathcal{P}_{\text{rank}-1}\left(\boldsymbol Y\right)=\max\left\{\lambda_{\max},0\right\}\boldsymbol v_1 \boldsymbol v_1^\top, 
\end{equation*}
where $\lambda_{\max}$ is the largest eigenvalue of $\boldsymbol Y$, having $\boldsymbol v_1$ as the associated eigenvector. An estimate of the underlying ground-truth signal ${\boldsymbol x}^*$ can be obtained as ${\boldsymbol x}^{t+1} = \sqrt{\lambda_{\text{max}}}{\boldsymbol v}_1$.\\
\indent Cai et al.~developed a similar singular-value thresholding algorithm, albeit with a soft-threshold, for solving the problem of low-rank matrix recovery from linear measurements \cite{svt_candes}. Calculating the gradient of $F\left(\boldsymbol X \right)$ requires the gradient of functions of the form $h\left(\boldsymbol X\right)=f\left(\text{Tr}\left(\boldsymbol A \boldsymbol X\right) -\tau\right)$, which is given by
\begin{equation*}
\nabla h\left(\boldsymbol X\right)=f'\left(\text{Tr}\left(\boldsymbol A \boldsymbol X\right) -\tau\right)\boldsymbol A^\top,
\label{grad_expr_eq}
\end{equation*}
where, for any $u \in \mathbb{R}$, $f'(u)$ is given by
\begin{equation*}
\label{def_cost_derivative}
 f'(u) = 
\begin{cases}
u, & \text{\,\,if\,\,}u\leq 0,\text{\,and}\\
0, & \text{\,\,otherwise}.
\end{cases}
\end{equation*}
To incorporate sparsity of ${\boldsymbol x}^*$, the QPR update is subjected to a hard-thresholding operation of the form
\begin{equation}
{\boldsymbol x}^{t+1} \leftarrow \mathcal{P}_{s}\left({\boldsymbol x}^{t+1}\right),
\label{pgd_algo_compressive}
\end{equation}
where $\mathcal{P}_{s}$ returns the best $s$-sparse approximation of its argument and is obtained by picking the top $s$ entries in magnitude. We refer to \eqref{pgd_algo_compressive} as the SQPR update. To determine $\eta^t$, we adopt the \textit{exact line-search} strategy by solving
\begin{equation}
\eta^t=\arg\underset{\eta>0}{\min}\text{\,\,}F\left(\boldsymbol X^t-\eta\boldsymbol G^t \right),
\label{exact_LS_eq}
\end{equation}
using a grid search over a chosen interval. The steps involved in QPR and SQPR are listed in Algorithm~\ref{qprSQPR}. The PGD algorithm for QPR possesses the \textit{descent property}, that is, the updates generated by \eqref{pgd_algo} satisfy $F\left(\boldsymbol X^{t+1}\right)\leq F\left(\boldsymbol X^{t}\right)$. A proof of this claim is provided in Section~\ref{pgd_descent_proof_qpr} of the supplementary material.
\begin{algorithm}[t]
\caption {QPR and SQPR (based on PGD) for phase retrieval in the presence of measurement quantization.}
\begin{algorithmic}
\STATE {\bf  1.} {\bf Initialization:} Set $t = 0$, $\boldsymbol X^{t}= \boldsymbol 0_{n\times n}$, and maximum iteration count $N_{\text{iter}}$.
\STATE \textbf{2. For} $t=1:N_{\text{iter}}$, \textbf{do}:
\begin{enumerate}
\item ${\boldsymbol X}^{t+1}=\mathcal{P}_{\text{rank}-1}\left({\boldsymbol X}^{t}-\eta^t  \nabla \left. F\left( \boldsymbol X\right)\right|_{\boldsymbol X=\boldsymbol X^t}\right)$.
\item For SQPR, perform the $s$-sparse approximation:
\begin{equation*}
{\boldsymbol x}^{t+1}\leftarrow\mathcal{P}_{s}\left({\boldsymbol x}^{t+1}\right), \text{\,where\,} {\boldsymbol X}^{t+1}={\boldsymbol x}^{t+1}\left({\boldsymbol x}^{t+1}\right)^\top.
\end{equation*}
\end{enumerate}
\STATE {\bf  3.} {\bf Output:} the current estimate ${\boldsymbol x}^{t}$.
\end{algorithmic}
\label{qprSQPR}
\end{algorithm}
\begin{algorithm}[t]
\caption{QPR-A and SQPR-A based on APGD for quantized PR.}
\begin{algorithmic}
\STATE {\bf  1.} {\bf Initialization:} Set $t = 0$, $\boldsymbol X^{t}=\boldsymbol Y^{t} = \boldsymbol 0_{n\times n}$, $\theta^0 = 1$, and $N_{\text{iter}}$ = Maximum iteration count.
\STATE \textbf{2. For} $t=1:N_{\text{iter}}$, \textbf{do}:
\begin{enumerate}
\item ${\boldsymbol X}^{t+1}=\mathcal{P}_{\text{rank}-1}\left({\boldsymbol Y}^{t}-\eta^t  \nabla \left. F\left( \boldsymbol Y\right)\right|_{\boldsymbol Y=\boldsymbol Y^t}\right)$,
\item For SQPR-A, perform a hard-thresholding operation 
\begin{equation*}
{\boldsymbol x}^{t+1}\leftarrow\mathcal{P}_{s}\left({\boldsymbol x}^{t+1}\right), \text{\,where\,} {\boldsymbol X}^{t+1}={\boldsymbol x}^{t+1}\left({\boldsymbol x}^{t+1}\right)^\top,
\end{equation*}
to enforce sparsity, where $s$ denotes the desired number of nonzero elements. In the absence of sparsity prior (that is, QPR-A), set $s=n$.
\item $\theta^{t+1}=2\left(1+\sqrt{1+\frac{4}{\left(\theta^{t}\right)^2}} \right)^{-1}$,
\item $\beta^{t+1}=\theta^{t+1} \left(\frac{1}{\theta^t}-1 \right)$, and 
\item $\boldsymbol Y^{t+1}={\boldsymbol X}^{t+1}+\beta^{t+1}\left({\boldsymbol X}^{t+1}-{\boldsymbol X}^{t}\right)$.
\end{enumerate}
\STATE {\bf  3.} {\bf Output:} the current estimate ${\boldsymbol x}^{t}$.
\end{algorithmic}
\label{algo_qpra_SQPRa}
\end{algorithm}
\subsubsection{Accelerated Projected Gradient-Descent (APGD)} 
\label{APGD_steps_sec} 
\indent Although the cost function in \eqref{F_def_eq} is convex, the rank-1 projection step in \eqref{pgd_algo} is not. In general, an acceleration of the PGD using Nesterov's scheme \cite{nesterov_momentum1} is not guaranteed in this setting. Nonetheless, motivated by the accelerated singular-value hard-thresholding strategy adopted in \cite{geng_yang} for low-rank matrix completion problems, we go ahead with incorporating a momentum factor in the QPR and SQPR algorithms and investigate empirically if it would result in acceleration. It turns out, from the simulation results, that incorporating the momentum factor indeed results in accelerated convergence (Section~\ref{QPR_vs_QPRA_sec} contains the simulation results). The steps of the QPR and SQPR algorithms with acceleration, referred to as QPR-A and SQPR-A, respectively, are summarized in Algorithm~\ref{algo_qpra_SQPRa}, where the step-sizes $\eta^t$ are chosen following \eqref{exact_LS_eq}. The updates for the momentum terms $\theta$ and $\beta$ in Algorithm~\ref{algo_qpra_SQPRa} are based on the recommendations given in \cite{PL2_candes}.
\section{Numerical Experiments Without the Sparsity Constraint}
\label{QPR_results_sec_dense}
\indent In this section, we demonstrate the following via numerical simulations: (i) effect of Nesterov's acceleration scheme; (ii) choice of quantizer design --- Lloyd-Max quantizer (LMQ) versus equiprobable quantizer (EQ); and (iii) comparison of the proposed quantized PR algorithm with state-of-the-art PR techniques in the absence of any external additive noise, and without the assumption of sparsity. An assessment of the robustness to noise and reconstruction with the incorporation of a sparsity prior will be presented in Sections~\ref{np_crb_sec_qpr_ch} and \ref{SQPR_results_sec}, respectively. The state-of-the-art techniques used for comparison are PhaseLift, TWF, and AltMinPhase.\\ 
\indent The PhaseLift approach is implemented using the PGD and the APGD algorithms, and referred to as PL and PL-A, respectively. The Matlab implementation of the TWF algorithm is taken from the authors' website\footnote{\url{http://web.stanford.edu/~yxchen/TWF/}.}. The spectral initialization technique of \cite{netrapalli_altminpr} is employed to initialize PhaseLift, TWF, and AltMinPhase, wherein one sets $\boldsymbol x^0$ to be equal to the eigenvector corresponding to the largest eigenvalue of the matrix $\boldsymbol S=\frac{1}{m}\sum_{i=1}^{m}y_i\boldsymbol a_i \boldsymbol a_i^\top$, normalized to have unity $\ell_2$-norm. The spectral initialization depends on the measurement vector $\boldsymbol y$, which is a function of the encoding symbols. The QPR-A algorithm, however, is initialized with an all-zero vector, thereby avoiding any dependence of the reconstructed signal on the choice of the codebook $\left\{s_1,s_2,\cdots,s_k \right\}$.

\indent Experiments are conducted for the real signal model, where $\boldsymbol x$ is drawn uniformly at random on $\mathbb{U}^n$ and the measurement vectors $\boldsymbol a_i \sim \mathcal{N}\left(\boldsymbol 0,\boldsymbol I_{n}\right)$, with $n=32$. The number of measurements is taken as $m=10n$. The step-size parameter $\eta^t$ is chosen by an exhaustive search over the range $[0,0.005]$, with a grid spacing of $10^{-5}$.
\subsection{Effect of the Momentum Factor: QPR Versus QPR-A}
\label{QPR_vs_QPRA_sec}
\begin{figure}[t]
\centering
\subfigure[QPR-A versus QPR]{
\includegraphics[width=1.5in]{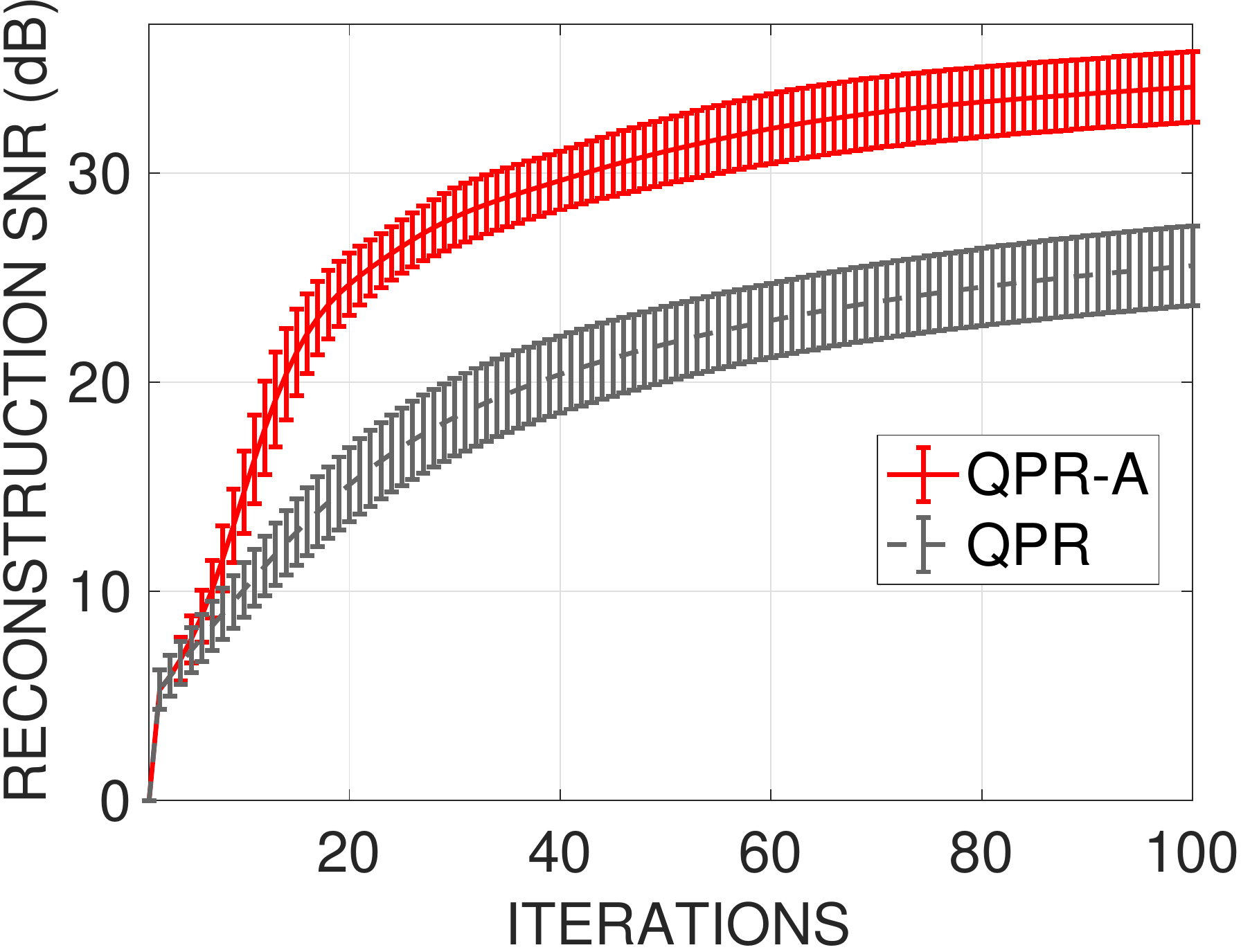}}
\subfigure[PL-A versus PL]{
\includegraphics[width=1.5in]{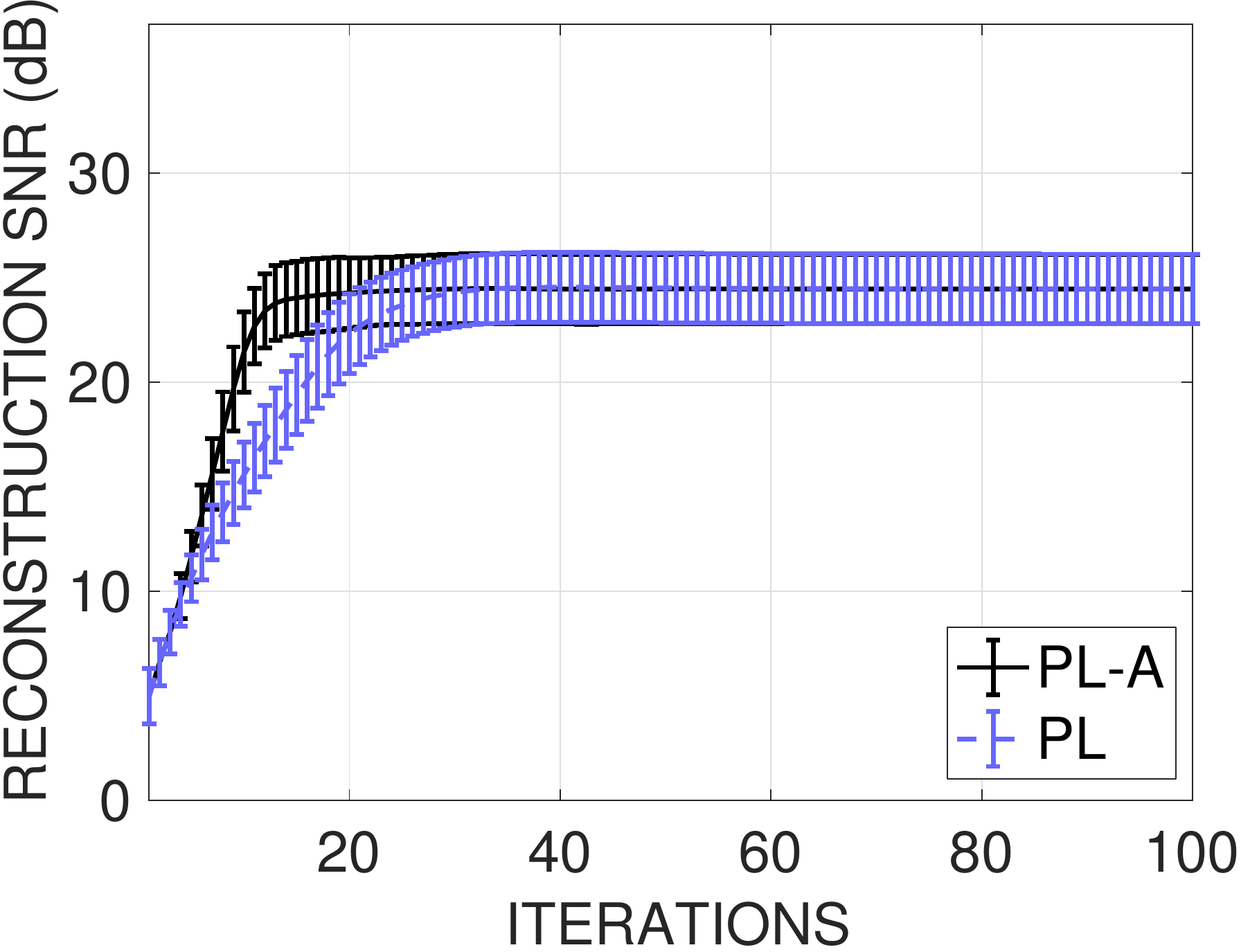}}\\
\subfigure[QPR-A reconstruction]{
\includegraphics[width=1.5in]{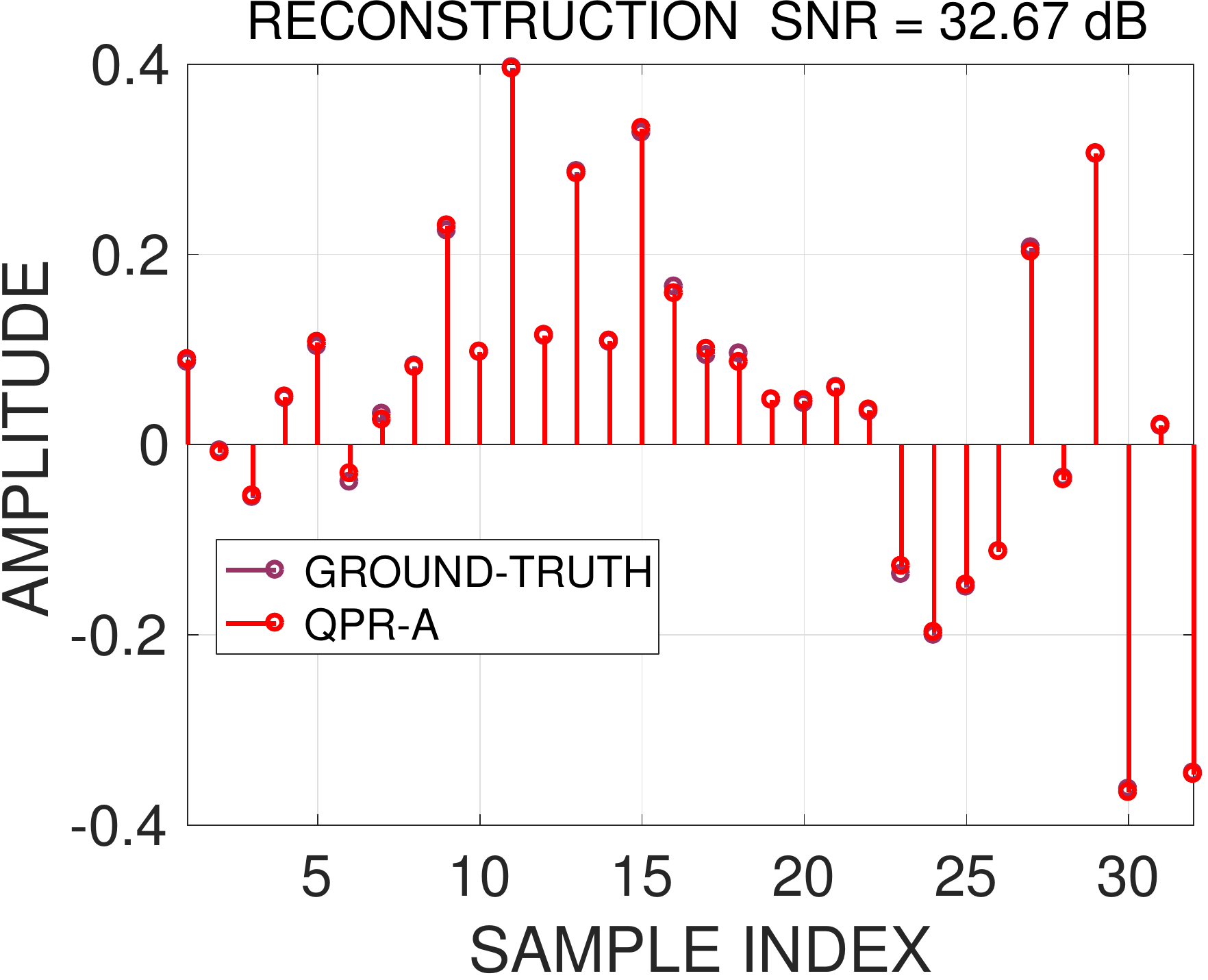}}
\subfigure[PL-A reconstruction]{
\includegraphics[width=1.5in]{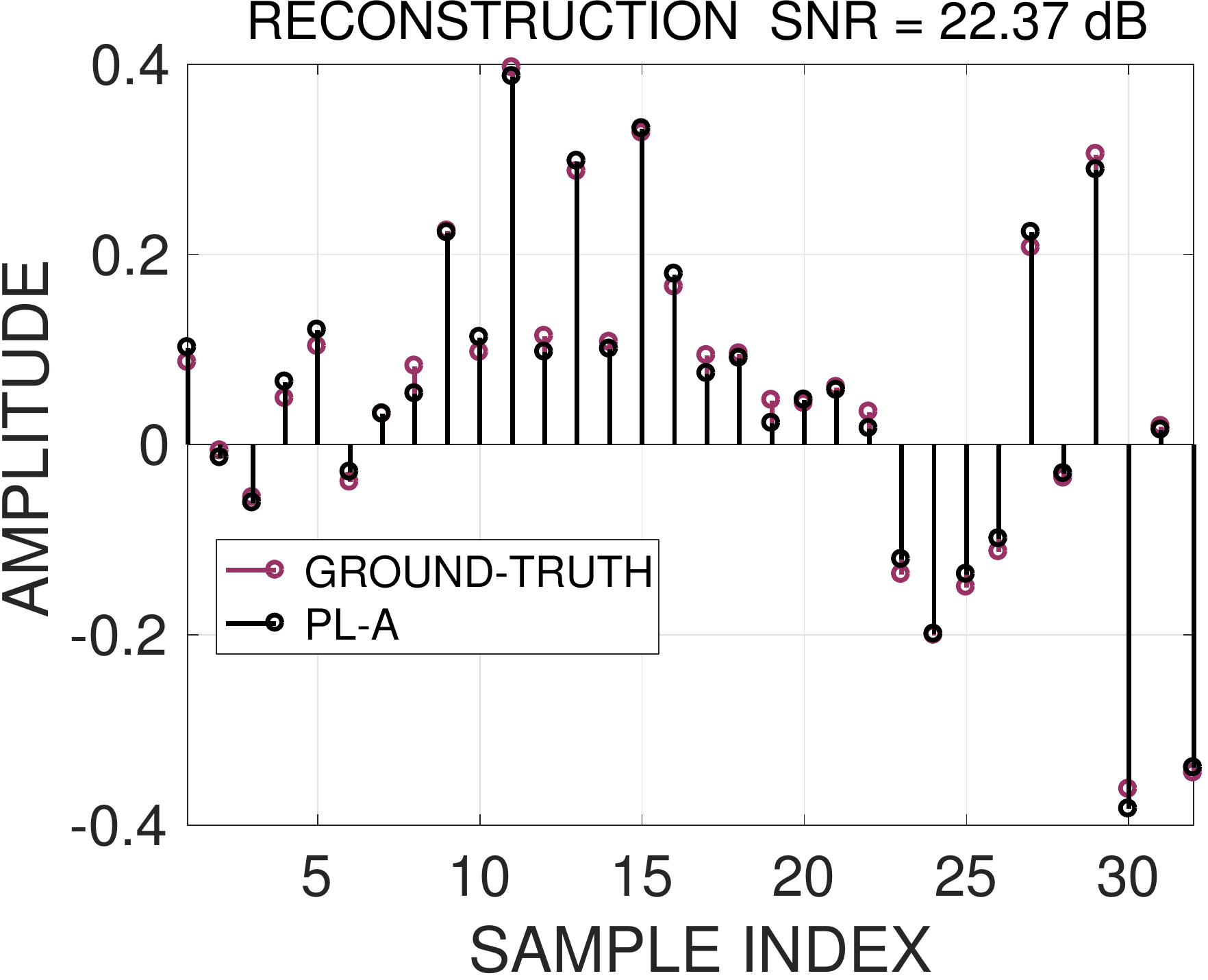}}
\caption{\small{The effect of acceleration: (a) QPR-A versus QPR; and (b) PL-A versus PL. The reconstructed signals obtained using QPR-A and PL-A in a typical random trial are compared against the ground-truth in (c) and (d), respectively. The corresponding output SNRs are also indicated. }}
\label{QPR_vs_QPRA_figure}
\end{figure}

\indent Since the rank-1 projection operator $\mathcal{P}_{\text{rank}-1}$ in the update rule \eqref{pgd_algo} is not convex, it is not obvious a priori that incorporating the momentum factor, explained in Section \ref{APGD_steps_sec}, would necessarily lead to fast convergence or give any performance gains. A similar dilemma was encountered in the context of PhaseLift as well. Therefore, we compare the performances of QPR and QPR-A to determine which of them results in superior reconstruction, and also compare the results to PhaseLift with and without acceleration. We consider $m=10n$ measurements quantized using $k=8$ levels, EQ for QPR and QPR-A, and LMQ for PL and PL-A. These are the optimal quantizer settings for the respective algorithms as will be demonstrated in the following subsection.\\
\indent The average reconstruction SNRs and their standard deviations for QPR and QPR-A, calculated over $20$ independent trials, are shown in Figure \ref{QPR_vs_QPRA_figure}(a) with respect to iterations. The same metrics for PL and PL-A are shown in Figure \ref{QPR_vs_QPRA_figure}(b). We observe from Figure \ref{QPR_vs_QPRA_figure}(a) that QPR-A results in an improvement of approximately $8$ dB over QPR after $100$ iterations. The variation of the reconstruction SNR around its average value is also found to be slightly smaller for QPR-A. On the other hand, PL-A leads to faster convergence than PL, as can be inferred from Figure \ref{QPR_vs_QPRA_figure}(b), although the final SNR after $100$ iterations settles to more or less the same value for both of them. This trend was found to be consistent for different values of $m$ and $k$. The reconstructed signals obtained in a random trial using QPR-A and PL-A are compared against the ground-truth in Figures \ref{QPR_vs_QPRA_figure}(c) and \ref{QPR_vs_QPRA_figure}(d), respectively. QPR-A yields an improvement of approximately $10$ dB in the reconstruction SNR over PL-A. For further comparisons, we consider only QPR-A and PL-A owing to their superiority over QPR and PL, respectively.
%%%%%%%%%%%%%%%%%%%%%%%%%%%%%%%%%%%%%%%%%%%%%
%%%%%%%%%%%%%%%%%%%%%%%%%%%%%%%%%%%%%%%%%%%%%
%%%%%%%%%%%%%%%%%%%%%%%%%%%%%%%%%%%%%%%%%%%%%
\begin{figure}[t]
\centering
\subfigure[$k=8$, LMQ]{
\includegraphics[width=1.5in]{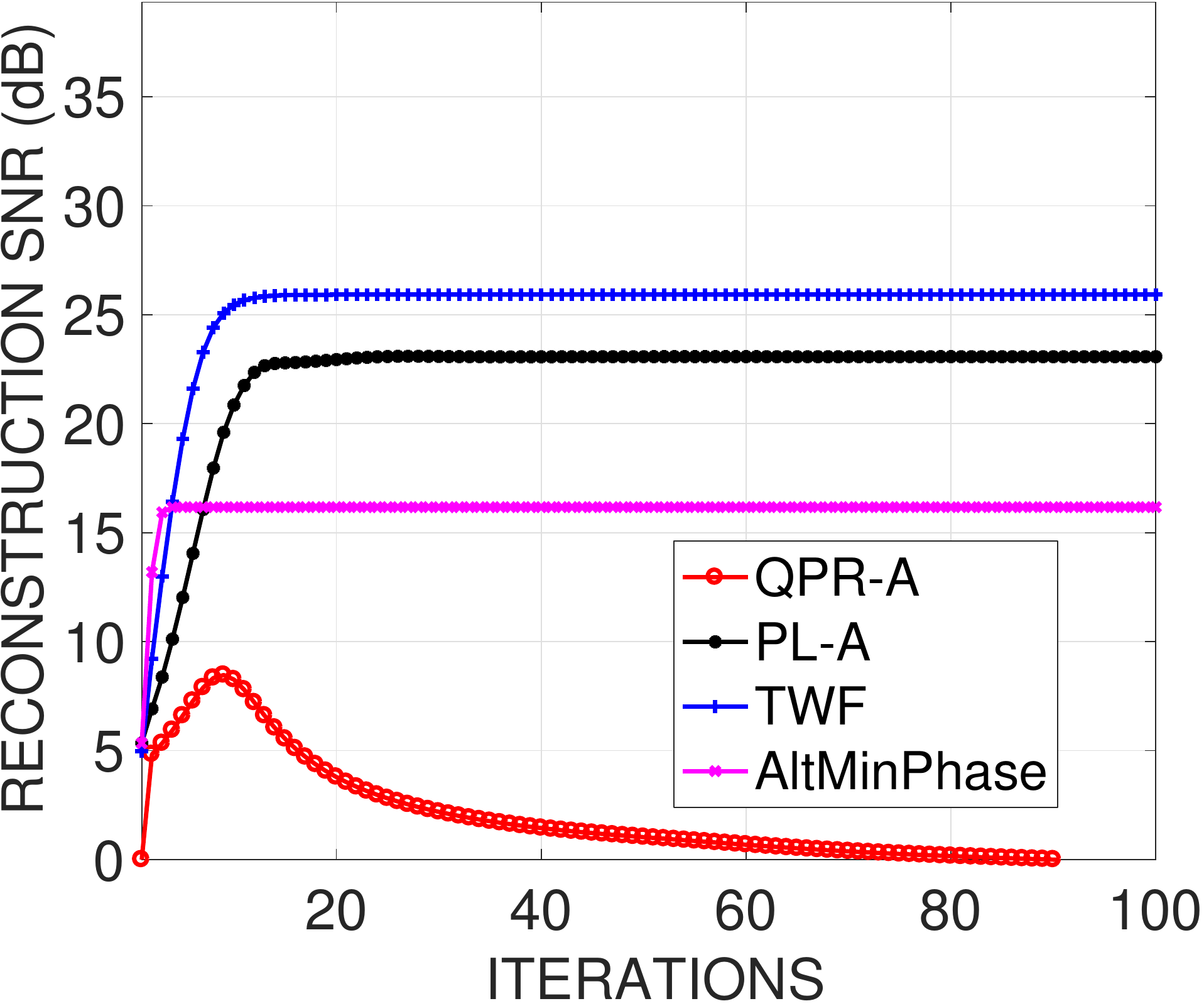}}
\subfigure[$k=8$, EQ]{
\includegraphics[width=1.5in]{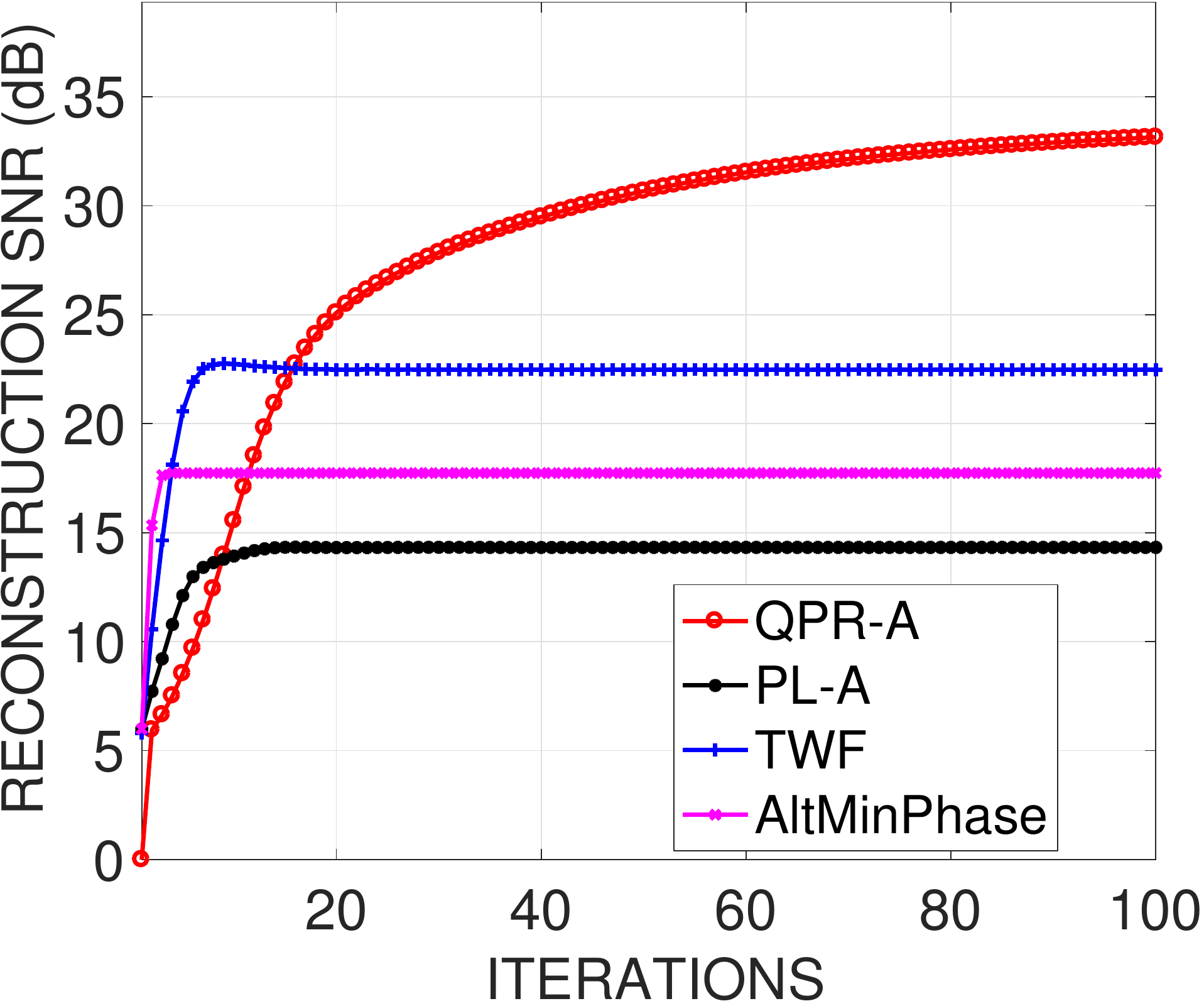}}\\
\subfigure[$k=16$, LMQ]{
\includegraphics[width=1.5in]{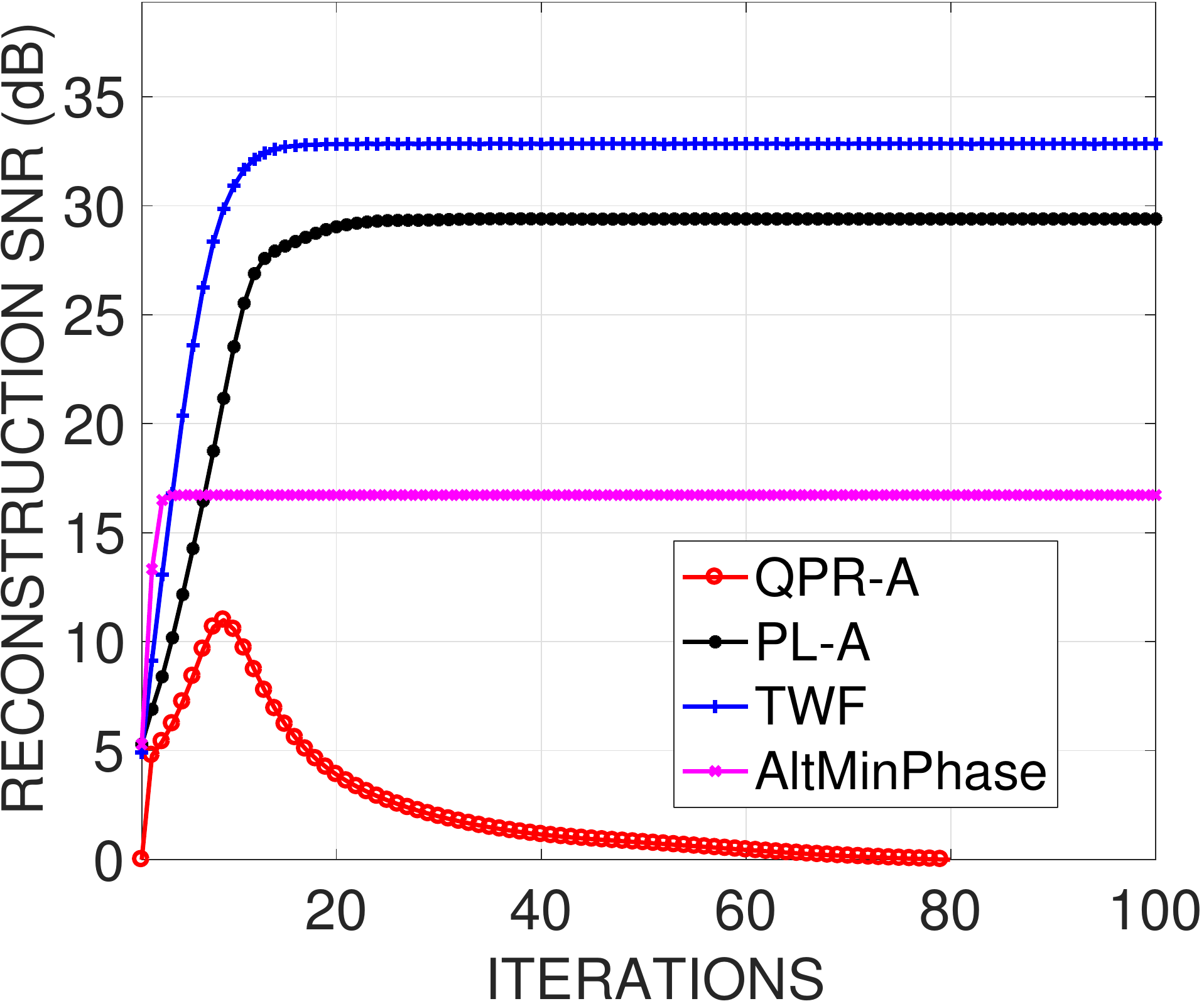}}
\subfigure[$k=16$, EQ]{
\includegraphics[width=1.5in]{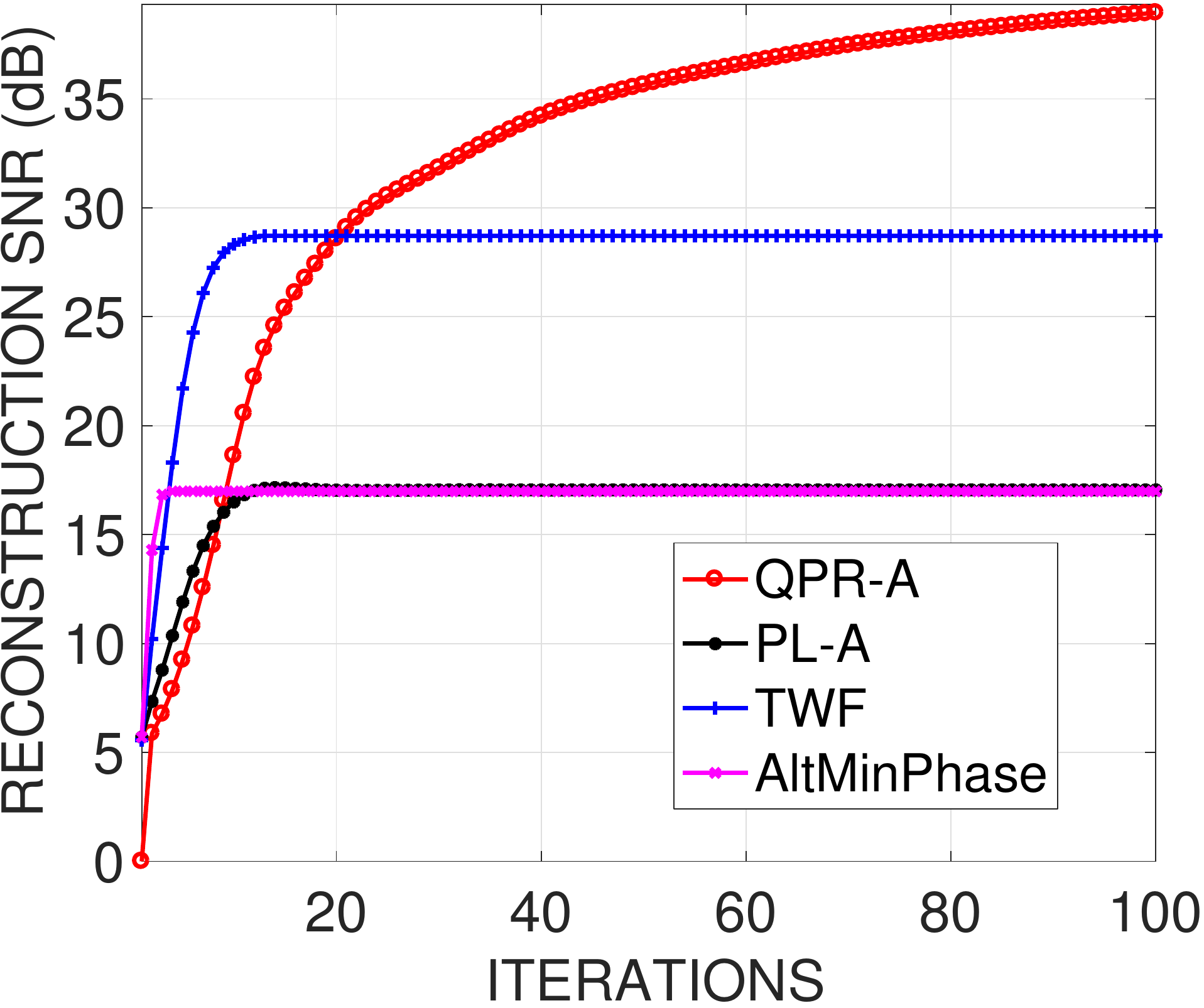}}
\caption{\small{Effect of quantizer: the output SNR versus iterations for different algorithms corresponding to the LMQ ((a) and (c)) and the EQ ((b) and (d)). Two different levels of quantization, namely, $k=8$ and $k=16$ are chosen for the experiment. }}
\label{compare_eq_with_opt_quant_figure}
\end{figure}
%%%%%%%%%%%%%%%%%%%%%%%%%%%%%%%%%%%%%%%%%%%%%
%%%%%%%%%%%%%%%%%%%%%%%%%%%%%%%%%%%%%%%%%%%%%
\begin{figure}[t]
\centering
\subfigure[$k=4$]{
\includegraphics[width=1.5in]{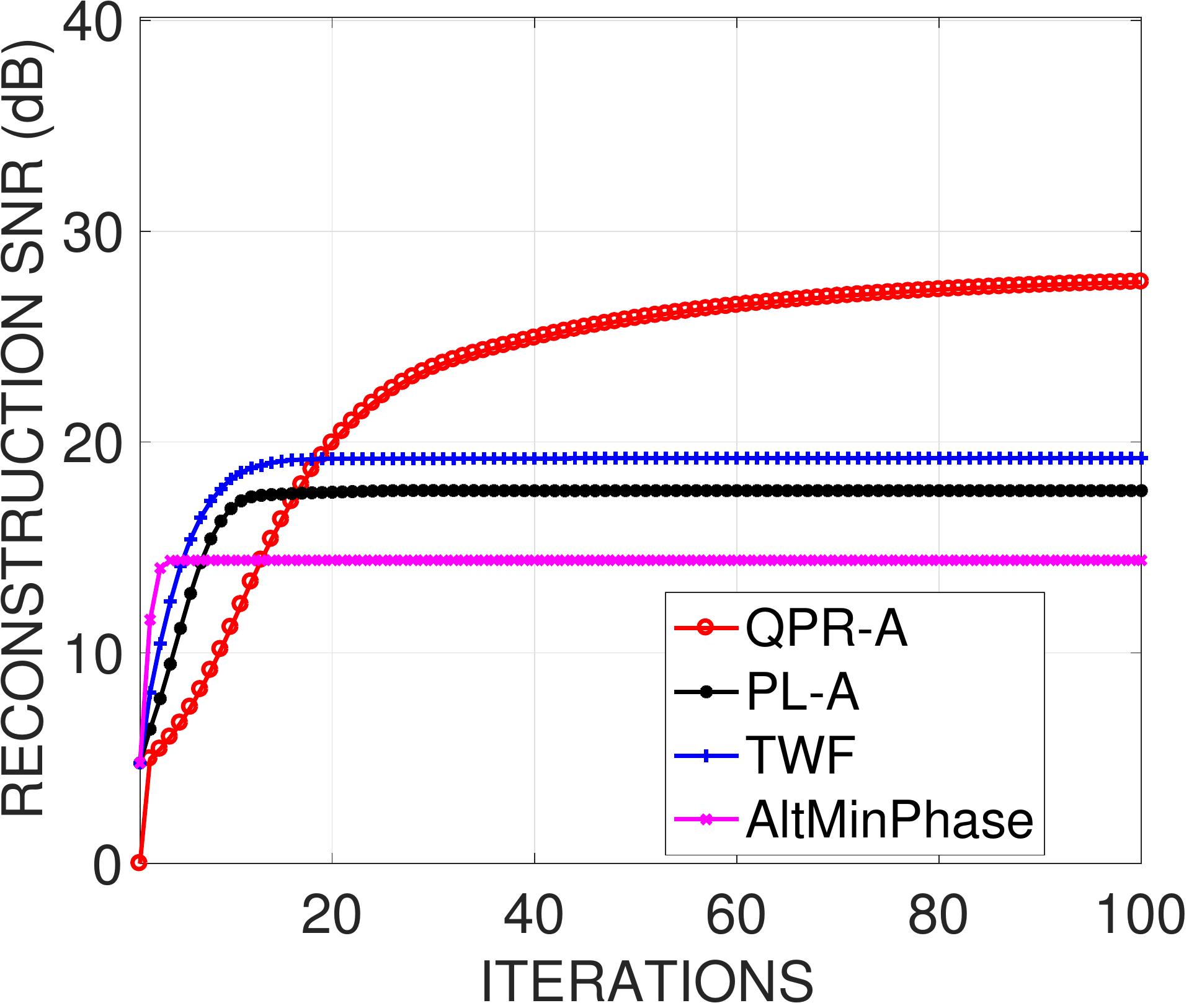}}
\subfigure[$k=8$]{
\includegraphics[width=1.5in]{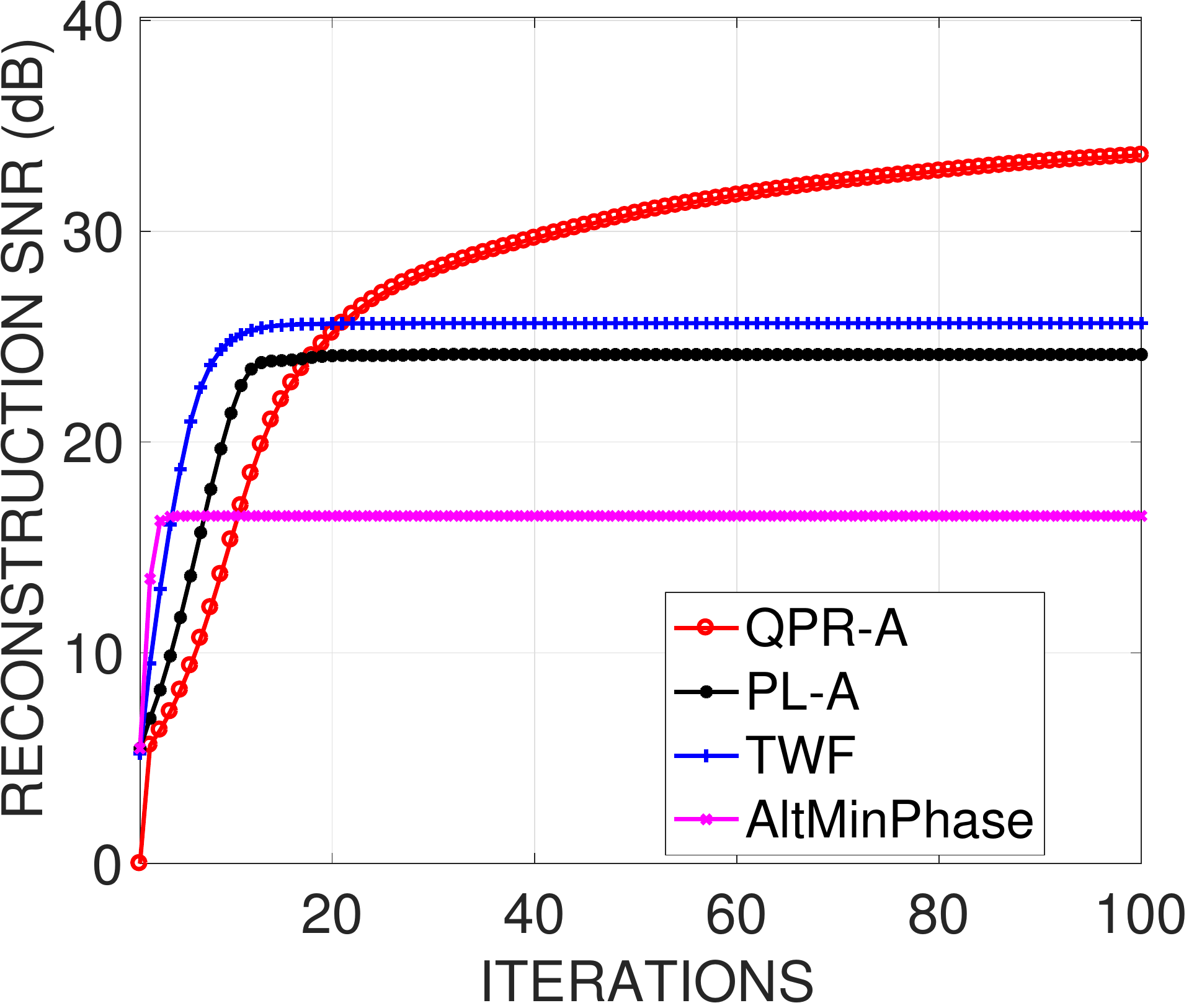}}\\
\subfigure[$k=12$]{
\includegraphics[width=1.5in]{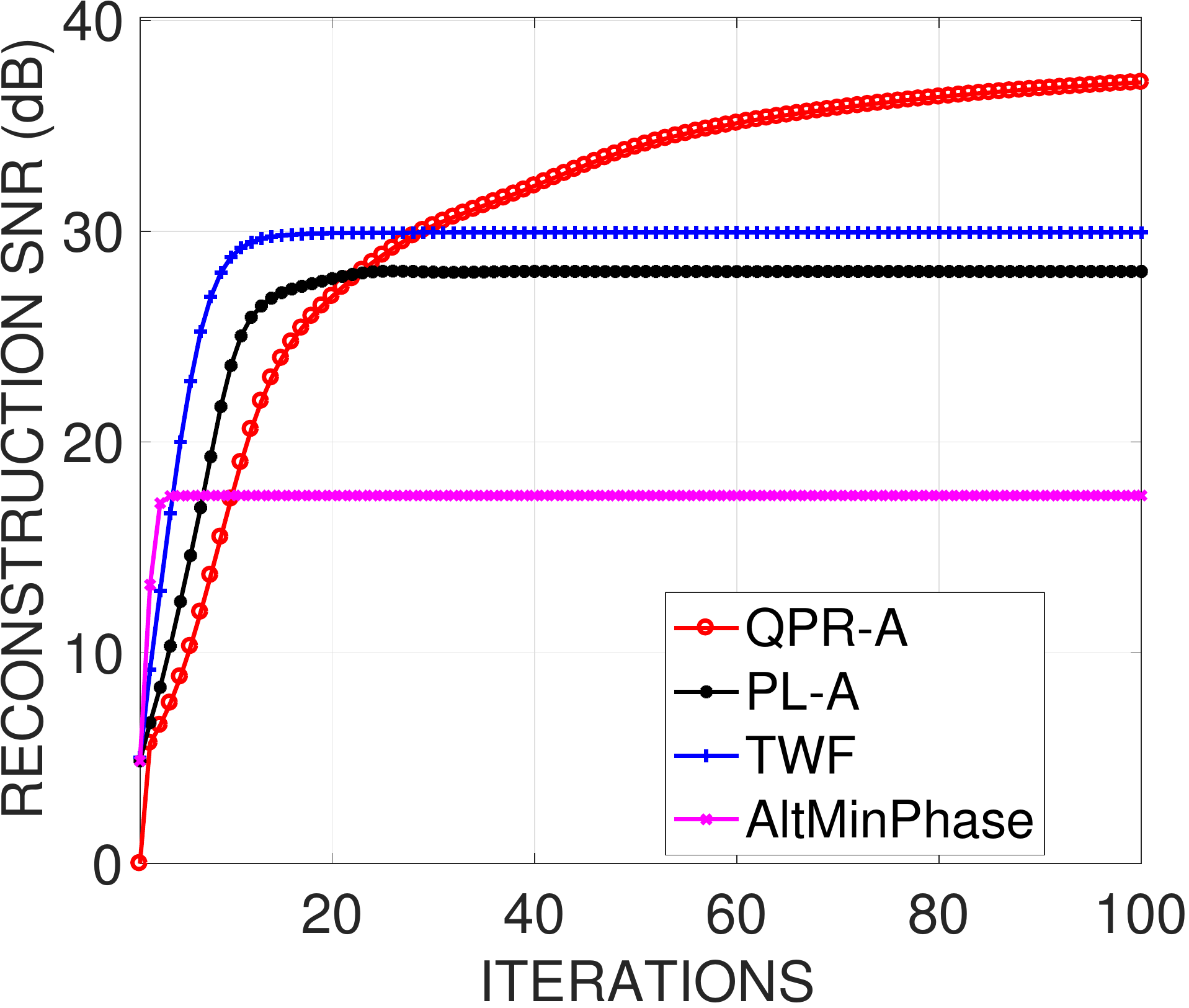}}
\subfigure[$k=16$]{
\includegraphics[width=1.5in]{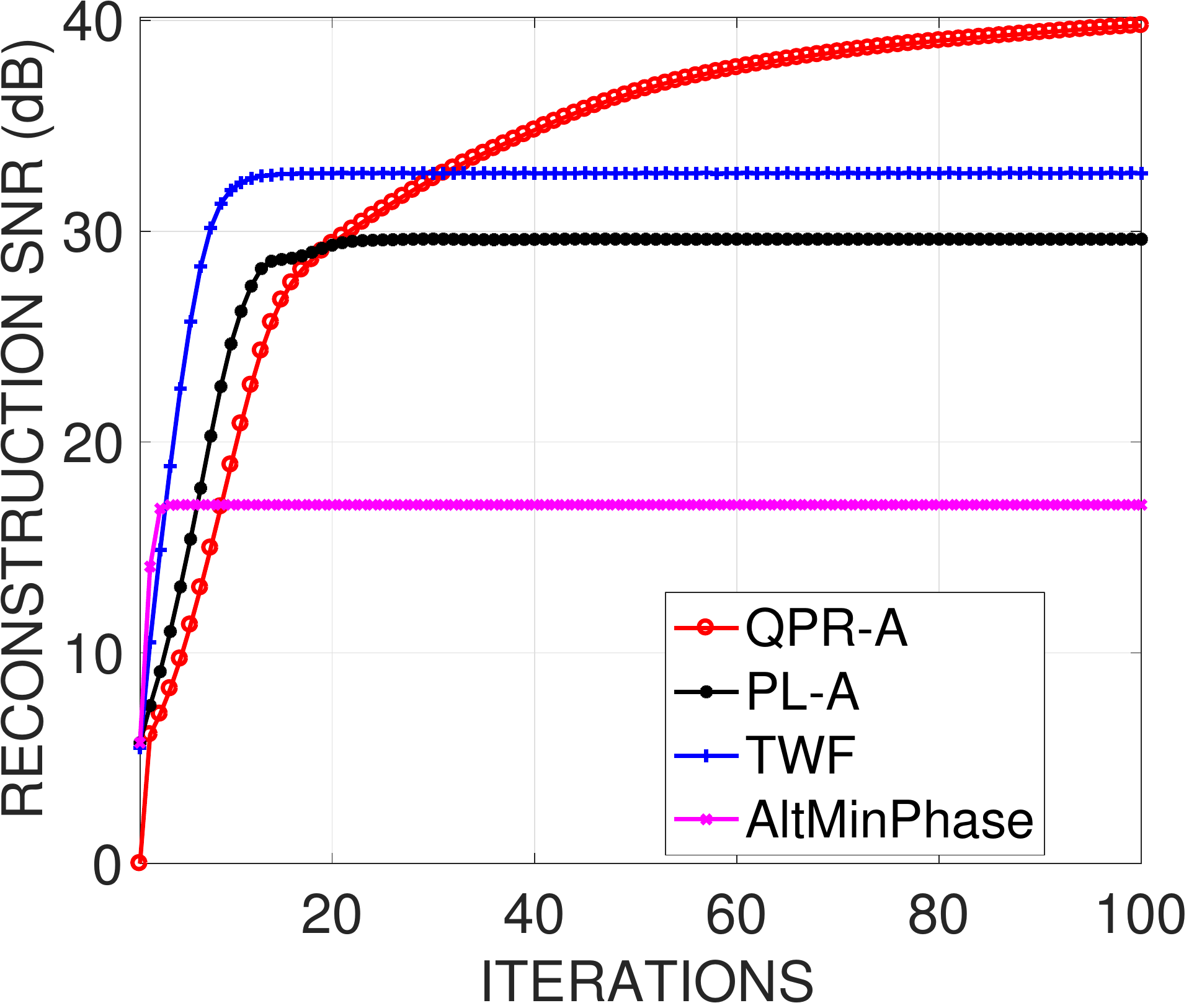}}
\caption{\small{Reconstruction SNR versus iterations for different algorithms corresponding to various quantization levels $k$. QPR-A leads to an improvement of approximately $5$ dB over the best performing PR technique, namely the TWF algorithm, not customized to tackle quantization. The output SNR improves with increasing $k$. }}
\label{QPR_comparison_SNR_figure}
\end{figure}
%%%%%%%%%%%%%%%%%%%%%%%%%%%%%%%%%%%%%%%%%%%%%
\subsection{Quantizer Design: Lloyd-Max Versus Equiprobable}
\label{opt_vs_eq_quant_sec}
\indent In LMQ, the thresholds and the encoding symbols are jointly optimized to maximize the quantization SNR. On the other hand, the thresholds in EQ are chosen such that each interval has equal probability. The encoding symbols in EQ are taken as $s_j=\frac{\tau_{j-1}+\tau_j}{2}$, for $j=1,2,\cdots,k-1$, and $s_k=\tau_{k-1}+2\delta$, where $\delta$ is as defined in \eqref{delta_def_eq}. Since the consistency criterion is enforced in QPR-A, the specific choice of the codebook has no bearing on the performance of QPR algorithms, as explained in Section \ref{quant_design_sec}. In other words, the estimated signals obtained using QPR-A corresponding to two different quantizers having the same intervals, but different codewords, would be the same. The variations of reconstruction SNR versus iterations for $k=8$ and $k=16$, averaged over $20$ random trials, are shown in Figure \ref{compare_eq_with_opt_quant_figure}. Comparing Figures~\ref{compare_eq_with_opt_quant_figure}(a) with \ref{compare_eq_with_opt_quant_figure}(b) and \ref{compare_eq_with_opt_quant_figure}(c) with \ref{compare_eq_with_opt_quant_figure}(d), we observe that the performances of PL-A and TWF improve significantly when LMQ is used for measurement quantization; whereas the performance of AltMinPhase remains approximately the same under LMQ and EQ. The reconstruction SNR of QPR-A corresponding to LMQ initially increases with iterations, but drops as the number of iterations exceeds $10$. However, when the EQ is used for measurement quantization, we observe that QPR-A leads to a steady increase of the reconstruction SNR as the iterations progress. The experiment indicates that the EQ is a better choice than LMQ for QPR-A, whereas for the competing algorithms, the LMQ is better. The superior performance of the competing algorithms with LMQ  is not too surprising since they are not quantization-aware. Any quantization noise would only be treated as additive noise and their performance would be the best when the quantization noise variance is the least, which is what the LMQ guarantees. Therefore, in order to facilitate a fair comparison, in the sequel, we report the performance of QPR-A with the EQ and its competitors with the LMQ.

\subsection{Comparison of QPR-A With the State-of-the-Art}
\label{compare_QPRA_with_state_of_the_art_sec}
\begin{figure}[t]
\centering
\subfigure[$k=4$]{
\includegraphics[width=1.5in]{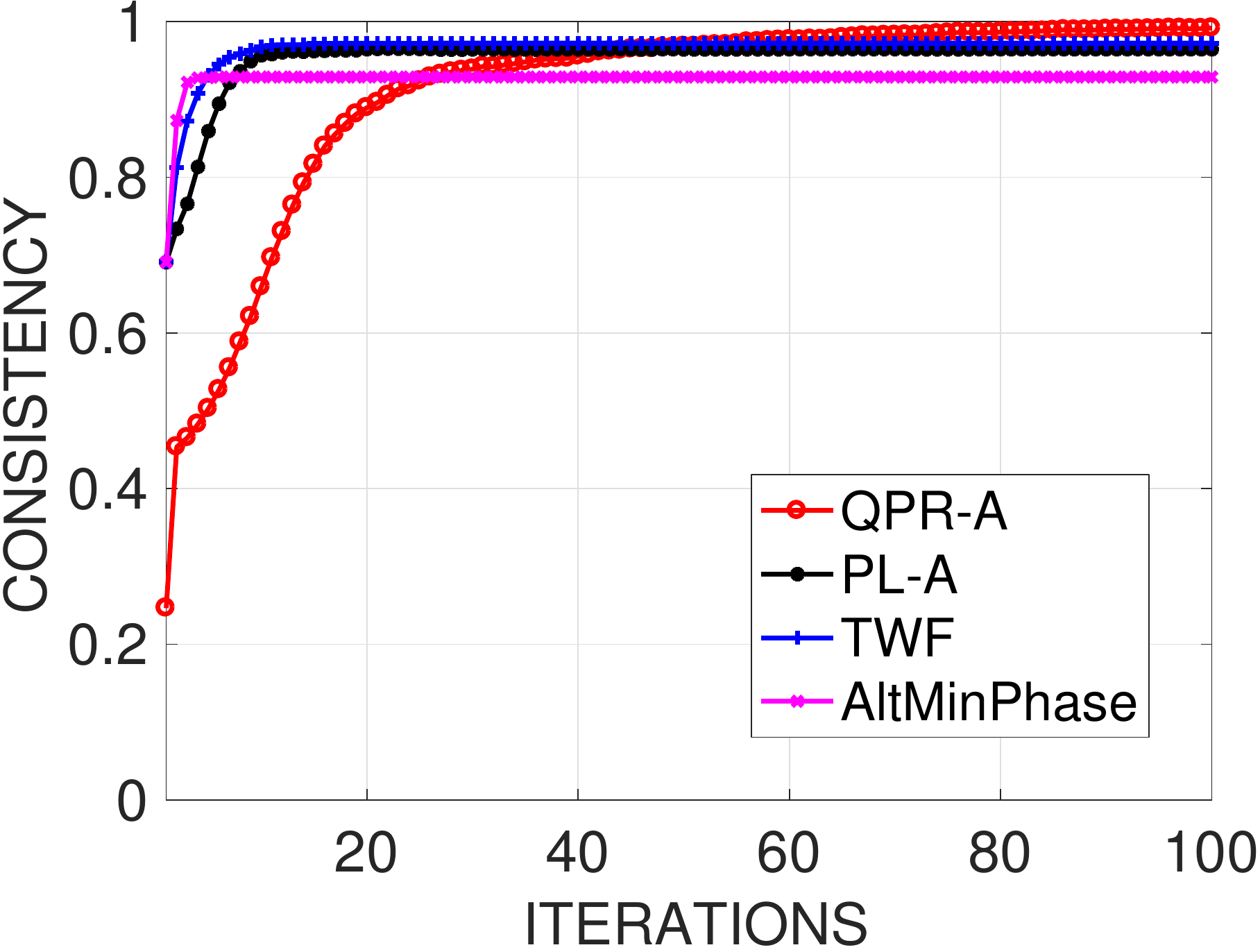}}
\subfigure[$k=8$]{
\includegraphics[width=1.5in]{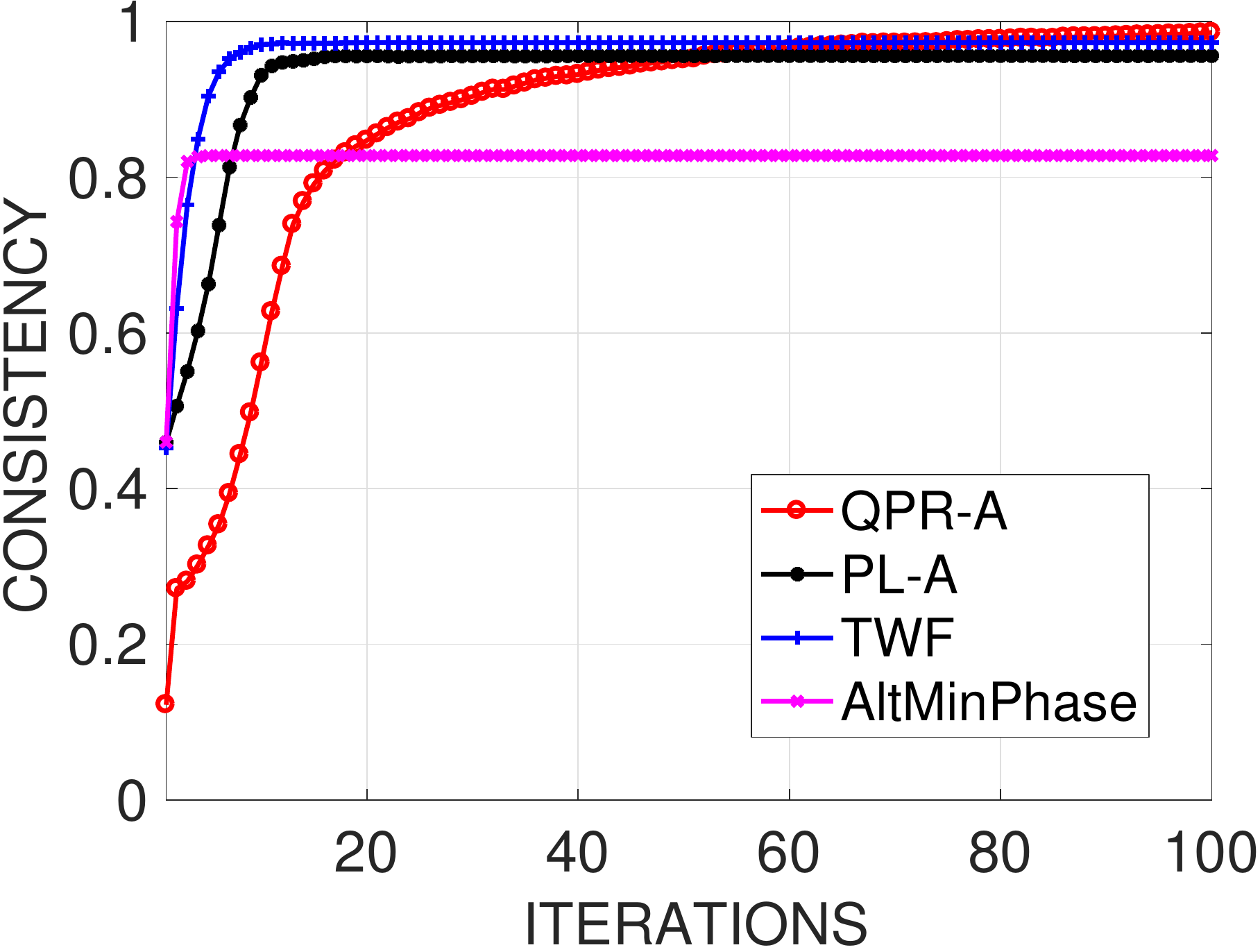}}\\
\subfigure[$k=12$]{
\includegraphics[width=1.5in]{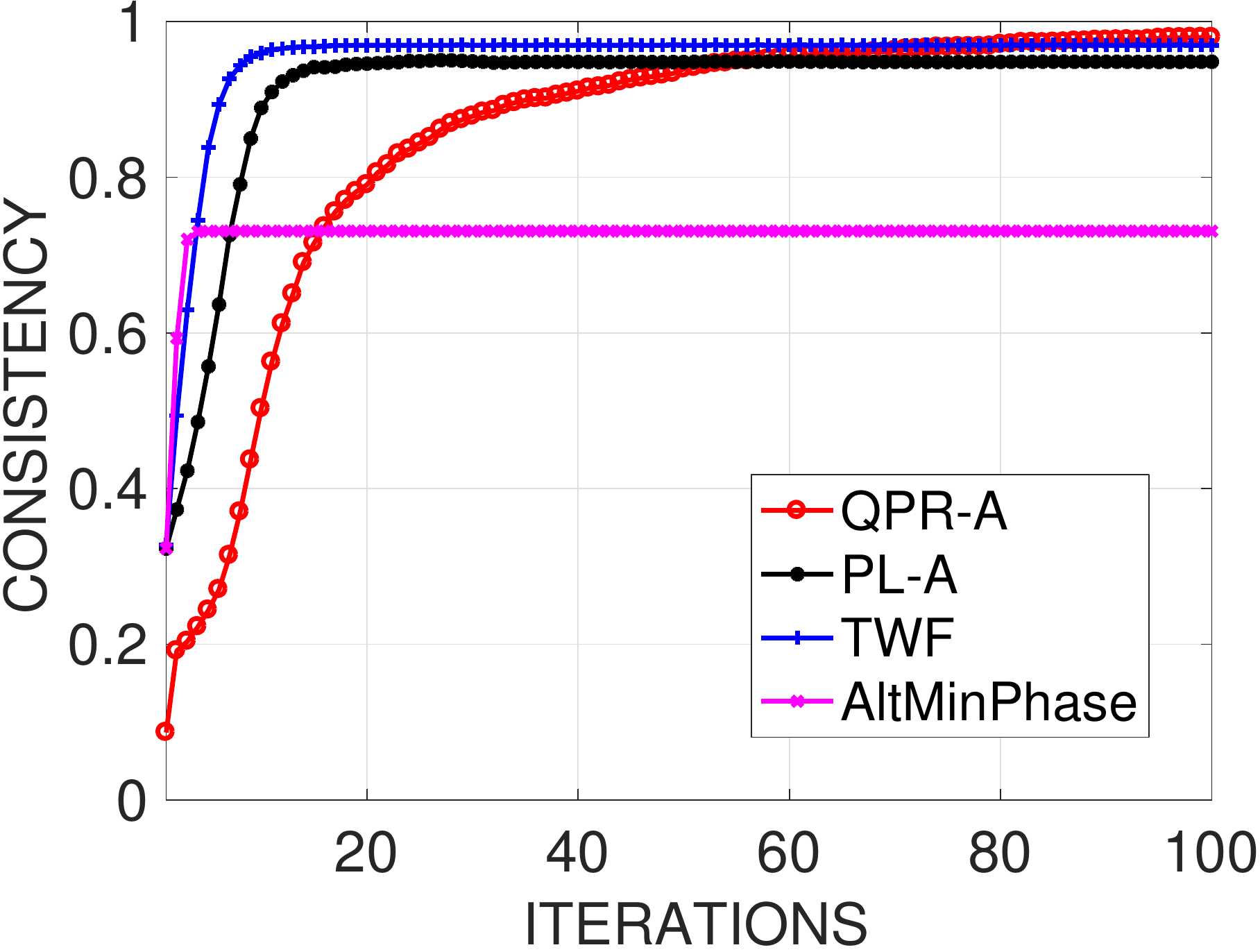}}
\subfigure[$k=16$]{
\includegraphics[width=1.5in]{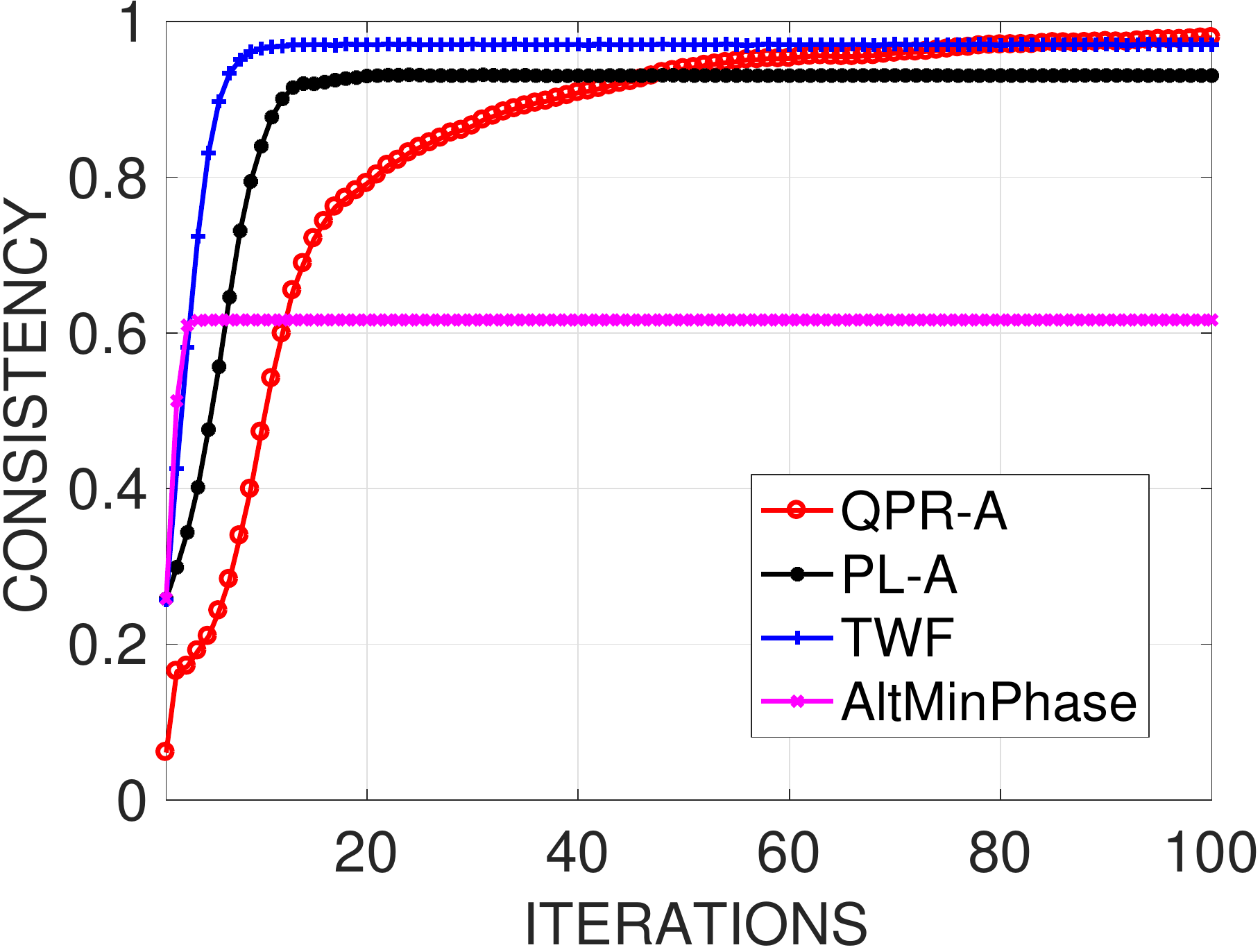}}
\caption{\small{Measurement consistency $c_m$ versus iterations for different algorithms corresponding to various quantization levels $k$. QPR-A eventually explains the measurements reasonably well despite an inferior initialization. }}
\label{QPR_comparison_consistency_figure}
\end{figure}

\indent  The comparative performances of QPR-A and the competing algorithms in terms of the reconstruction SNR and measurement consistency are shown in Figures \ref{QPR_comparison_SNR_figure} and \ref{QPR_comparison_consistency_figure}, respectively, corresponding to various quantization levels $k$. We observe from Figure \ref{QPR_comparison_SNR_figure} that QPR-A ultimately results in superior reconstruction SNR  despite having a suboptimal initialization, and this trend is consistent for all values of $k$. We observe from Figure \ref{QPR_comparison_SNR_figure} that the improvement in reconstruction SNR obtained using QPR-A over TWF, the best performing competing technique, is approximately $8$ dB for $k=4$, and reduces to nearly $6$ dB as $k$ increases to $16$. Moreover, unlike the competing techniques, the reconstruction SNR of QPR-A does not seem to saturate fast. Naturally, the reconstruction SNR increases with $k$ for all algorithms because the quantization gets finer. As far as measurement consistency is concerned, we observe from Figure \ref{QPR_comparison_consistency_figure} that QPR-A steadily improves with iterations and eventually performs on par with its competitors. In other words, as the iterations progress, the reconstruction produced by QPR-A provides an accurate explanation of the acquired measurements.\\
%%%%%%%%%%%%%%%%%%%%%%%%%%%%%%%%%%%%%%%%%%%%%
\indent To summarize, the accelerated algorithms QPR-A and PL-A offer superior reconstruction performance than QPR and PL, respectively. An EQ is a better choice for QPR-A and LMQ for the competing algorithms. In the absence of noise, the QPR-A technique is at least $6$ dB better than the best performing PR technique, namely, the TWF. Before concluding this section, we show an application of QPR-A to the reconstruction problem in \textit{frequency-domain optical coherence tomography} (FDOCT). An example of natural image reconstruction using QPR-A is shown in Section~\ref{qpr_image_recon_sec} of the supplementary document.
%%%%%%%%%%%%%%%%%%%%%%%%%%%%%%%%%%%%%%%%%%%%%
\begin{figure}[t]
\subfigure[]{
\includegraphics[width=1.5in]{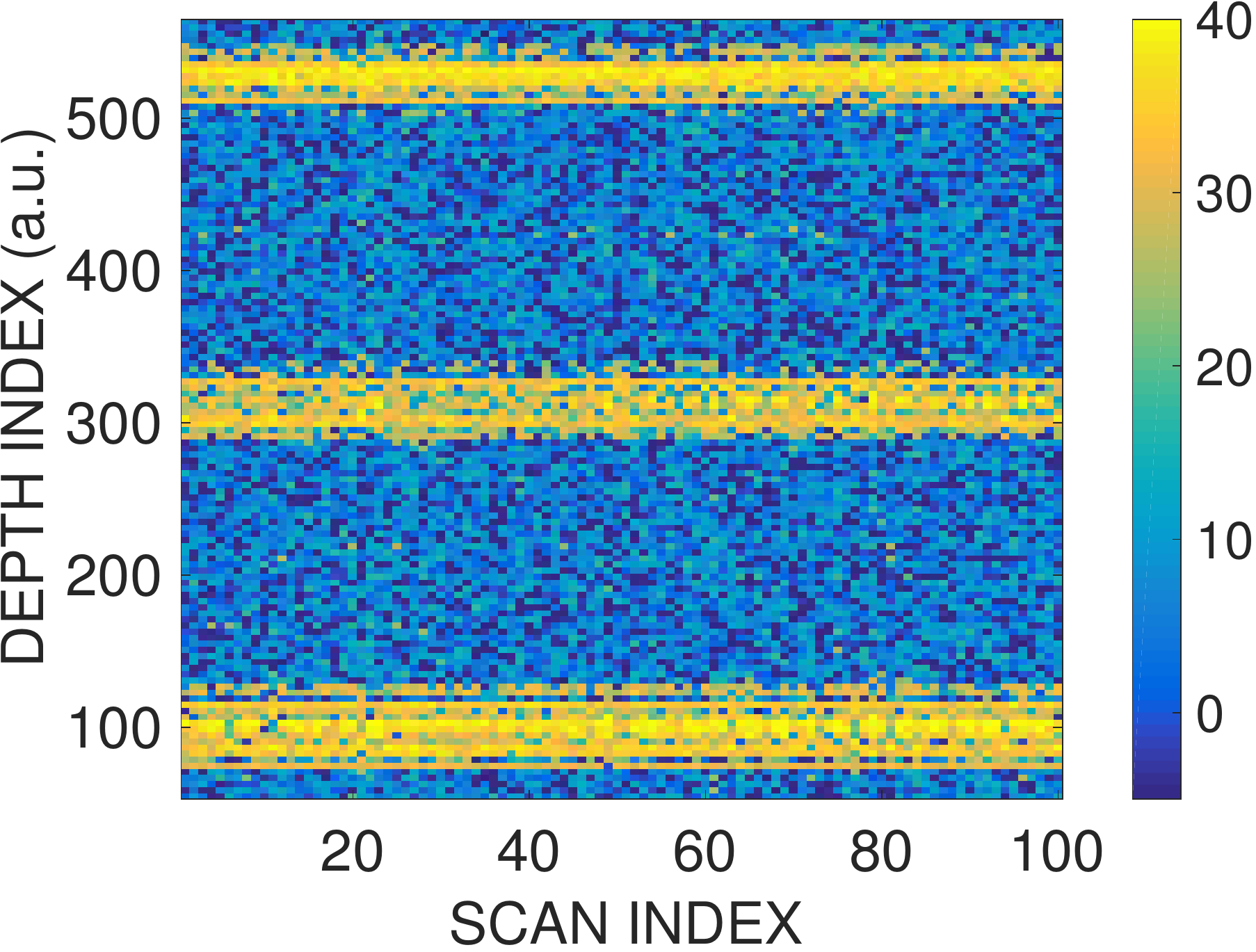}}
\subfigure[]{
\includegraphics[width=1.5in]{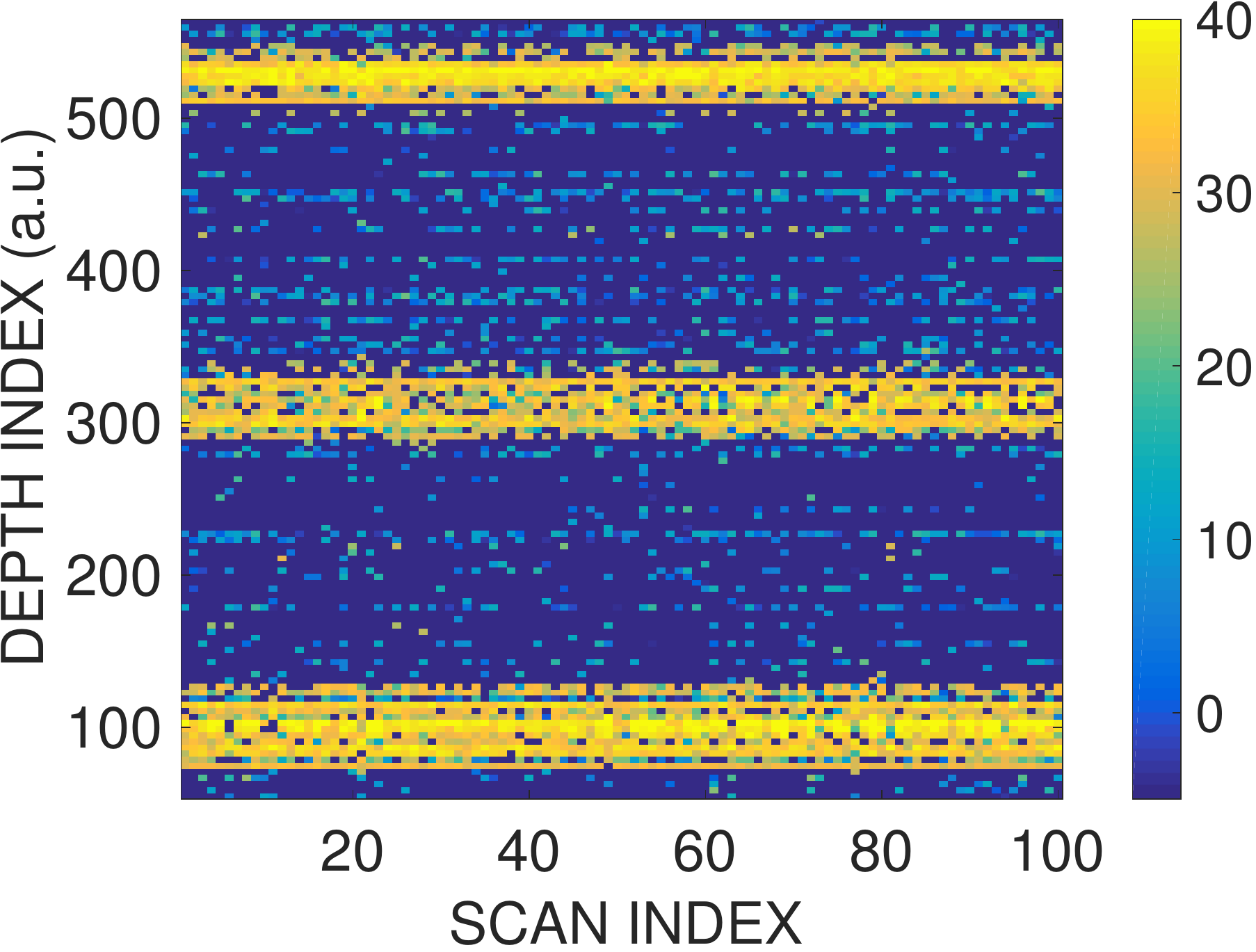}}\\
\subfigure[]{
\includegraphics[width=1.5in]{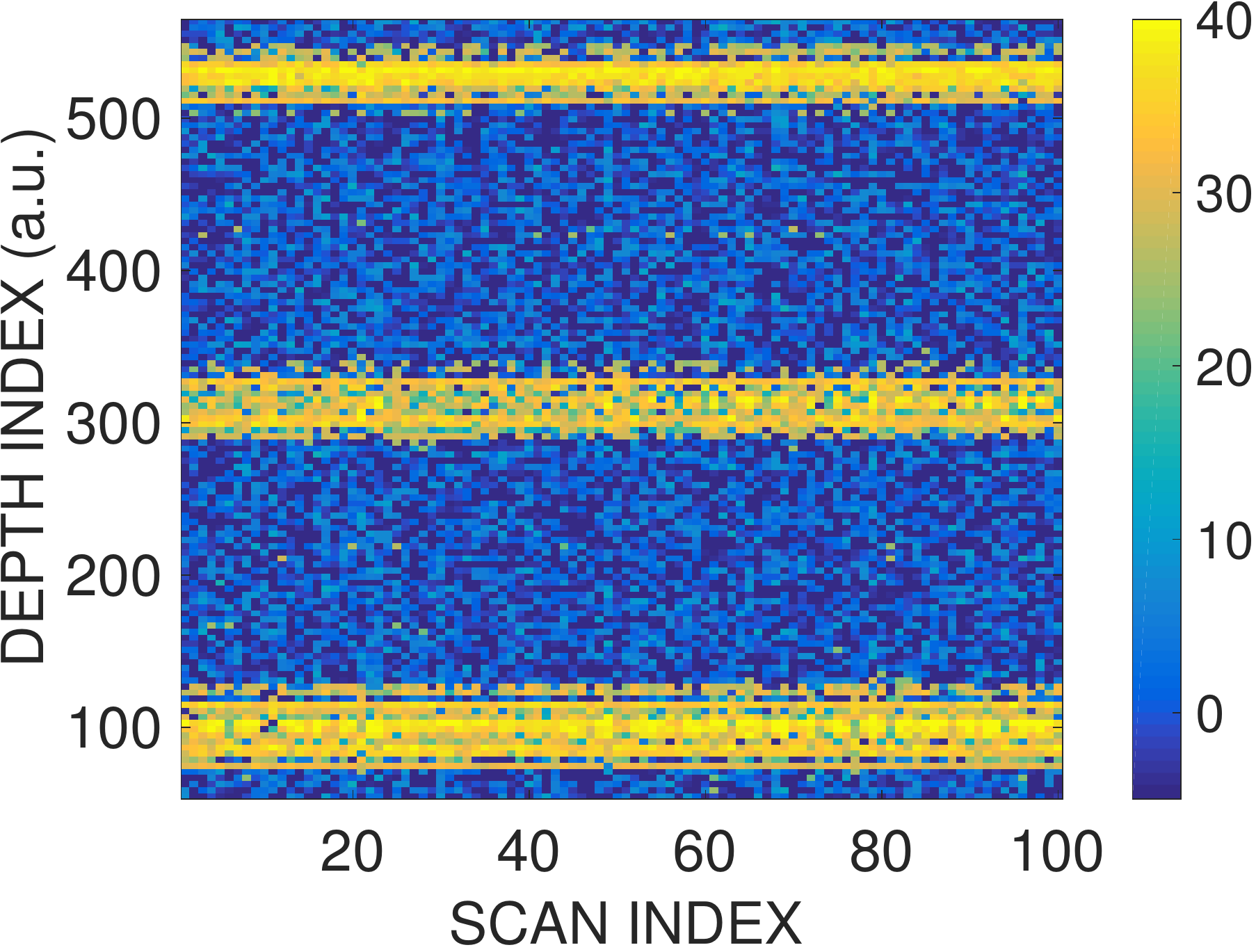}}
\subfigure[]{
\includegraphics[width=1.5in]{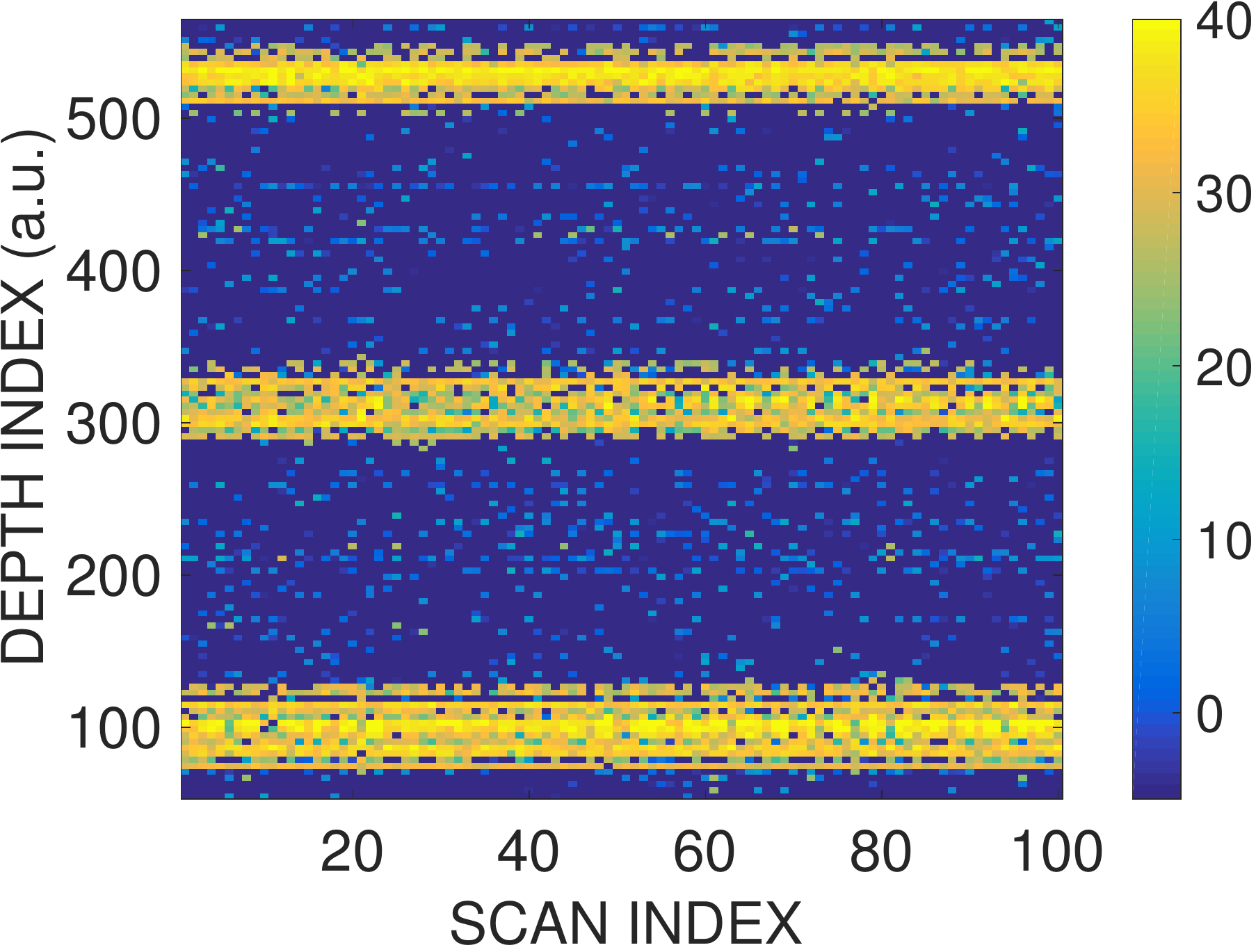}}\\
\subfigure[]{
\includegraphics[width=1.5in]{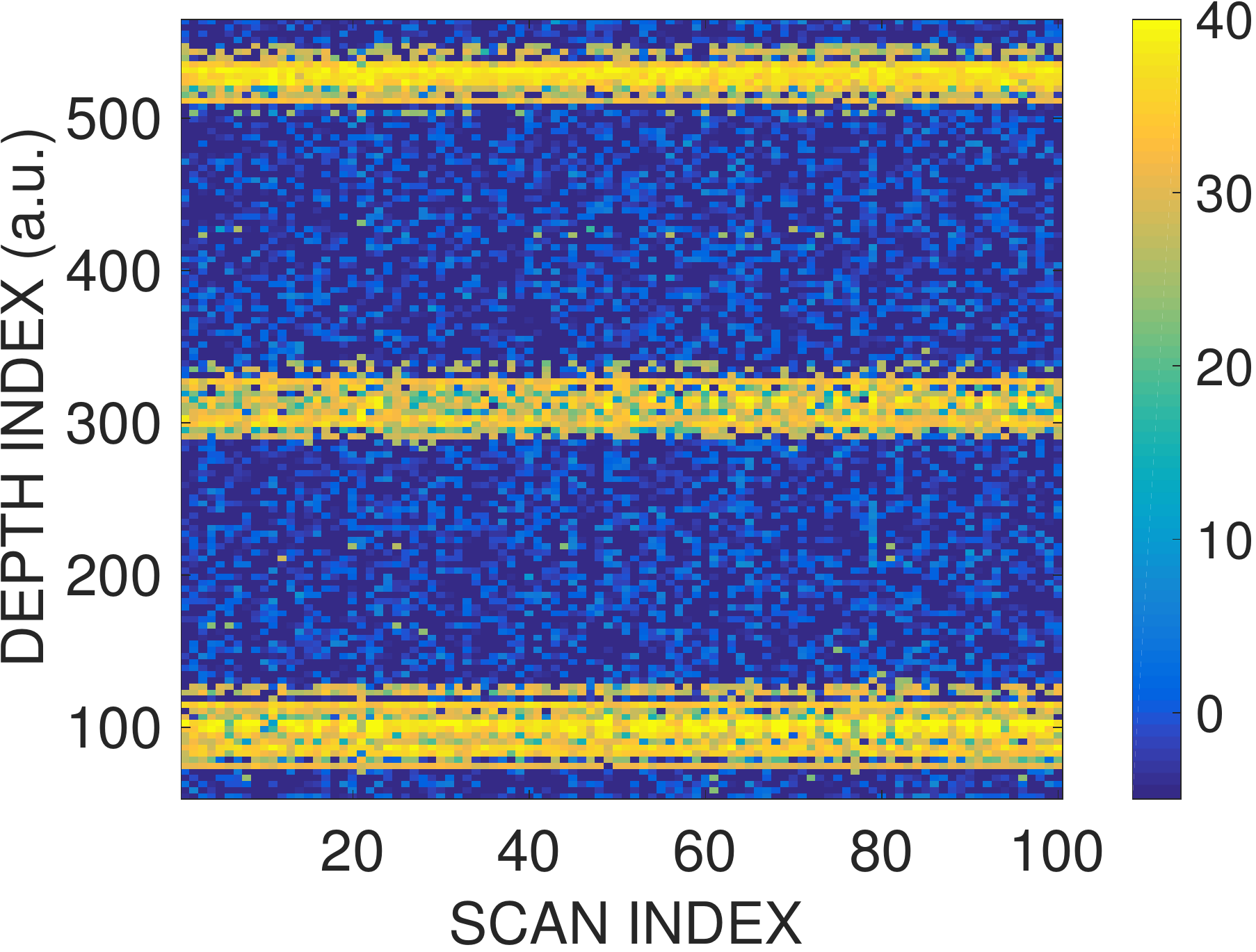}}
\subfigure[]{
\includegraphics[width=1.5in]{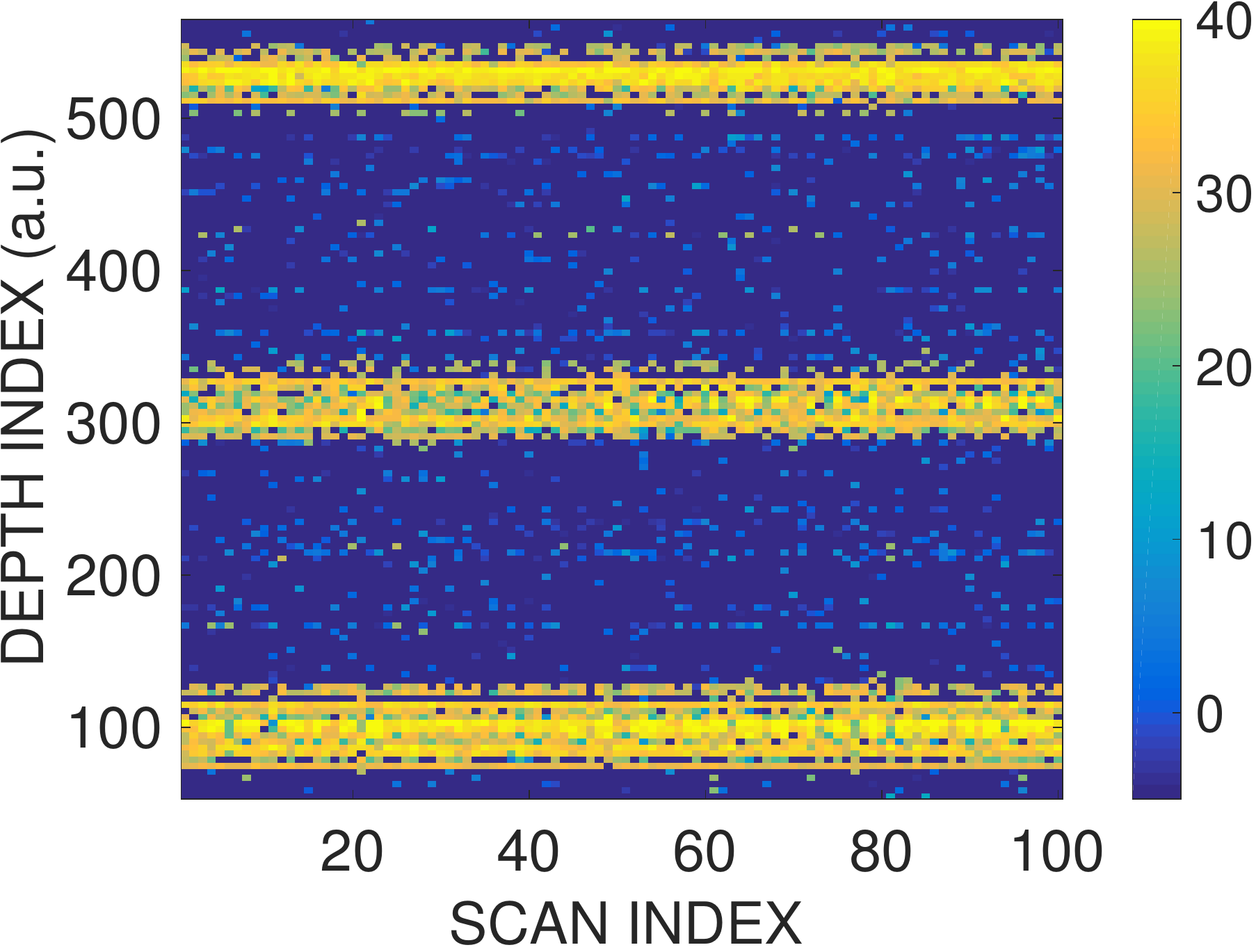}}
\caption{\small{Performance of QPR-A and SQPR-A corresponding to $k=4$ (first row), $k=8$ (second row), and $k=16$ (third row) on FDOCT image reconstruction. For each scan-line, we consider measurements down-sampled by a factor of $4$ over the region of interest, leading to a signal dimension of $n=128$. }}
\label{QPR_fdoct_figure}
\end{figure}
%%%%%%%%%%%%%%%%%%%%%%%%%%%%%%%%%%%%%%%%%%%%%
\subsection{Application of QPR-A to Frequency-Domain Optical Coherence Tomography (FDOCT)}
We consider signal reconstruction in FDOCT, a non-invasive imaging technique used for obtaining structural details of biological specimens.  A detailed description of the acquisition setup and the signal model in FDOCT that is relevant to the present discussion can be found in \cite{css_oct_2008}. The interference pattern formed by the reflected signals from the object and reference arms approximates the Fourier transform of the object wave and is recorded by the spectrometer. The key challenge is to reconstruct the reflected wave from the object arm, which carries structural information about the specimen, from the intensity recordings of the spectrometer. Since the reflected wave exhibits a strong peak only when there is a significant change of refractive index in the specimen, the assumption of sparsity is appropriate in this context. However, the QPR-A algorithm leads to a fairly accurate reconstruction of the tomograms even without the sparsity assumption, as we shall show next.\\
\indent FDOCT reconstruction of the glass specimen\footnote{The FDOCT data is the courtesy of Prof. R. A. Leitgeb, Medical University of Vienna, Austria.} produced by the QPR-A algorithm is shown in Figure~\ref{QPR_fdoct_figure} corresponding to three different quantization levels, namely $k=4$, $k=8$, and $k=16$. Considering the max-$s$ reconstruction \cite{css_sparsePR} as the ground-truth, $m=5n$ finite-precision measurements of the form \eqref{QPR_measurement_eq} are collected corresponding to every scan-line. We consider a downsampled version (by a factor of four) of the back-scattered wave along each scan-line, leading to a signal dimension of $n=128$, for the purpose of illustration. We observe that the QPR-A algorithm can reconstruct the back-scattered wave and recover the structural details in the specimen reliably. Imposing the sparsity constraint iteratively (more on the effect of sparsity in Section~\ref{SQPR_results_sec}), with a sparsity level $s=0.35n$, helps eliminate the background noise significantly, as one can observe from Figures~\ref{QPR_fdoct_figure}(b), \ref{QPR_fdoct_figure}(d), and \ref{QPR_fdoct_figure}(f). These results also show that increasing the quantizer precision leads to a higher accuracy in tomogram reconstruction.
%%%%%%%%%%%%%%%%%%%%%%%%%%%%%%%%%%%%%%%%%%%%% 
\begin{figure*}[t]
\centering
\subfigure[$k=4$ (2-bit quantization)]{
\includegraphics[width=2.0in]{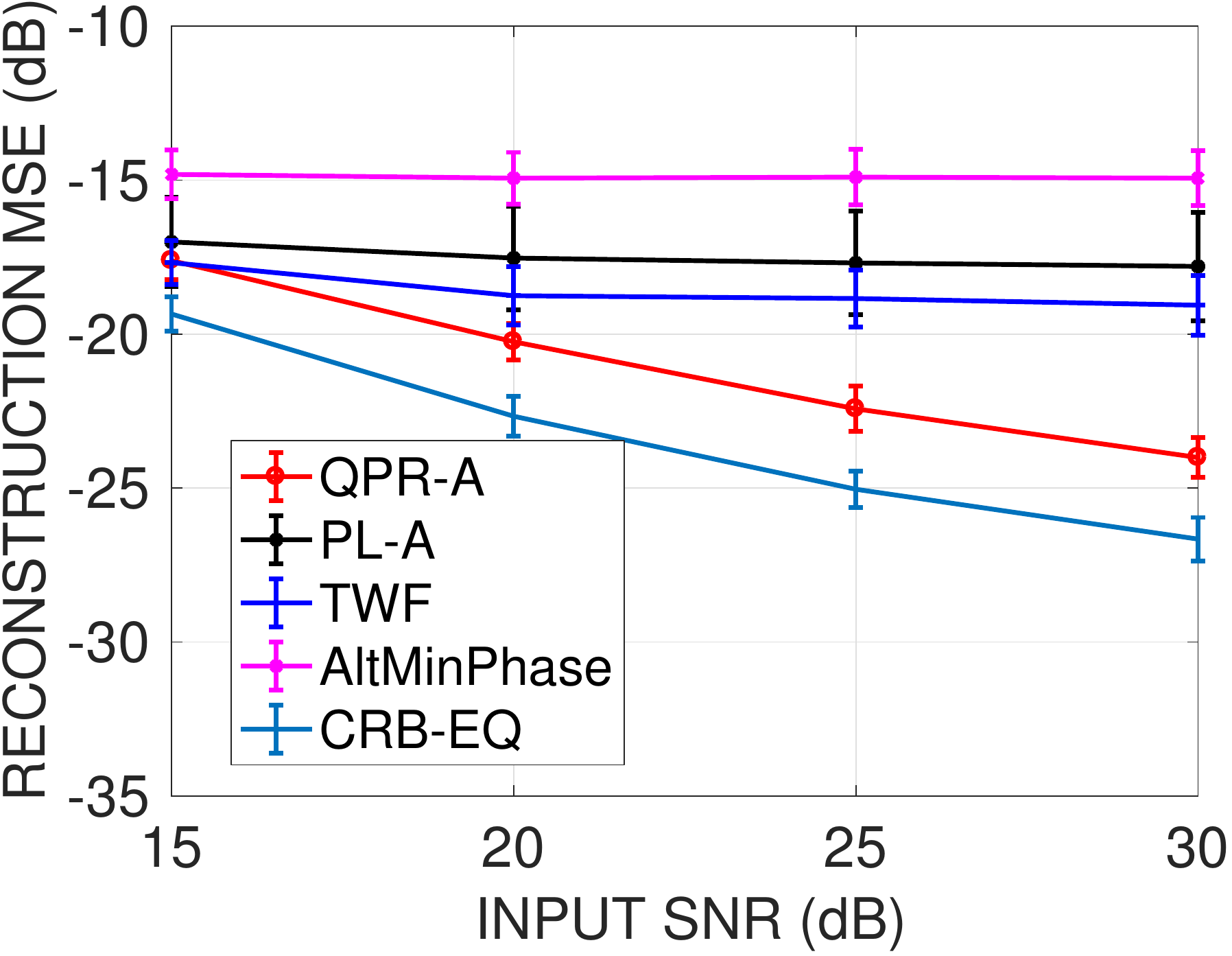}}
\subfigure[$k=8$ (3-bit quantization)]{
\includegraphics[width=2.0in]{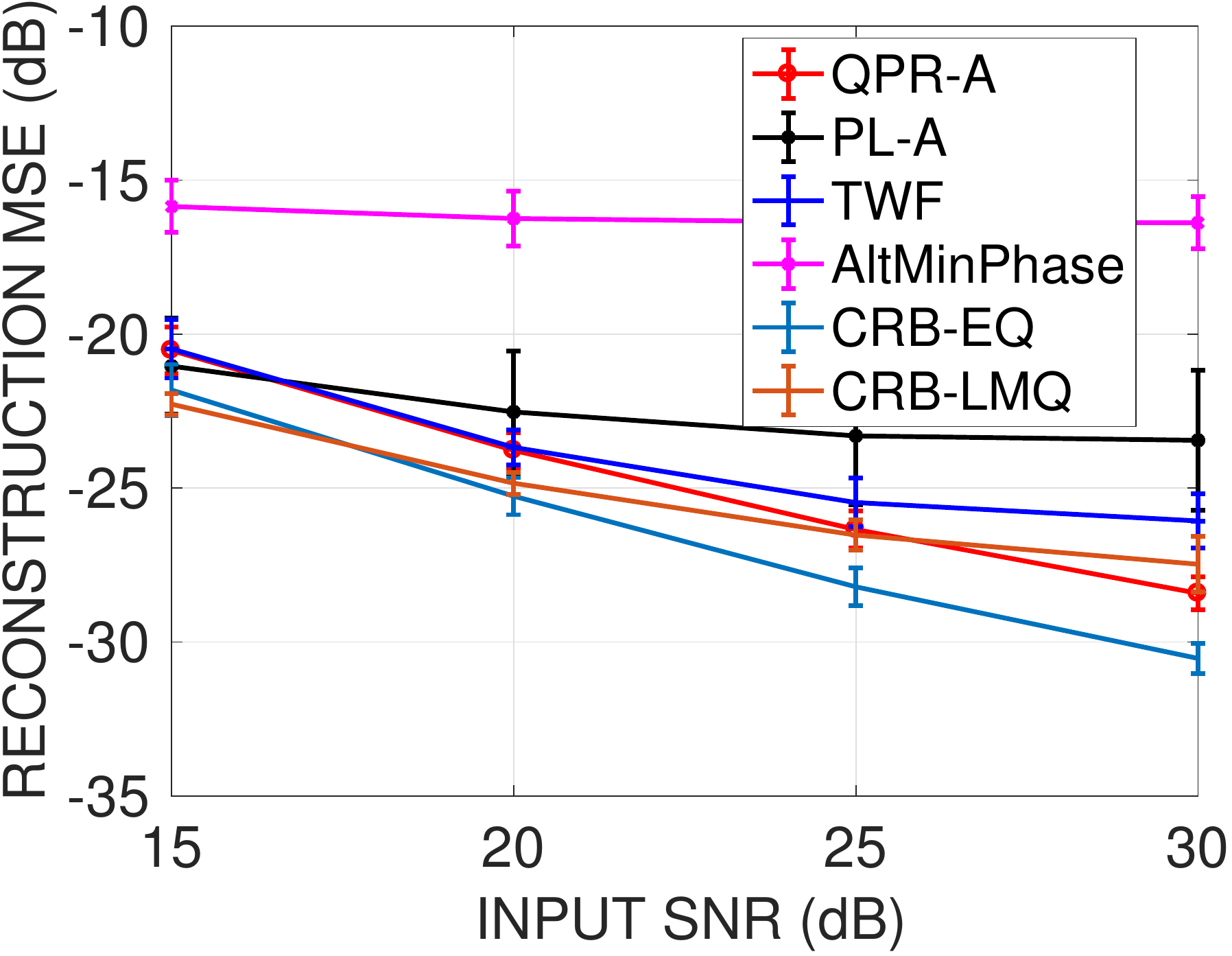}}
\subfigure[$k=16$ (4-bit quantization)]{
\includegraphics[width=2.0in]{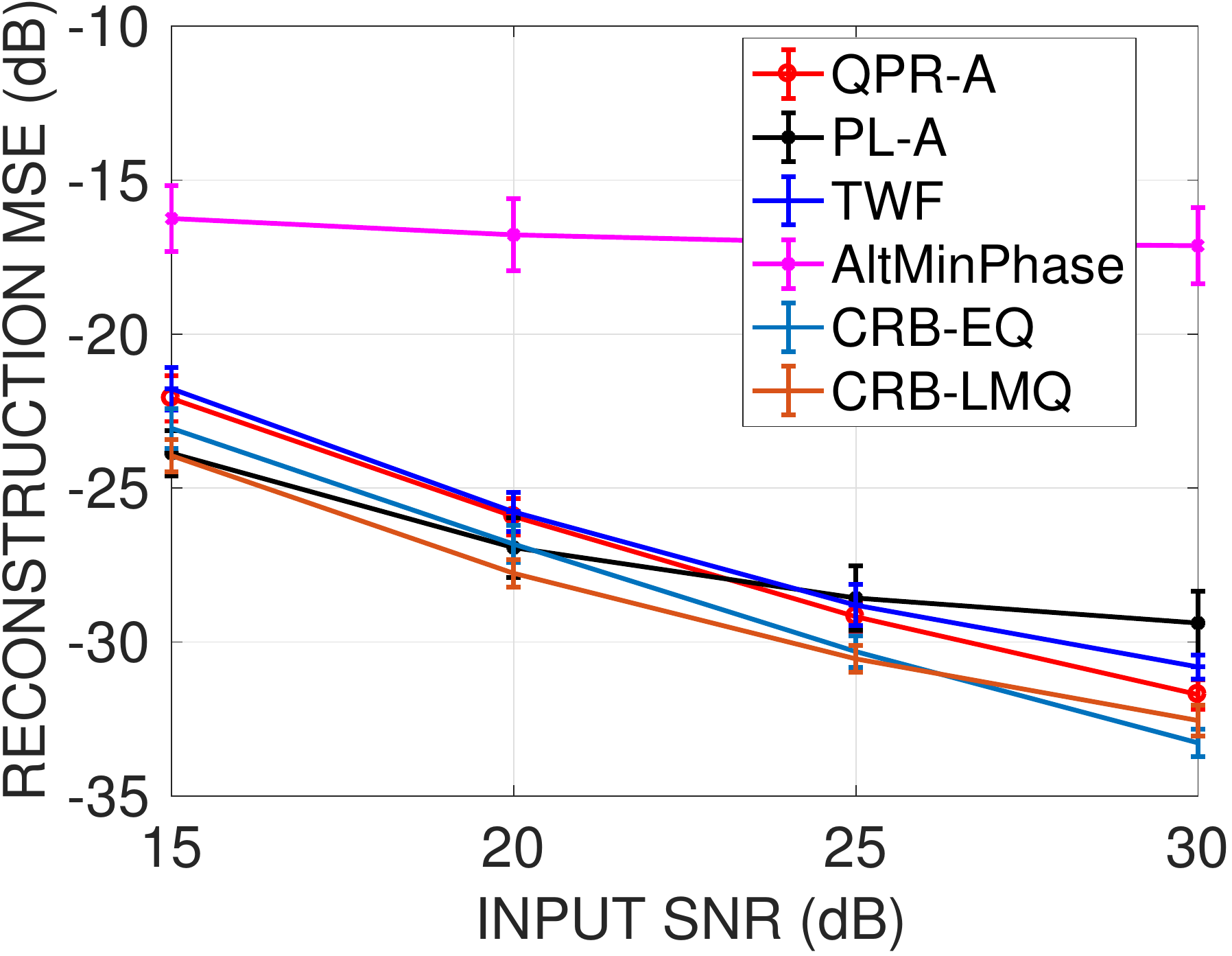}}
\caption{\small{Reconstruction MSE values of different algorithms versus the CRB. The over-sampling factor is $\frac{m}{n}=10$. }}
\label{algo_vs_crb_fig}
\end{figure*}
%%%%%%%%%%%%%%%%%%%%%%%%%%%%%%%%%%%%%%%%%%%%%  
\section{Noise Robustness: MSE vis-\`a-vis the CRB}
\label{np_crb_sec_qpr_ch}
\indent We now consider the effect of additive white Gaussian noise, prior to quantization, giving rise to quantized measurements\begin{equation}
y_i=\mathcal{Q}\left(\left| \boldsymbol a_i^\top \boldsymbol x^*  \right|^2 +\xi_i\right), i=1:m,
\label{klevel_noisy_meas_eq}
\end{equation}
where the noise samples $\xi_i$ are drawn independently from the $\mathcal{N}\left(0,\sigma_{\xi}^2\right)$ distribution. For illustration, the ground-truth signal $\boldsymbol x^*$ is taken as a sum of two sinusoids:
\begin{equation}
x_{\ell}^*=C_{\boldsymbol x^*}\left[1.5 \sin\left(\frac{4\pi(\ell-1)}{n}\right) + 2.5 \cos\left(\frac{14\pi(\ell-1)}{n}\right)\right],
\label{ground_truth_crb}
\end{equation}
where $\ell=1, 2,\cdots,n=32$ and the constant $C_{\boldsymbol x^*}$ is chosen such that $\left\|\boldsymbol x^*\right\|_2=1$. The derivation of the CRB for a $k$-level quantizer is given in Section~\ref{crb_derivation_kbit} of the supplementary material. Since the CRB is used as a theoretical benchmark, we use the reconstruction MSE as a performance metric. The reconstruction MSEs of different algorithms are compared against the CRB for three different quantization levels, namely $k=4, 8,$ and $16$. The MSEs are computed according to \eqref{mse_def2}, and averaged over $20$ noise realizations corresponding to a fixed measurement matrix $\boldsymbol A= (\boldsymbol a_1, \boldsymbol a_2,\cdots, \boldsymbol a_m)^\top$, whose entries are i.i.d. and follow the $\mathcal{N}(0,1)$ distribution. Since the measurement matrix is random, we need one more level of averaging, which is performed over $20$ different measurement matrices. The results are shown in Figure~\ref{algo_vs_crb_fig} as a function of the input SNR defined as $\text{SNR}_\text{in}=\frac{1}{m\sigma_{\xi}^2}{\sum_{i=1}^{m}\left| \boldsymbol a_i^\top \boldsymbol x^*  \right|^4}$.
The legends CRB-EQ and CRB-LMQ in Figure~\ref{algo_vs_crb_fig} denote the Cram\'er-Rao bounds corresponding to quantizers EQ and LMQ, respectively. The Fisher information matrix corresponding to the LMQ was found to be nearly rank-deficient for $k=4$ and therefore we omitted CRB-LMQ in Figure~\ref{algo_vs_crb_fig}(a).\\
\indent We observe that QPR-A attains reconstruction MSEs within $2$--$3$ dB of the corresponding CRB, whereas the other algorithms do not follow the CRB with increasing input SNR, especially for coarse quantization ($k=4$). For input SNR greater than $20$ dB, QPR-A has a reconstruction MSE closer to the CRB than other techniques. At low input SNRs (below $15$ dB), the additive noise leads to a violation of the consistency condition in \eqref{recovery_eq1}, which forms the basis of QPR-A. As a result, the QPR-A algorithm does not lead to a significant improvement over TWF and PL-A. We believe that a relaxed consistency condition in \eqref{recovery_eq1} to account for noise might result in more accurate recovery at low input SNRs. This aspect requires a separate investigation. 
\begin{figure}[t]
\centering
\subfigure[SQPR-A reconstruction]{
\includegraphics[width=1.5in]{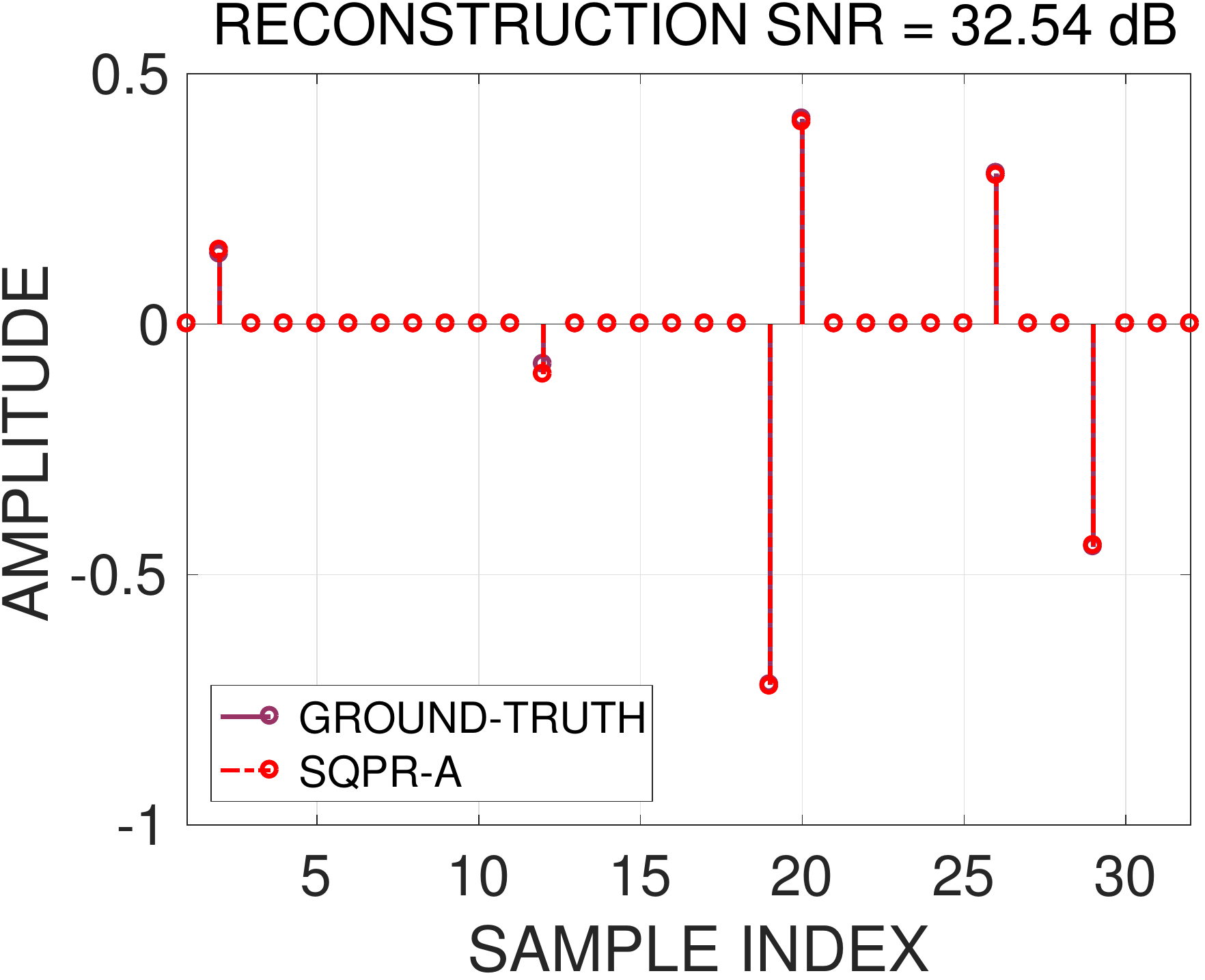}}
\subfigure[QPR-A reconstruction]{
\includegraphics[width=1.5in]{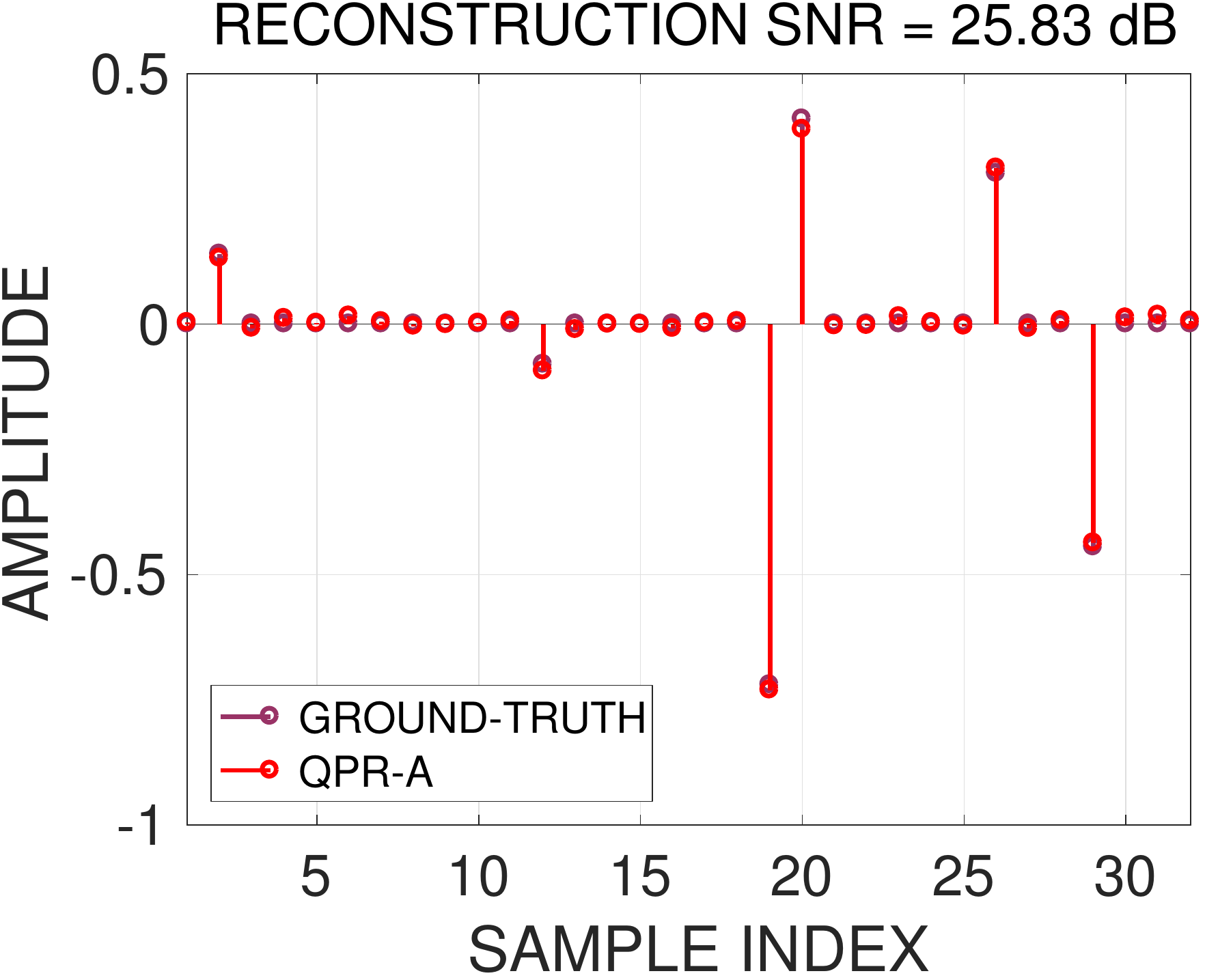}}
\caption{\small{Comparison of SQPR-A with QPR-A with $m=20n$ binary measurements ($k=2$). We set the sparsity level as $s=0.2n$, and the total number of iterations is $N_{\text{iter}}=100$. }}
\label{SQPR_vs_QPR_comparison_figure_an_example}
\end{figure}
%%%%%%%%%%%%%%%%%%%%%%%%%%%%%%%%%%%%%%%%%%%%%
%%%%%%%%%%%%%%%%%%%%%%%%%%%%%%%%%%%%%%%%%%%%%
\begin{figure}[t]
\centering
\subfigure[$s=0.1n$, $m=10n$]{
\includegraphics[width=1.5in]{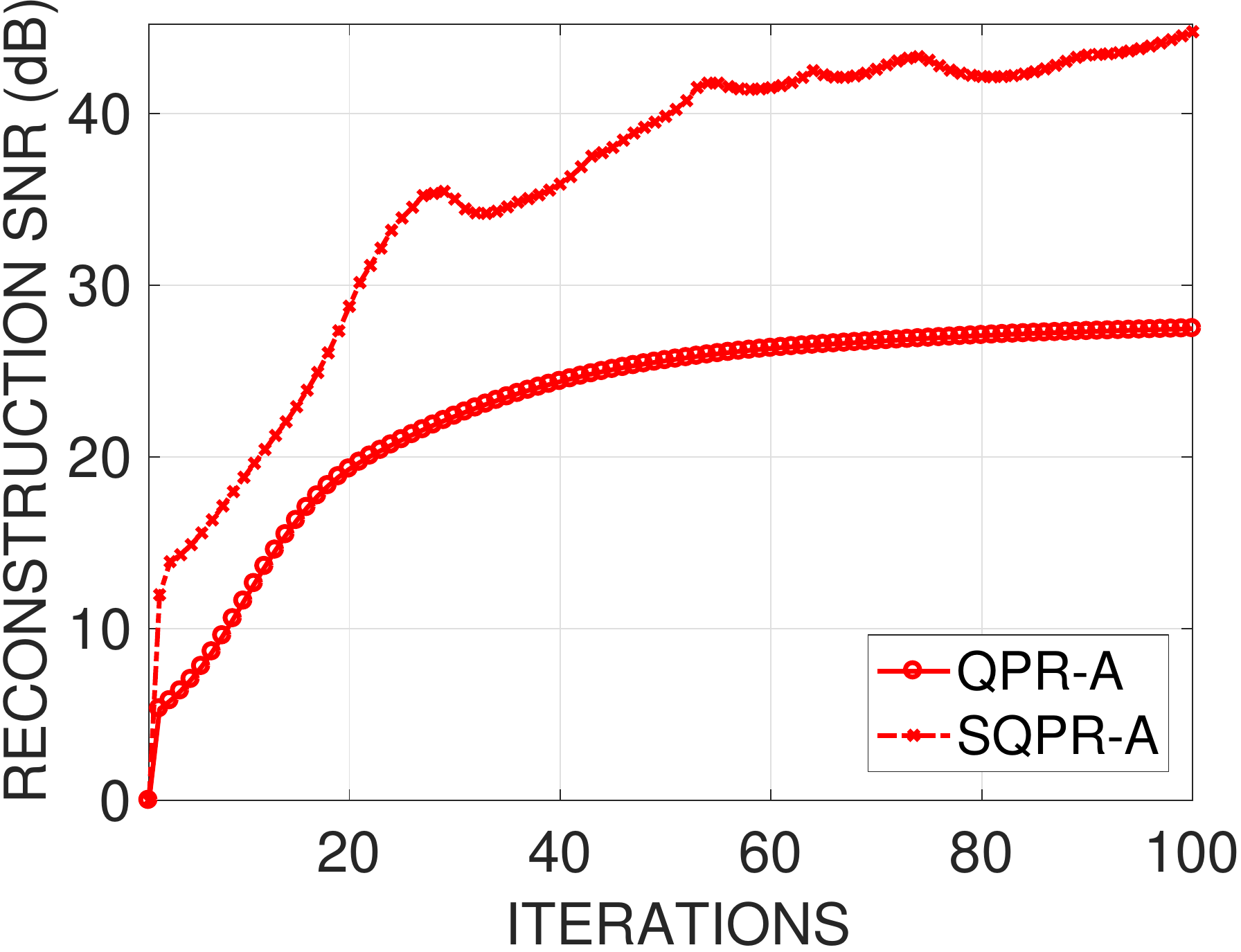}}
\subfigure[$s=0.1n$, $m=10n$]{
\includegraphics[width=1.5in]{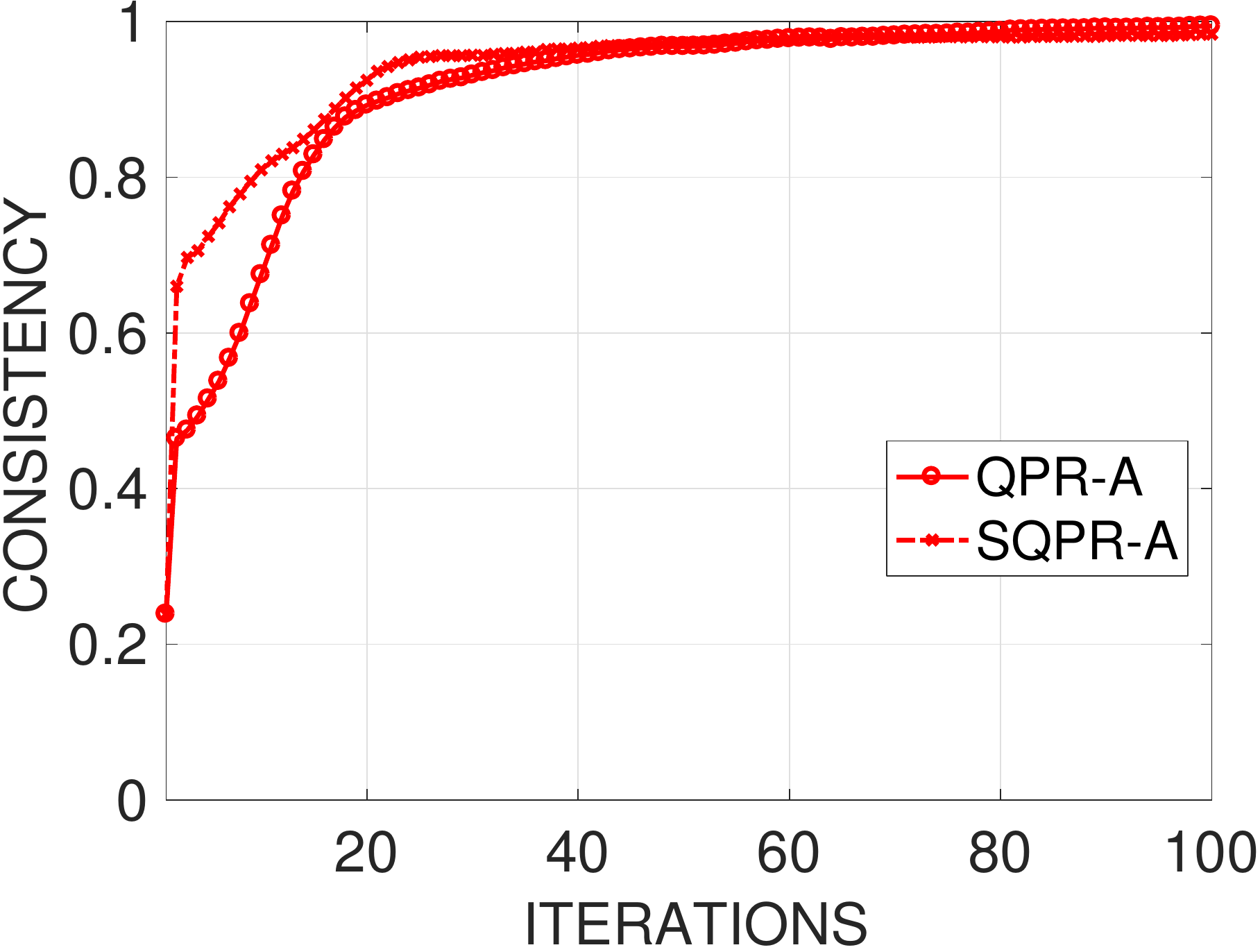}}\\
\subfigure[$s=0.3n$, $m=10n$]{
\includegraphics[width=1.5in]{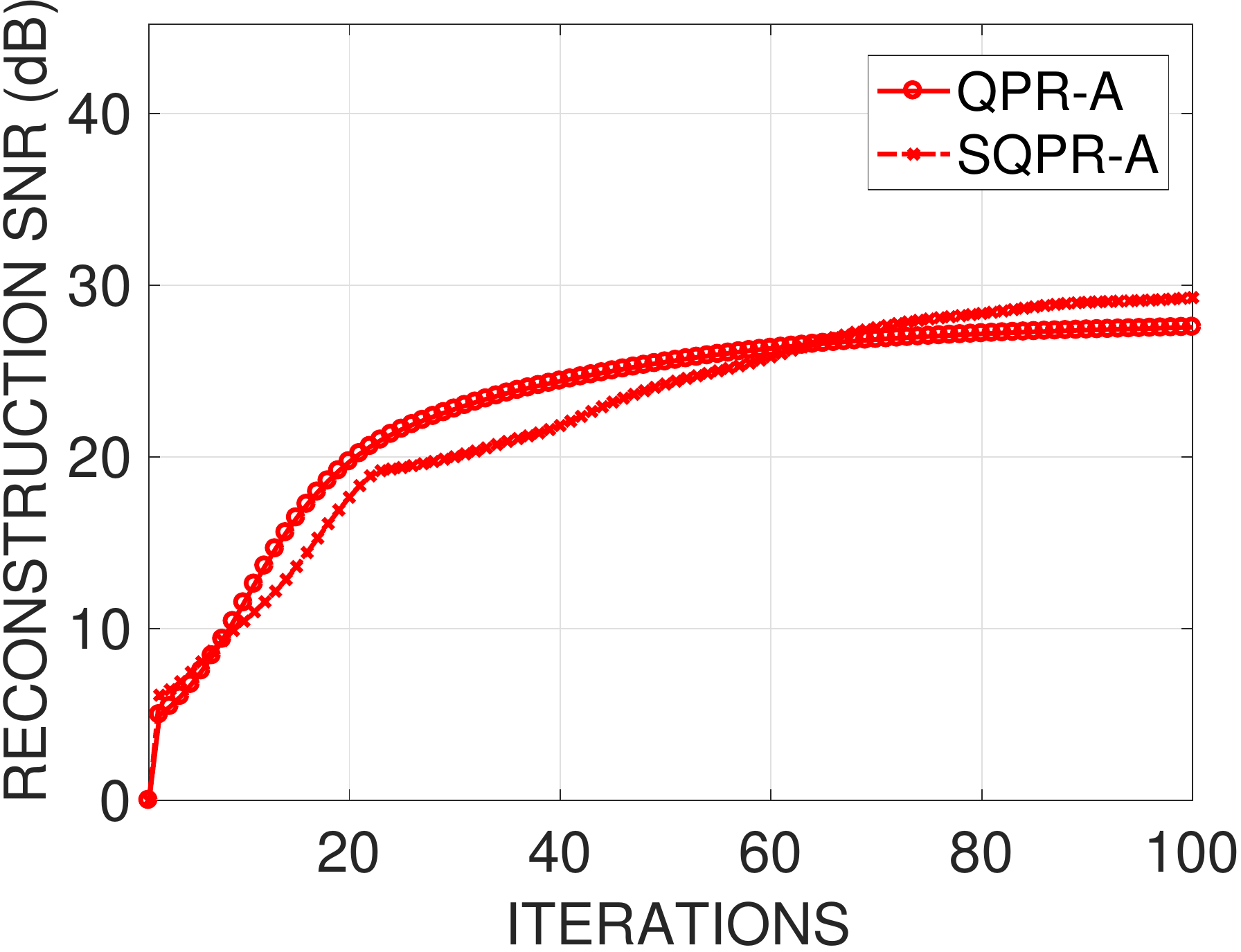}}
\subfigure[$s=0.3n$, $m=10n$]{
\includegraphics[width=1.5in]{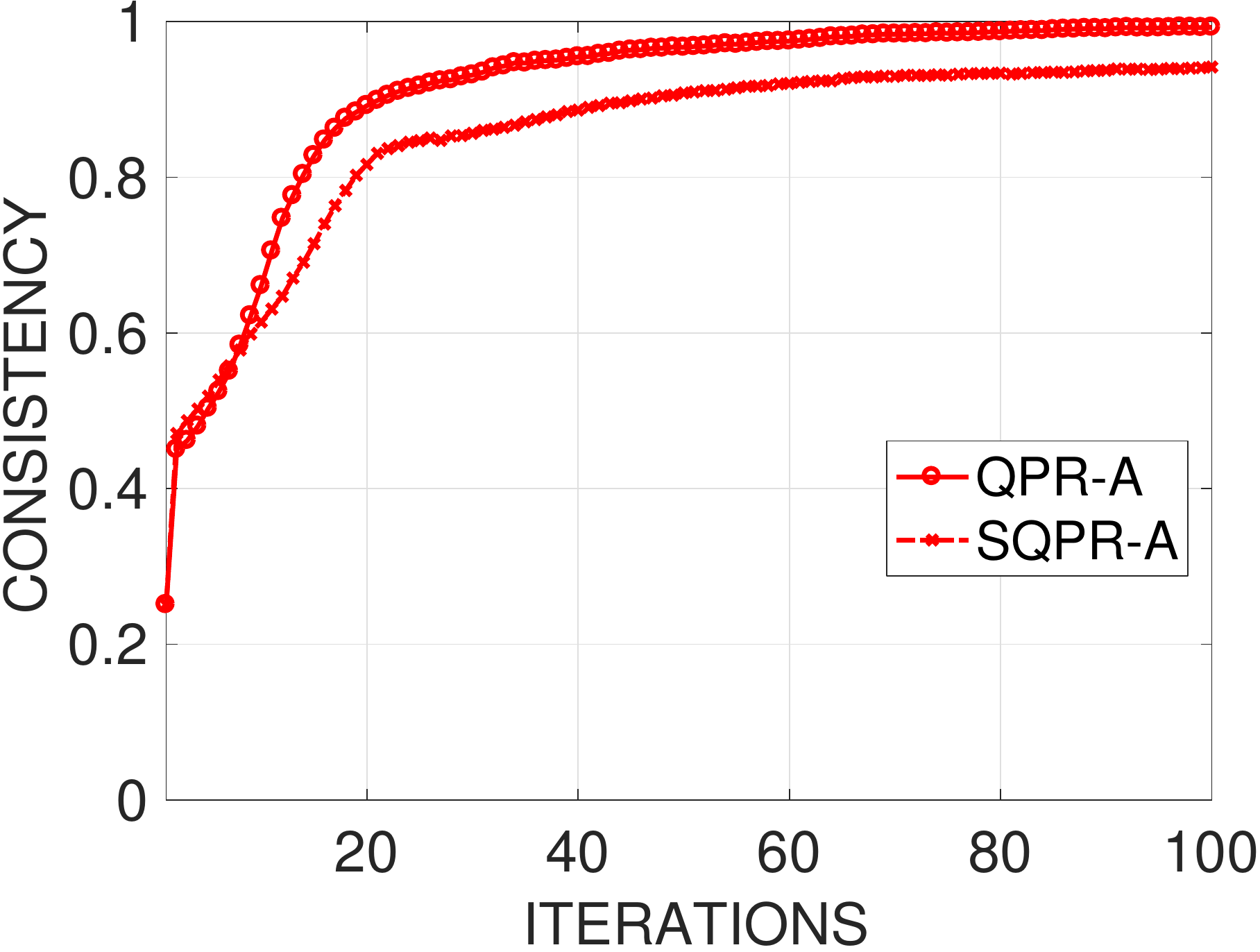}}\\
\caption{\small{Incorporation of the sparsity constraint: A comparison of SQPR-A with QPR-A. The first column shows reconstruction SNR and the second one shows the measurement consistency index. The number of quantization levels is $k = 4$. The values of $m$ and sparsity $s$ are indicated below each figure. }}
\label{QPR_sparse_versus_dense_figure}
\end{figure}
%%%%%%%%%%%%%%%%%%%%%%%%%%%%%%%%%%%%%%%%%%%%%
\section{Numerical Experiments on QPR With Sparsity}
\label{SQPR_results_sec}
\indent We now take into account sparsity of the signal and analyze the reconstruction capability of SQPR-A vis-\`a-vis the state-of-the-art algorithms for sparse PR. As the experimental results show, when the ground-truth is indeed sparse, incorporating that prior actually helps improve the reconstruction performance. This point is illustrated by comparing the reconstructed signals using SQPR-A and QPR-A for binary quantization ($k=2$) in Figure~\ref{SQPR_vs_QPR_comparison_figure_an_example}. We observe that imposition of the sparsity prior leads to an improvement of about $7$ dB in the reconstruction SNR. 
\subsection{Impact of the Sparsity Prior: SQPR-A versus QPR-A}
\indent We compare SQPR-A with QPR-A for different values of $m$ and the relative sparsity level $\rho=\frac{s}{n}$, the fraction of nonzero entries in the ground-truth. The results are averaged over $20$ trials and presented in Figures~\ref{QPR_sparse_versus_dense_figure}(a) and \ref{QPR_sparse_versus_dense_figure}(b). The figures show a distinct improvement coming from the sparsity prior. For example, for sparsity level $\rho=0.1$, and $m=10n$, SQPR-A is about $12$ dB better than QPR-A. The measurement consistency index is also higher and exhibits a faster convergence with iterations. As the underlying signal gets denser, the improvement one can achieve by enforcing sparsity diminishes (cf. Figures~\ref{QPR_sparse_versus_dense_figure}(c) and \ref{QPR_sparse_versus_dense_figure}(d)). 
\begin{figure}[t]
\centering
\subfigure{
\includegraphics[width=2.9in]{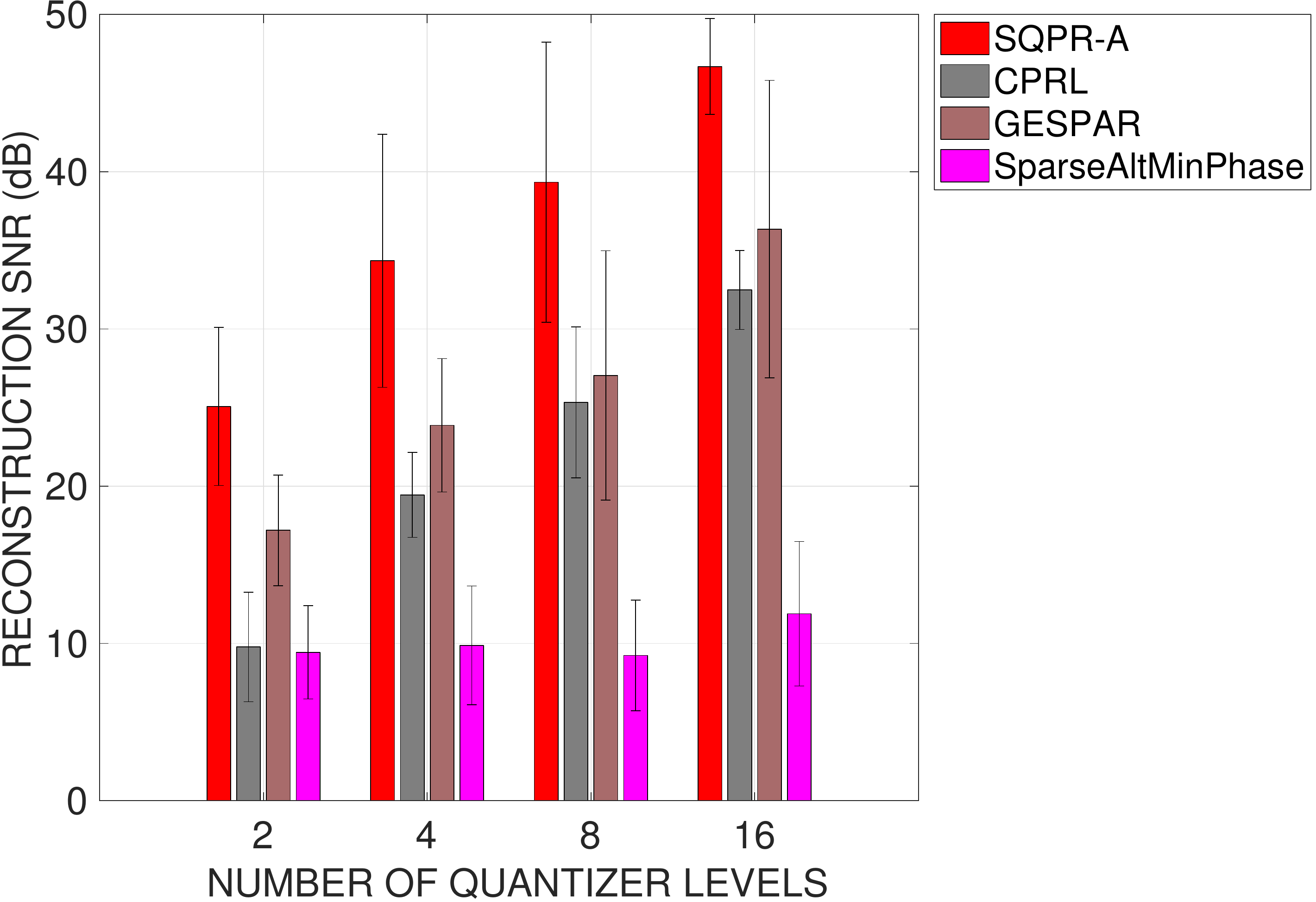}}
\caption{\small{Comparison of the average and the standard deviation of the output SNR obtained using different sparse PR algorithms, calculated over $20$ trials, with $m=10n$, $s=0.2n$, and $N_{\text{iter}}=100$. The total number of swaps in GESPAR is taken as $N_{\text{swap}}=1000$. }}
\label{snr_comparison_sparse_fig}
\end{figure}
%%%%%%%%%%%%%%%%%%%%%%%%%%%%%%%%%%%%%%%%%%%%%
\subsection{SQPR-A Versus State-of-the-Art Sparse PR Algorithms}
\indent We now compare SQPR-A with three state-of-the-art \textit{sparse PR} techniques: (i) CPRL \cite{cpr_pl}; (ii) GESPAR \cite{gespar}; and (iii) \textit{SparseAltMinPhase} \cite{netrapalli_altminpr}, the sparse counterpart of AltMinPhase. Recall that in CPRL, sparsity is enforced by incorporating an $\ell_1$ penalty term, thereby leading to the following semi-definite program:
\begin{equation}
\arg\underset{\boldsymbol X \succeq 0}{\min}\text{\,}\text{Tr}(\boldsymbol X) + \lambda_1 \| \boldsymbol X\|_1 \text{\,\,s.t.\,\,}\sum_{i=1}^{m}\left(\text{Tr}\left( \boldsymbol A_i \boldsymbol X\right)-y_i\right)^2\leq \epsilon^2.
\label{cprl_opt}
\end{equation}
We employ the Matlab implementation of CPRL available on the authors' web-page\footnote{\url{http://users.isy.liu.se/rt/ohlsson/code.html}.}, which uses the CVX package \cite{grant_cvx} to solve \eqref{cprl_opt}. To make a fair comparison, we set
\begin{equation*}
\epsilon=\left(\sum_{i=1}^{m}\left(\text{Tr}\left( \boldsymbol A_i \boldsymbol X^*\right)-\mathcal{Q}\left(\text{Tr}\left( \boldsymbol A_i \boldsymbol X^*\right) \right)\right)^2\right)^{\frac{1}{2}},
\end{equation*}
where $\boldsymbol X^*=\boldsymbol x^* \boldsymbol x^{*\top}$. The parameter $\lambda_1$ is set to $50$. In implementing GESPAR, we choose the total number of swaps \cite{gespar} as $N_{\text{swap}}=1000$. We found experimentally that increasing the number of swaps beyond $1000$ led to no significant improvement in reconstruction performance. The number of iterations in SparseAltMinPhase and SQPR-A is taken as $N_{\text{iter}}=100$. The sample size is $m=10n$, and the relative sparsity is taken as $\frac{s}{n}=0.2$. The support indices of the ground-truth are drawn uniformly at random from ${n}\choose{s}$ possibilities and their magnitudes are sampled uniformly on the unit-sphere in $\mathbb{R}^{s}$.\\
\indent The average and the standard deviations of the output SNR corresponding to the algorithms under comparison, calculated over $20$ trials, are shown in Figure \ref{snr_comparison_sparse_fig}, for four different values of $k$. We observe that SQPR-A consistently outperforms the closest competing technique, namely GESPAR, by approximately $8$ to $10$ dB. However, the standard deviations of the reconstruction SNR of SQPR-A and the competing techniques are high. The variations tend to reduce with finer quantization. This experiment demonstrates that SQPR-A, which is custom-designed for finite-precision measurements, leads to a more accurate reconstruction of sparse signals compared with the sparse PR algorithms that do not take measurement quantization explicitly into account.
\section{Summary and Conclusions}
\label{conclusion_sec}
We have introduced the problem of phase retrieval in the presence of measurement quantization and developed an iterative rank-1 projection-based algorithm, referred to as QPR-A, by combining the idea of lifting with the principle of consistent reconstruction. We have analyzed the effects of quantization on quadratic measurements, in terms of distinguishability and noise robustness, when the sampling vectors are drawn from a Gaussian ensemble. We have shown that incorporating signal sparsity, an important prior in a variety of practical applications, is possible by applying a hard-thresholding operator on the estimates produced within every iteration of QPR-A, resulting in the sparse version of QPR-A, which is referred to as SQPR-A. Extensive performance comparison of QPR-A and SQPR-A with several state-of-the-art PR and sparse PR techniques, respectively, has established the superiority of the proposed approach in terms of the reconstruction accuracy. In particular, the gain in the output SNR has been shown to be about $5$ to $10$ dB, depending on the number of bits allocated by the quantizer for each measurement and the total number of measurements.~Although we have demonstrated that quantizers with equiprobable intervals work reasonably well, the issue of selecting optimal thresholds by minimizing the expected reconstruction error requires further investigation. We have demonstrated that the PGD scheme for QPR has the descent property (cf. supplementary material). Establishing analytical convergence guarantees for QPR-A and SQPR-A (with the momentum factor included) requires a separate investigation.   
\section{Supplementary Material}
This supplementary material contains the following:
\begin{enumerate}
	\item An example of image reconstruction using the proposed QPR-A algorithm;
	\item An analysis of the optimization objective in the QPR framework; 
	\item Derivation of the Cram\'er-Rao Bound for QPR; and
	\item A proof of the \textit{descent property} of the projected gradient scheme for QPR. 
\end{enumerate}
%%%%%%%%%%%%%%%%%%%%%%%%%%%%%%%%%%%%%%%%%%%%%
\begin{figure}[t]
	\centering
	\subfigure[Ground-truth]{
		\includegraphics[width=1.5in]{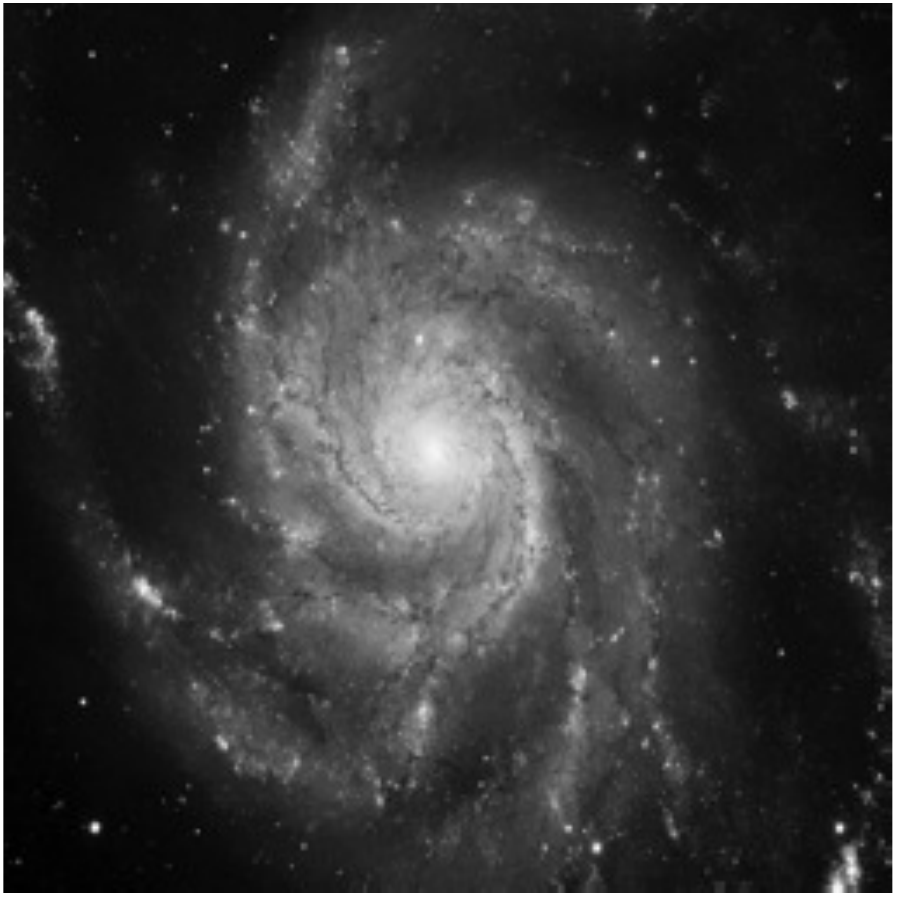}}
	\subfigure[$m=10n$, $\left(41.94 \text{\,dB}, 0.95\right)$]{
		\includegraphics[width=1.5in]{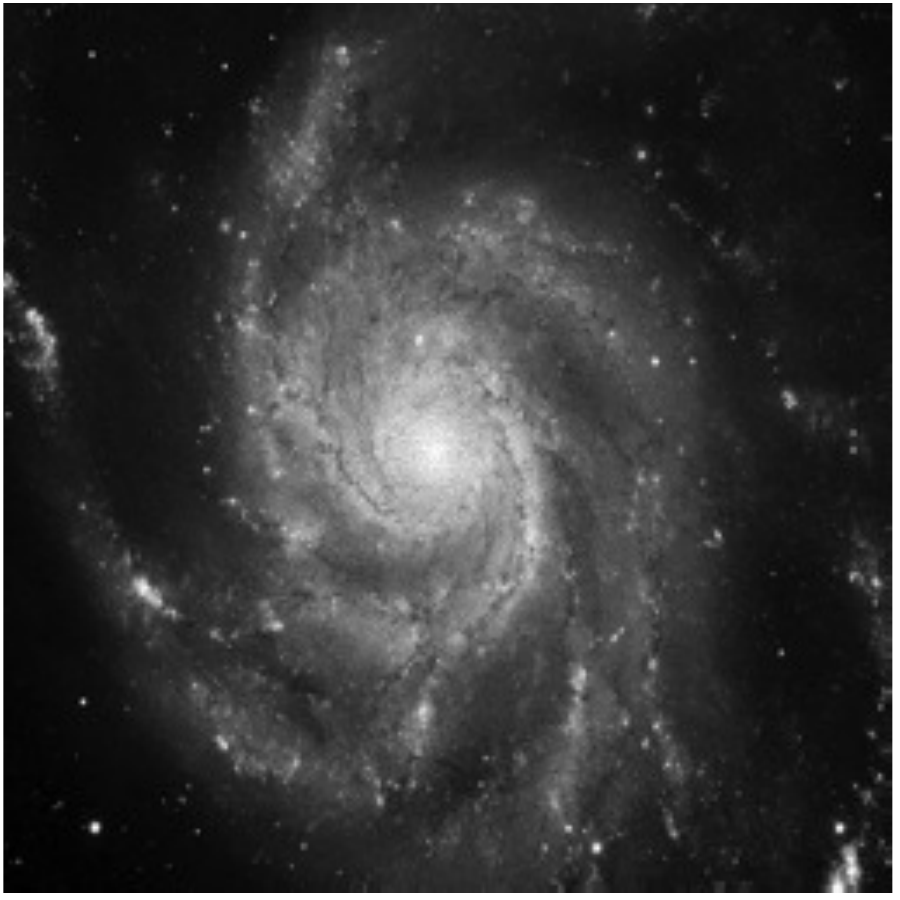}}\\
	\subfigure[$m=7.5n$, $\left(39.64 \text{\,dB}, 0.92\right)$]{
		\includegraphics[width=1.5in]{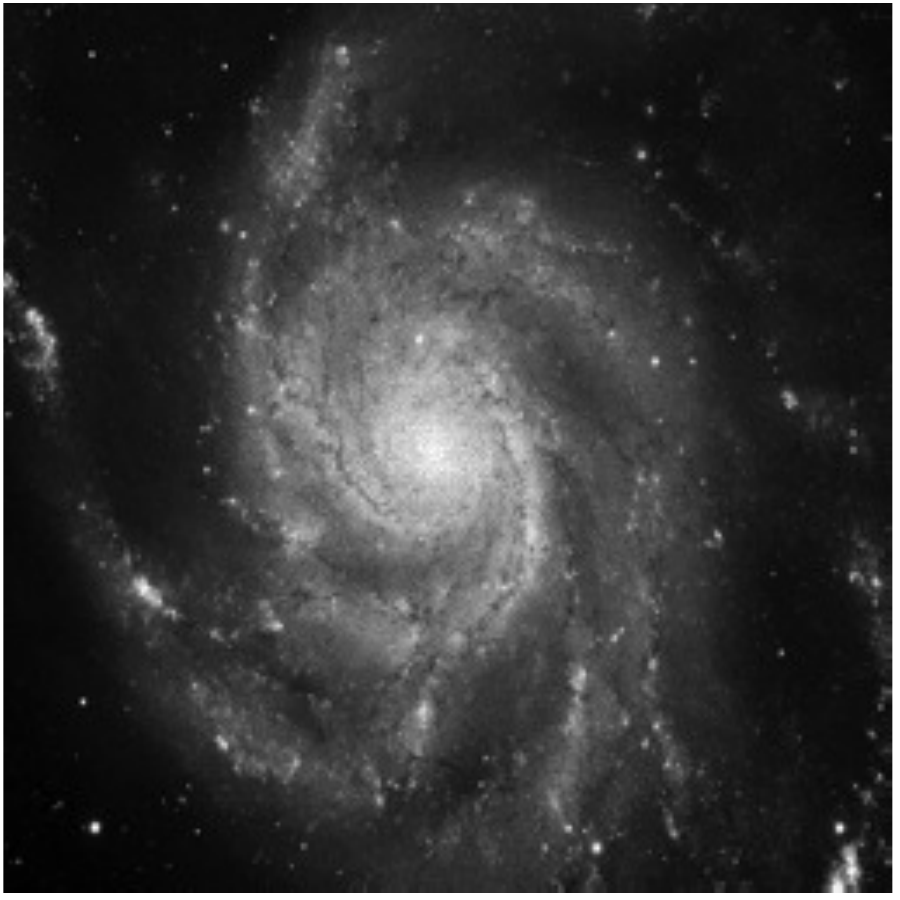}}
	\subfigure[$m=5n$, $\left(35.90 \text{\,dB}, 0.84\right)$]{
		\includegraphics[width=1.5in]{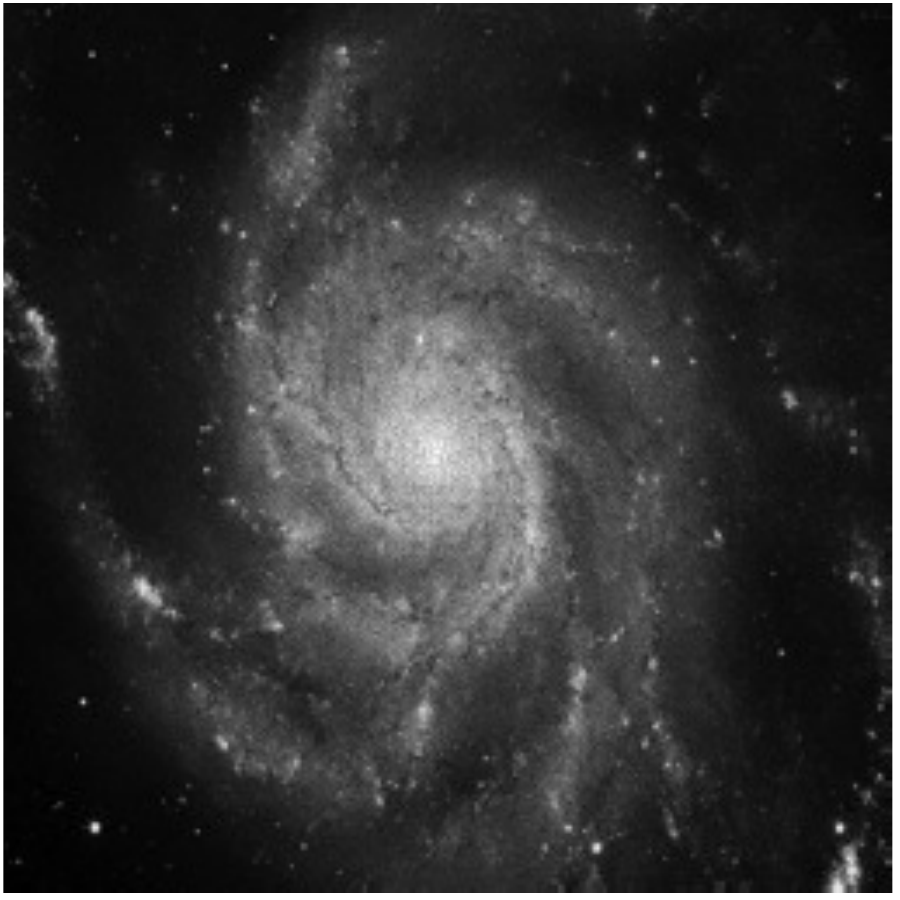}}
	\caption{\small{Image reconstruction by QPR-A, where the measurements are quantized using three bits, for different number of measurements $m$. The PSNR (in dB) and SSIM are indicated inside the parentheses. The number of iterations is $N_{\text{iter}}=60$.}}
	\label{QPRA_image_recon}
\end{figure}
%%%%%%%%%%%%%%%%%%%%%%%%%%%%%%%%%%%%%%%%%%%%%
\begin{figure}[t]
	\centering
	\subfigure[$m=10n$, $\left(35.76 \text{\,dB}, 0.84\right)$]{
		\includegraphics[width=1.5in]{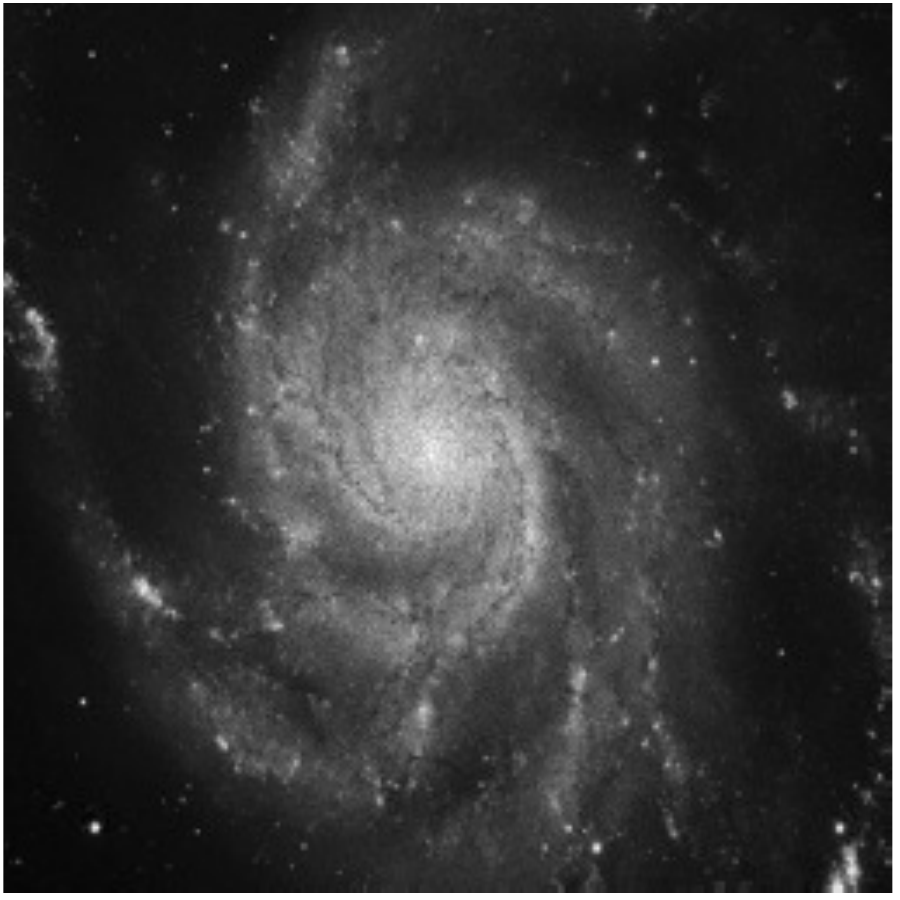}}
	\subfigure[$m=5n$, $\left(56.88 \text{\,dB}, 1.00\right)$]{
		\includegraphics[width=1.5in]{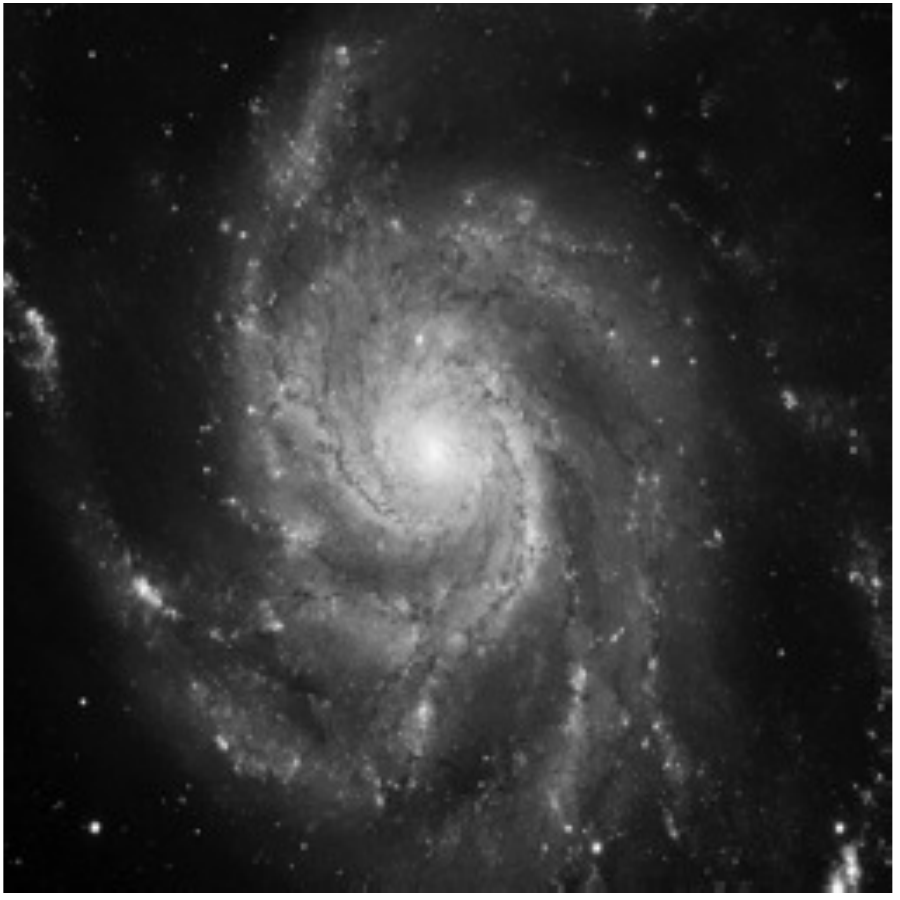}}\\
	\subfigure[PSNR and SSIM for QPR-A and PL-A for $m=5n$.]{
		\includegraphics[width=2.2in]{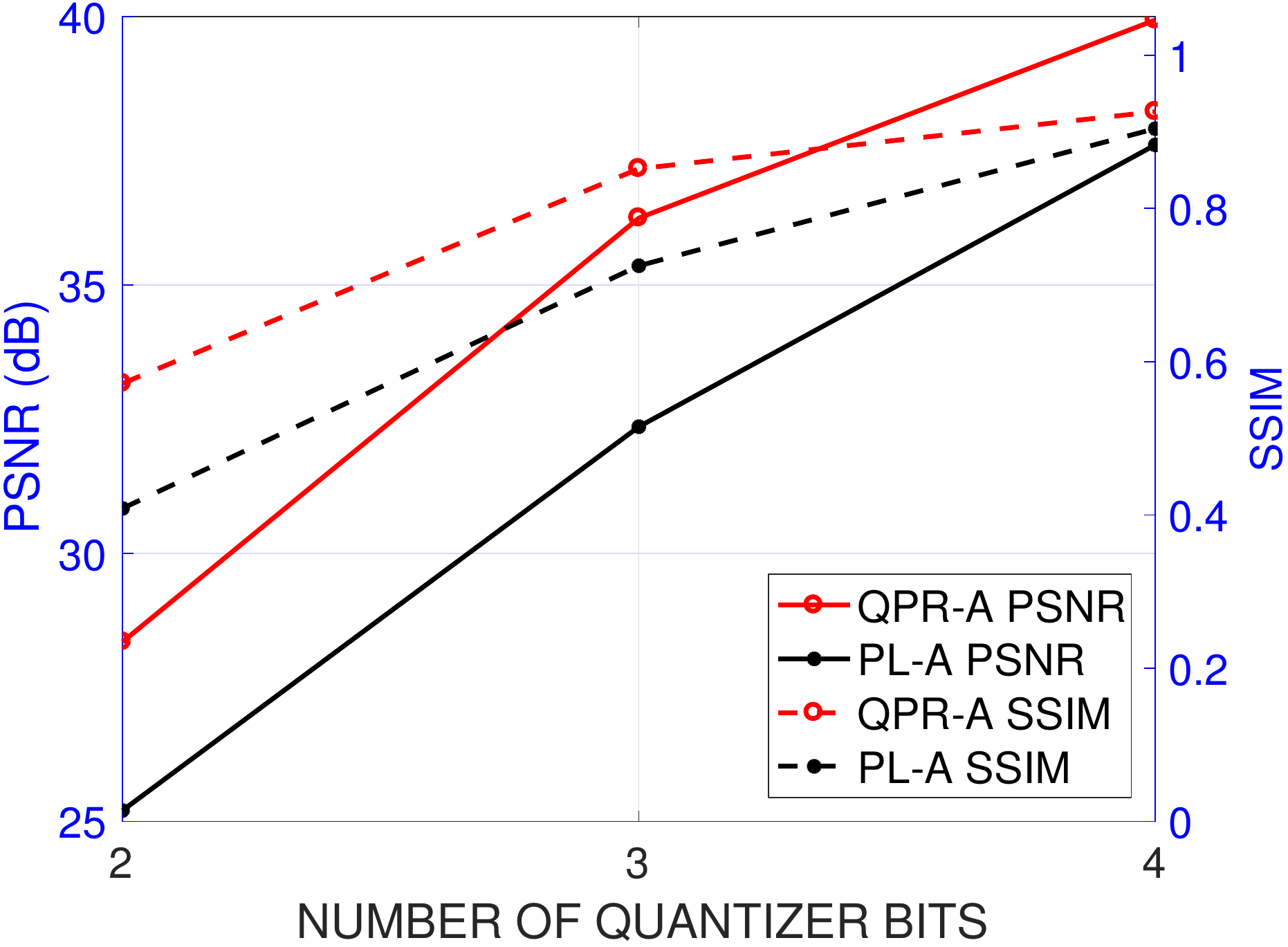}}
	\caption{\small{Image reconstruction by PL-A and QPR-A: (a) PL-A reconstruction where measurements are quantized with three bits; (b) PL-A reconstruction where the measurements are acquired with double precision; and (c) QPR-A versus PL-A for different values of quantization bits $\log_2k$.  The number of iterations is $N_{\text{iter}}=60$.}}
	\label{PLA_image_recon}
\end{figure}
%%%%%%%%%%%%%%%%%%%%%%%%%%%%%%%%%%%%%%%%%%%%%
\subsection{Image Reconstruction Using QPR-A}
\label{qpr_image_recon_sec}
\indent In Figure~\ref{QPRA_image_recon}, we show results of the application of QPR-A for image reconstruction from intensity measurements quantized using only three bits. The optimal step-size $\eta^t$ for QPR-A is selected by a grid-search over $[0,5.5\times 10^{-3}]$ with a spacing of $10^{-6}$. The ground-truth image is shown in Figure~\ref{QPRA_image_recon}(a); and the reconstructed images corresponding to $m=10n$, $m=7.5n$, and $m=5n$, where $n=256\times 256$ is the number of pixels in the image, are shown in Figures~\ref{QPRA_image_recon}(b), \ref{QPRA_image_recon}(c), and \ref{QPRA_image_recon}(d), respectively. The quality of reconstruction is measured using peak signal-to-noise ratio (PSNR) and structural similarity index (SSIM) \cite{ssim_paper_ref}, which show that the reconstructed images obtained using QPR-A are close to the ground-truth.\\
\indent A comparison of QPR-A with PL-A is shown in Figure~\ref{PLA_image_recon}. The reconstruction in Figure~\ref{PLA_image_recon}(a) corresponds to PL-A with three-bit quantization; and is approximately equivalent to the reconstructed image in Figure~\ref{QPRA_image_recon}(d) in terms of PSNR and SSIM. This experiment demonstrates that QPR-A with $m=5n$ achieves a performance comparable to that of PL-A with $m=10n$ measurements. In Figure~\ref{PLA_image_recon}(b), we show the output of PL-A for double-precision measurements for the sake of visual comparison. We observe that PL-A leads to near-accurate reconstruction in the absence of quantization noise; but tends to underperform when the measurements are of lower precision. Finally, the output PSNR and SSIM of QPR-A and PL-A are plotted in Figure~\ref{PLA_image_recon}(c) with respect to the number of bits $\log_2k$. The number of measurements is set to $m=5n$ and both algorithms are iterated $N_{\text{iter}}=60$ times. The QPR-A algorithm consistently dominates PL-A with a significant margin, both in terms of PSNR and SSIM. These results underline the importance of designing a quantization-aware PR algorithm particularly when one can only encode the measurements with a lower precision.
%\appendices 
\subsection{Analysis of the Cost Function}
\label{analysiscost}
\indent We provide upper and lower bounds on the cost $F(\boldsymbol X)$ as functions of $\rho_{\boldsymbol x}=\left| \boldsymbol x^{*\top}\boldsymbol x\right|$, where $\boldsymbol x$ is an estimate of the ground-truth $\boldsymbol x^{*}$, and the quantizer parameters, namely, precision $\delta$ defined in \eqref{delta_def_eq}, the penultimate quantizer threshold $\tau_{k-1}$, and a squared precision parameter $\delta_{\text{sq}}$ defined below:
\begin{equation}
\delta_{\text{sq}}=\underset{1\leq j \leq k}{\max}\left(s_j^2-\tau_{j-1}^2\right).
\label{del_sq_def}
\end{equation}
The motivation for providing such bounds is to guarantee that if the quantization precision is high, then forcing the cost $F(\boldsymbol X)$ to $0$ drives $\rho_{\boldsymbol x} \rightarrow 1$, which in turn guarantees that $\boldsymbol x \rightarrow \pm\boldsymbol x^{*}$, with a high probability. Recall that we are working with real-valued signals and sampling vectors, which restricts the phase ambiguity only up to a global sign flip.\\
\indent To begin with, we establish a relationship between $F(\boldsymbol X)$ defined in \eqref{F_def_eq} with the usual two-sided quadratic cost $Q(\boldsymbol X)$ given as
\begin{eqnarray}
Q\left(\boldsymbol X \right)&=&\frac{1}{2}\sum_{i=1}^{m}\left( y_i- \text{Tr}\left(\boldsymbol A_i \boldsymbol X\right)\right)^2\nonumber\\&=&\frac{1}{2}\sum_{j=1}^{k}\sum_{i=1}^{m_j}( s_j-\text{Tr}(\boldsymbol A_{i}^j \boldsymbol X) )^2.\nonumber
\end{eqnarray}
Suppose the encoding symbols $s_j$ are chosen such that they satisfy \eqref{encoding}. Consider a fixed $i$ and assume that $y_i=s_j$, for some $j\in\left\{1,2,\cdots,k\right\}$. 
\begin{propos}
	\label{cost_anal1}
	For any $d_0\geq\tau_{k-1}$ and $\boldsymbol X=\boldsymbol x\boldsymbol x^\top$, the cost function $F\left(\boldsymbol X \right)$ defined in \eqref{F_def_eq} satisfies
	\begin{equation*}
	0\leq Q\left(\boldsymbol X \right)-F\left(\boldsymbol X \right)\leq \frac{m}{2}\max\left\{ \delta^2+2\delta d_0, \delta_{\text{sq}}\right\},
	\end{equation*}
	with probability $\left[\frac{\gamma\left(\frac{d_0}{2},\frac{1}{2}\right)}{\sqrt{\pi}}\right]^m$. Therefore, for a high-precision quantizer for which $\delta$ and $\delta_{\text{sq}}$ are small, $F\left(\boldsymbol X \right)$ is close to $Q\left(\boldsymbol X \right)$ with high probability.
\end{propos}
\noindent\textit{Proof}: Let $d_0\geq \tau_{k-1}$ be a constant such that $\left| \boldsymbol a_i^\top \boldsymbol x\right|^2<d_0$, for all $i$, so that all measurements fall in the interval $\left[0,d_0\right]$, with probability
\begin{equation}
p_{\text{bdd}}=\left[\frac{\gamma\left(\frac{d_0}{2},\frac{1}{2}\right)}{\sqrt{\pi}}\right]^m,
\label{p_bdd_eq}
\end{equation}
where the subscript bdd is a shorthand to indicate that the intensity measurements are bounded. The following three cases are considered separately:\\
Case-I: $\tau_{j-1}<\text{Tr}(\boldsymbol A_{i}^j \boldsymbol X)<\tau_j$, thereby implying that 
$f_{i,j}^1+f_{i,j}^2=0$, where $f_{i,j}^1$ and $f_{i,j}^2$ are defined as 
\begin{eqnarray*}
	f_{i,j}^1=f(\tau_j-\text{Tr}(\boldsymbol A_{i}^j \boldsymbol X)) \text{\,\,and\,\,}f_{i,j}^2 = f(\text{Tr}(\boldsymbol A_{i}^j \boldsymbol X)-\tau_{j-1}),
\end{eqnarray*}
respectively. As a consequence, we have that
\begin{eqnarray}
\frac{1}{2}( s_j-\text{Tr}(\boldsymbol A_{i}^j \boldsymbol X))^2-\left[f_{i,j}^1+f_{i,j}^2\right] &\leq& \frac{\delta^2}{2}.
\label{case1_ineq}
\end{eqnarray}
Case-II: $\text{Tr}(\boldsymbol A_{i}^j \boldsymbol X)<\tau_{j-1}$ (cannot happen for $j=1$, which corresponds to $\tau_0=0$, due to the innate nonnegativity of the intensity measurements), which leads to $f_{i,j}^1=0$ and $f_{i,j}^2=\frac{1}{2}(\tau_{j-1}-\text{Tr}(\boldsymbol A_{i}^j \boldsymbol X))^2$. Consequently, we observe that 
\begin{eqnarray}
\frac{1}{2}( s_j-\text{Tr}(\boldsymbol A_{i}^j \boldsymbol X))^2-\left[f_{i,j}^1+f_{i,j}^2\right]\leq\frac{1}{2}\left(s_j^2-\tau_{j-1}^2\right),
\label{case2_ineq}
\end{eqnarray}
where the inequality in \eqref{case2_ineq} follows from the fact that 
\begin{equation*}
-s_j \text{Tr} (\boldsymbol A_{i}^j \boldsymbol X )+\tau_{j-1}\text{Tr}(\boldsymbol A_{i}^j \boldsymbol X)\leq 0,
\end{equation*}
which, in turn, is a consequence of \eqref{encoding}. Using the definition of $\delta_{\text{sq}}$ in \eqref{del_sq_def}, we get
\begin{eqnarray}
\frac{1}{2}(s_j-\text{Tr}(\boldsymbol A_{i}^j \boldsymbol X))^2-\left[f_{i,j}^1+f_{i,j}^2\right]\leq\frac{\delta_{\text{sq}}}{2}.
\label{case2_ineq_final}
\end{eqnarray}
Case-III: $\tau_{j}<\text{Tr}\left(\boldsymbol A_{i,j} \boldsymbol X\right)$, $j\in\left\{1,2,\cdots,k-1\right\}$ (since this case cannot arise for $j=k$, as $\tau_k=+\infty$), which results in $f_{i,j}^2=0$ and $f_{i,j}^1=\frac{1}{2}\left( \tau_{j}-\text{Tr}\left(\boldsymbol A_{i,j} \boldsymbol X\right) \right)^2$. Defining $r_1=\tau_j-s_j$ and $r_2=\text{Tr}\left(\boldsymbol A_{i,j}\boldsymbol X\right)-\tau_j$, we have that
\begin{eqnarray}
&&\frac{1}{2}\left( s_j-\text{Tr}\left(\boldsymbol A_{i,j} \boldsymbol X\right) \right)^2-\left[f_{i,j}^1+f_{i,j}^2\right]\nonumber\\&=&\frac{1}{2}\left[\left(r_1+r_2\right)^2-r_2^2\right]\nonumber\\ &\leq& \frac{\delta^2}{2}+\delta d_0,
\label{case3_ineq}
\end{eqnarray}
since $r_1\leq \delta$ and $r_2 \leq d_0$. The proposition follows by combining \eqref{case1_ineq}, \eqref{case2_ineq_final}, and \eqref{case3_ineq} for all $m$ measurements.\hfill $\blacksquare$\\
In the following proposition, we upper-bound the difference $F\left(\boldsymbol X \right)-F\left(\boldsymbol X^* \right)$, for any $\boldsymbol X =\boldsymbol x \boldsymbol x^\top$, $\boldsymbol X^* =\boldsymbol x^* \boldsymbol x^{*\top}$, where $\boldsymbol x \in \mathbb{U}^n$. Since $F\left(\boldsymbol X^* \right)=0$, it suffices to bound $F\left(\boldsymbol X \right)$.
\begin{propos}
	For any $d_0\geq\tau_{k-1}$ and $\boldsymbol X=\boldsymbol x \boldsymbol x^\top$, where $\boldsymbol x \in \mathbb{U}^n$, the following happens with probability $\left[\frac{\gamma\left(\frac{d_0}{2},\frac{1}{2}\right)}{\sqrt{\pi}}\right]^m$:
	\begin{eqnarray*}
		F\left(\boldsymbol X \right)&\leq&\frac{m}{2}\max \left\{\delta^2,\left(d_0-\tau_{k-1}\right)^2\right\}\nonumber\\&+&\left(1-\rho_{\boldsymbol x}^2\right)\sum_{i=1}^{m}\left({\tilde a_{i1}^2+\tilde a_{i2}^2}\right)^2,
	\end{eqnarray*}
	where $\rho_{\boldsymbol x}=\left| \boldsymbol x^{*\top}\boldsymbol x\right|$, and ${\tilde a}_{i1}$ and ${\tilde a}_{i2}$ are the first two entries of ${\tilde{\boldsymbol a}_i} =  \boldsymbol R^\top \boldsymbol a_i$, where in turn $\boldsymbol R \boldsymbol D \boldsymbol R^\top$ is the spectral decomposition of $\boldsymbol x^* \boldsymbol x^{*\top}-\boldsymbol x \boldsymbol x^\top$.
	\label{cost_anal2}
\end{propos}
\noindent\textit{Proof}: Using Proposition \ref{cost_anal1}, we have that 
\begin{equation}
F\left(\boldsymbol X \right)\leq Q\left(\boldsymbol X \right)=\frac{1}{2}\sum_{i=1}^{m}\left( y_i- \text{Tr}\left(\boldsymbol A_i \boldsymbol X\right)\right)^2.
\label{prop5_anal1}
\end{equation}
Considering the additive noise model for quantization, we write $y_i=\mathcal{Q}\left(\text{Tr}\left(\boldsymbol A_i \boldsymbol X^*\right)\right)=\text{Tr}\left(\boldsymbol A_i \boldsymbol X^*\right)+w_i$ in \eqref{prop5_anal1}, where $w_i$ denotes the quantization noise. Consequently, 
\begin{eqnarray}
F\left(\boldsymbol X \right)&\leq& \frac{1}{2}\sum_{i=1}^{m}\left(\text{Tr}\left(\boldsymbol A_i \left(\boldsymbol X^*-\boldsymbol X\right)\right)+w_i\right)^2,\nonumber\\
&\leq& \sum_{i=1}^{m}(\,\underbrace{\boldsymbol a_i^\top \left(\boldsymbol X^*-\boldsymbol X\right)\boldsymbol a_i}_{\leq \sqrt{1-\rho_{\boldsymbol x}^2}\left( \tilde{a}_{i1}^2  + \tilde{a}_{i2}^2\right)}\,)^2+w_i^2,
\label{prop5_anal2}
\end{eqnarray}
where the inner inequality has been established using Proposition~\ref{error_prob_prop1} and \eqref{gamma_eval_eq}. The quantization noise $w_i$ satisfies
\begin{equation}
w_i\leq \max\left\{\delta,\left(d_0-\tau_{k-1}\right)\right\},
\end{equation}
whenever $\left| \boldsymbol a_i^\top \boldsymbol x\right|^2<d_0$, for all $i$, which occurs with probability $p_{\text{bdd}}$ given in \eqref{p_bdd_eq}. Therefore, adding over all measurements $i = 1:m$ establishes the proposition.\hfill $\blacksquare$
\begin{propos}
	For any $d_0 \geq \tau_{k-1}$ and $\boldsymbol X=\boldsymbol x \boldsymbol x^\top$, where $\boldsymbol x \in \mathbb{U}^n$, the cost $F\left(\boldsymbol X \right)$ satisfies the following lower-bound with probability $p_{\text{bdd}}$:
	\begin{eqnarray*}
		F\left(\boldsymbol X \right)&\geq& \frac{1-\rho_{\boldsymbol x}^2}{2} \sum_{i=1}^{m} ({\tilde a}_{i1}^2-{\tilde a}_{i2}^2)^2\\&-&\sqrt{1-\rho_{\boldsymbol x}^2}\,{\text{max}\{\delta,d_0-\tau_{k-1}\}} \sum_{i=1}^{m} ({\tilde a}_{i1}^2+{\tilde a}_{i2}^2)\\&-&\frac{m}{2}\max\left\{ \delta^2+2\delta d_0, \delta_{\text{sq}}\right\}.
	\end{eqnarray*}
	\label{cost_anal_lb}
\end{propos}
\noindent\textit{Proof}: From Proposition \ref{cost_anal1}, we have with probability $p_{\text{bdd}}$ that 
\begin{equation}
F\left(\boldsymbol X \right)\geq Q\left(\boldsymbol X \right)-\frac{m}{2}\max\left\{ \delta^2+2\delta d_0, \delta_{\text{sq}}\right\}.
\label{eq47}
\end{equation}
Recalling \eqref{prop5_anal1} and considering $y_i=\text{Tr}\left(\boldsymbol A_i \boldsymbol X^*\right)+w_i$, we get
\begin{eqnarray*}
	Q\left(\boldsymbol X \right)=\frac{1}{2}\sum_{i=1}^{m}\left(\text{Tr}\left(\boldsymbol A_i (\boldsymbol{X^*-X})\right)+w_i\right)^2.\\
\end{eqnarray*}
The quantity $Q\left(\boldsymbol X \right)$ could be lower-bounded as follows:
\begin{eqnarray*}
	Q\left(\boldsymbol X \right)&\geq& \frac{1}{2}\sum_{i=1}^{m}\underbrace{(\text{Tr}\left(\boldsymbol A_i (\boldsymbol{X^*-X})\right))^2}_{\geq (1-\rho_{\boldsymbol x}^2)({\tilde a}_{i1}^2-{\tilde a}_{i2}^2)^2} \\&-& \sum_{i=1}^{m}\underbrace{|w_i|}_{\leq \text{max}\{\delta,d_0-\tau_{k-1}\}}\underbrace{|\text{Tr}\left(\boldsymbol A_i (\boldsymbol{X^*-X})\right)|}_{\leq \sqrt{1-\rho_{\boldsymbol x}^2}({\tilde a}_{i1}^2+{\tilde a}_{i2}^2)},\\
	&\geq& \frac{1-\rho_{\boldsymbol x}^2}{2} \sum_{i=1}^{m} ({\tilde a}_{i1}^2-{\tilde a}_{i2}^2)^2\\&-&\sqrt{1-\rho_{\boldsymbol x}^2}\,{\text{max}\{\delta,d_0-\tau_{k-1}\}} \sum_{i=1}^{m} ({\tilde a}_{i1}^2+{\tilde a}_{i2}^2).
	\label{prop5_anal2_lb1}
\end{eqnarray*}
Combining the above lower bound on $Q\left(\boldsymbol X \right)$ with \eqref{eq47} completes the proof. \hfill $\blacksquare$\\
In summary, for high-precision quantization, minimizing the objective $F(\boldsymbol X)$ has the desirable effect of resulting in a solution that is close to $\boldsymbol x^*$ with a high probability. Our simulation results show that accurate reconstruction is possible even with coarse quantization.

\subsection{Derivation of the Cram\'er-Rao Bound}
\label{crb_derivation_kbit}
In this section, we derive the analytical expressions for the Cram\'er-Rao bounds (CRBs) corresponding to the multi-level quantization model in the presence of additive noise contamination prior to acquiring finite-precision measurements. The CRB serves as a benchmark for comparing the variance of the estimates produced by different algorithms in the presence of noise before quantization. The CRB expressions for PR are available  corresponding to various measurement models such as Gaussian noise corrupting the quadratic measurements \cite{balan1_qpr}, non-additive Gaussian noise prior to computing the quadratic measurement \cite{balan2_qpr}, uniformly distributed additive noise encountered in high-rate quantization \cite{pr_crb_fpp_qpr}, frame-based measurements \cite{bandeira_qpr}, and Fourier measurements \cite{cederquist_qpr}. However, to the best of our knowledge, the CRB expression for coarsely quantized measurements, which is considered our work, is not available in the literature. The derivation given in the following fills this gap.\\
\indent We generalize the CRB expression for binary quantization derived in \cite{css_binaryPR} to the case where a general multi-level quantizer is employed to encode the measurements. Recall that the quantized measurements in the presence of noise is given by
\begin{equation*}
y_i=\mathcal{Q}\left(\left| \boldsymbol a_i^\top \boldsymbol x^*  \right|^2 +\xi_i\right), i=1:m,
\end{equation*}
where $\xi_i$ denotes the noise sample corresponding to measurement index $i$. We assume that the sensing vectors $\left\{\boldsymbol{a}_i\right\}_{i=1}^{m}$ are constant and the randomness in the measurements is only due to additive noise. To derive the CRB for a $k$-level quantizer, where $k \geq 2$, we assume that the noisy quadratic measurement corresponding to $y_i$ falls in the interval $\left(\tau_i^{\text{L}},\tau_i^{\text{R}}\right)$, so that $y_i$ is the encoding symbol for $\left[\tau_i^{\text{L}},\tau_i^{\text{R}}\right)$. Observe that $\tau_i^{\text{L}}\in\left\{\tau_0,\cdots,\tau_{k-1}\right\}$, whereas $\tau_i^{\text{R}}\in\left\{\tau_1,\cdots,\tau_{k}\right\}$, for every $i$. Given this notation, the log-likelihood function of $\left\{y_i\right\}_{i=1}^{m}$ for estimating $\boldsymbol x^*$ is given by 
\begin{eqnarray}
f_{\text{ml}}^{\log}\left(\boldsymbol x^*\right) &=& \log \left[\prod_{i=1}^{m}P\left\{y_i\in \left[\tau_i^{\text{L}},\tau_i^{\text{R}}\right)\right\}\right]\nonumber\\&=&\sum_{i=1}^{m}\log\left[P\left\{\xi_i\in \left[\tau_i^{\text{L}}-u_i^2,\tau_i^{\text{R}}-u_i^2\right)\right\} \right]\nonumber\\ &=&\sum_{i=1}^{m}\log\left(\Phi\left(\tau_i^{\text{R}}-u_i^2\right)-\Phi\left(\tau_i^{\text{L}}-u_i^2\right)\right),\nonumber\\
\label{ml_func_step1_gen}
\end{eqnarray}
where $u_i = \boldsymbol a_i^\top \boldsymbol x^*$ and $\Phi(\cdot)$ denotes the cumulative distribution function (c.d.f.) of the additive noise. The interval parameters $\tau_i^{\text{R}}$ and $\tau_i^{\text{L}}$ are functions of $y_i$. Differentiating both sides of \eqref{ml_func_step1_gen} with respect to $\boldsymbol x^*$, we get that 
\begin{eqnarray}
\nabla f_{\text{ml}}^{\log}\left(\boldsymbol x^*\right)=\sum_{i=1}^{m} 2u_i\frac{ \varphi_i^{'\text{L}}-\varphi_i^{'\text{R}}}{\varphi_i^{\text{R}} - \varphi_i^{\text{L}}}\boldsymbol a_i,
\label{ml_func_step2_gen}
\end{eqnarray}
where $\varphi_i^{\text{R}}=\Phi\left(\tau_i^{\text{R}}-u_i^2\right)$, $\varphi_i^{\text{L}}=\Phi\left(\tau_i^{\text{L}}-u_i^2\right)$, $\varphi_i^{'\text{R}}=\Phi'\left(\tau_i^{\text{R}}-u_i^2\right)$, and $\varphi_i^{'\text{L}}=\Phi'\left(\tau_i^{\text{L}}-u_i^2\right)$. Differentiating both sides of \eqref{ml_func_step2_gen} again with respect to $\boldsymbol x^*$ gives
\begin{eqnarray}
\nabla^2 f_{\text{ml}}^{\log}\left(\boldsymbol x^*\right)&=&\sum_{i=1}^{m} \beta_i\left(\tau_i^{\text{L}},\tau_i^{\text{R}}\right)\boldsymbol a_i \boldsymbol a_i^\top,
\label{ml_func_step3_gen}
\end{eqnarray}
where $\beta_i\left(\tau_i^{\text{L}},\tau_i^{\text{R}}\right)$ is defined as
\begin{eqnarray}
\beta_i\left(\tau_i^{\text{L}},\tau_i^{\text{R}}\right)&=&2\frac{ \varphi_i^{'\text{L}}-\varphi_i^{'\text{R}}}{\varphi_i^{\text{R}} - \varphi_i^{\text{L}}}\nonumber\\&+&4u_i^2 \frac{ \left(\varphi_i^{\text{R}} - \varphi_i^{\text{L}}\right)(\varphi_i^{''\text{R}}-\varphi_i^{''\text{L}})-(\varphi_i^{'\text{L}} - \varphi_i^{'\text{R}})^2}{\left(\varphi_i^{\text{R}} - \varphi_i^{\text{L}}\right)^2}.\nonumber
\label{ml_func_step4_gen}
\end{eqnarray}
Finally, to compute the Fisher matrix, one has to evaluate the expectation of both sides of \eqref{ml_func_step3_gen} over the distribution of $y_i$. This in turn requires the computation of 
\begin{eqnarray}
\bar{\beta}_i&=&\mathbb{E}_{\bar{\boldsymbol y}}\left[\beta_i\left(\tau_i^{\text{L}},\tau_i^{\text{R}}\right)\right]\nonumber\\&=&\sum_{j=1}^{k}\beta_i\left(\tau_{j-1},\tau_{j}\right)P\left\{\tau_{j-1}\leq u_i^2+\xi_i\leq \tau_{j}\right\}\nonumber\\
&=&\sum_{j=1}^{k}\beta_i\left(\tau_{j-1},\tau_{j}\right)\left(\Phi\left(\tau_j-u_i^2\right)-\Phi\left(\tau_{j-1}-u_i^2\right)\right).\nonumber\\
\label{beta_mean}
\end{eqnarray}
Defining $\varphi_{i}\left(\tau\right)=\Phi\left(\tau-u_i^2\right)$, $\varphi'_{i}\left(\tau\right)=\Phi'\left(\tau-u_i^2\right)$, and $\varphi''_{i}\left(\tau\right)=\Phi''\left(\tau-u_i^2\right)$, for $\tau\in\mathbb{R}$, we write $\bar{\beta}_i$ in \eqref{beta_mean} as
\begin{eqnarray}
\bar{\beta}_i&=& \sum_{j=1}^{k} 2\left[\varphi'_{i}\left(\tau_{j-1}\right)-\varphi'_{i}\left(\tau_{j}\right)\right]\nonumber\\&+&4u_i^2 \left[\varphi''_{i}\left(\tau_{j}\right)-\varphi''_{i}\left(\tau_{j-1}\right)\right]\nonumber\\&-&4u_i^2 \frac{\left[\varphi'_{i}\left(\tau_{j}\right)-\varphi'_{i}\left(\varphi_{j-1}\right)\right]^2}{\varphi_{i}\left(\tau_{j}\right)-\varphi_{i}\left(\tau_{j-1}\right)}.
\label{beta_mean_final1}
\end{eqnarray}
For the binary quantization case, that is, $k=2$, we have $\tau_0=-\infty$, $\tau_1=\tau$, and $\tau_2=+\infty$. Making the substitutions 
\begin{equation*}
\varphi'_{i}\left(-\infty\right)=\varphi'_{i}\left(+\infty\right)=\varphi''_{i}\left(-\infty\right)=\varphi''_{i}\left(+\infty\right)=0;
\end{equation*}
\begin{equation*}
\varphi_{i}\left(-\infty\right)=0; \text{\,\,and\,\,}\varphi_{i}\left(+\infty\right)=1
\end{equation*}
in \eqref{beta_mean_final1} leads to the same expression as developed in \cite{css_binaryPR}, where $\varphi_i=\varphi_{i}\left(\tau\right)$ and $\varphi'_i=\varphi'_{i}\left(\tau\right)$. Further simplifications in the expression for $\bar{\beta}_i$ in \eqref{beta_mean_final1} can be achieved by using the fact that $\sum_{j=1}^{k}\left(t_j-t_{j-1}\right)=t_k-t_0$, for any sequence of numbers $t_1,t_2,\cdots, t_k$; thereby leading to
\begin{eqnarray}
\bar{\beta}_i&=& 2\left(\varphi'_{i}\left(\tau_{0}\right)-\varphi'_{i}\left(\tau_{k}\right)\right)+4u_i^2\left(\varphi''_{i}\left(\tau_{k}\right)-\varphi''_{i}\left(\tau_{0}\right)\right)\nonumber\\&-&4u_i^2\sum_{j=1}^{k} \frac{\left[\varphi'_{i}\left(\tau_{j}\right)-\varphi'_{i}\left(\tau_{j-1}\right)\right]^2}{\varphi_{i}\left(\tau_{j}\right)-\varphi_{i}\left(\tau_{j-1}\right)}.
\label{beta_mean_final2}
\end{eqnarray}
Substituting $\tau_0=-\infty$ and $\tau_k=+\infty$ in \eqref{beta_mean_final2}, we arrive at 
\begin{eqnarray}
\bar{\beta}_i&=& -4u_i^2\sum_{j=1}^{k} \frac{\left[\varphi'_{i}\left(\tau_{j}\right)-\varphi'_{i}\left(\tau_{j-1}\right)\right]^2}{\varphi_{i}\left(\tau_{j}\right)-\varphi_{i}\left(\tau_{j-1}\right)}.
\label{beta_mean_final3}
\end{eqnarray}
The expression for $\bar{\beta}_i$ in \eqref{beta_mean_final3} is used to evaluate the Fisher information matrix, and therefore the CRB.  
\subsection{Proof of the Descent Property of the PGD Algorithm}
\label{pgd_descent_proof_qpr}
In this section, we show that the PGD algorithm for QPR leads to a monotonically non-increasing objective function value. Consider the update rule of a projected gradient-descent (PGD) algorithm for QPR:
\begin{equation}
{\boldsymbol X}^{t+1}=\mathcal{P}_{\text{rank}-1}\left({\boldsymbol X}^{t}-\eta^t  \nabla \left. F\left( \boldsymbol X\right)\right|_{\boldsymbol X=\boldsymbol X^t}\right),
\label{update1}
\end{equation}
which can be rewritten as
\begin{equation}
{\boldsymbol X}^{t+1}=\arg \underset{\boldsymbol X \in \mathcal{R}_1}{\min}\text{\,\,}\frac{1}{2\eta^t}\left\|\boldsymbol X-\left({\boldsymbol X}^{t}-\eta^t  \nabla \left. F\left( \boldsymbol X\right)\right|_{\boldsymbol X=\boldsymbol X^t}\right)\right\|_{\textsc{F}}^2,
\label{update2}
\end{equation}
where $\|\cdot\|_\textsc{F}$ denotes the Frobenius norm and $\mathcal{R}_1$ is the set of all symmetric rank-1 matrices. Rearranging terms, the update turns out to be equivalent to
\begin{equation}
{\boldsymbol X}^{t+1}=\arg \underset{\boldsymbol X \in \mathcal{R}_1}{\min}\text{\,}P\left(\boldsymbol X,\boldsymbol X^t\right),
\label{update3}
\end{equation}
where $P\left(\boldsymbol X,\boldsymbol X^t\right)$ is defined as 
\begin{eqnarray*}
	P\left(\boldsymbol X,\boldsymbol X^t\right) &=& F\left( \boldsymbol X^t\right)+ \text{Tr}\left(\nabla  F\left( \boldsymbol X^t\right)^\top \left( \boldsymbol X -\boldsymbol X^t \right)\right)\\&+&\frac{1}{2\eta^t}\left\|\boldsymbol X -\boldsymbol X^t \right\|_{\textsc{F}}^2. 
\end{eqnarray*}
Suppose for now that the gradient of $F\left( \boldsymbol X\right)$ is Lipschitz continuous (which we shall actually establish in the following), that is, there exists a constant $L>0$ such that
\begin{equation*}
\left\| \nabla F\left( \boldsymbol X\right)-\nabla F\left( \boldsymbol Y\right) \right\|_{\textsc{F}} \leq L \left\| \boldsymbol X-\boldsymbol Y\right\|_{\textsc{F}},
\label{lipschitz}
\end{equation*}
for every pair of symmetric matrices $\left( \boldsymbol X,\boldsymbol Y\right)$. Then, for $\eta^t<\frac{1}{L}$, we have $$F\left( \boldsymbol X\right)\leq P\left(\boldsymbol X,\boldsymbol X^t\right)$$ for any symmetric $\boldsymbol X$, and, in particular, $$F\left( \boldsymbol X^{t+1}\right)\leq P\left(\boldsymbol X^{t+1},\boldsymbol X^t\right).$$ Since $\boldsymbol X^{t}$ and $\boldsymbol X^{t+1}$ belong to $\mathcal{R}_1$, we have that 
\begin{equation*}
F\left( \boldsymbol X^{t+1}\right)\leq P\left(\boldsymbol X^{t+1},\boldsymbol X^t\right) \stackrel{\text{(i)}}{\leq} P\left(\boldsymbol X^{t},\boldsymbol X^t\right)=F\left( \boldsymbol X^t\right),
\end{equation*}
where the inequality (i) is a consequence of \eqref{update3}. Therefore, the PGD algorithm does not increase the objective provided that $F\left( \boldsymbol X\right)$ has a Lipschitz-continuous gradient. That $\nabla F\left( \boldsymbol X\right)$ is indeed Lipschitz continuous is established in the following.\\
\noindent \textit{Lipschitz Continuity of $\nabla F\left( \boldsymbol X\right)$}:
Recall from \eqref{F_def_eq} that the optimization objective in QPR is given by
\begin{equation*}
F\left(\boldsymbol X \right)=\sum_{j=1}^{k}\sum_{i=1}^{m_j}f\left(\tau_j-\text{Tr}(\boldsymbol A_{i}^j \boldsymbol X)\right)+f\left(\text{Tr}(\boldsymbol A_{i}^j \boldsymbol X)-\tau_{j-1}\right).
\end{equation*}
For convenience, we denote 
\begin{equation*}
u_{ij}^1=\tau_j-\text{Tr}\left(\boldsymbol A_{i}^j \boldsymbol X\right)\text{\,\,and\,\,}  u_{ij}^2 = \text{Tr}\left(\boldsymbol A_{i}^j \boldsymbol X\right)-\tau_{j-1}.
\end{equation*}
The $(j_1,j_2)^{\text{th}}$ entry of the gradient $\boldsymbol G=\nabla F\left( \boldsymbol X\right)$ is given by
\begin{equation}
\boldsymbol G_{j_1,j_2}=\sum_{j=1}^{k}\sum_{i=1}^{m_j}f' \left(u_{ij}^2\right) a_{i j_1}^j a_{i j_2}^j - f' \left(u_{ij}^1\right) a_{i j_1}^j a_{i j_2}^j,
\label{def_G}
\end{equation}
where $f'$ denotes the derivative of $f$, and $a_{i \ell}^j$ is the $\ell^{\text{th}}$ entry of the vector $\boldsymbol a_i^j$. Differentiating \eqref{def_G} further with respect to ${\boldsymbol X}_{k_1,k_2}$, we get the Hessian (which is a tensor):
\begin{equation*}
\mathbb{H}_{j_1,j_2,k_1,k_2}=\sum_{j=1}^{k}\sum_{i=1}^{m_j}\left[f''\left(u_{ij}^1\right)+f''\left(u_{ij}^2\right)\right] a_{i j_1}^ja_{i j_2}^j a_{i k_1}^j a_{i k_2}^j,
\end{equation*}
where the function $f''$ denotes the sub-differential of $f'$ and since $f(u)=\frac{1}{2}u^2\mathbbm{1}_{(u\leq0)}$, it follows that $0\leq f''(u)\leq 1$ for any $u$.\\
For a symmetric positive-definite matrix $\boldsymbol U\in \mathbb{R}^{n\times n}$, we have
\begin{eqnarray*}
&&	\sum_{j_1,j_2=1}^{n}\sum_{k_1,k_2=1}^{n}{\boldsymbol U}_{j_1,j_2}\mathbb{H}_{j_1,j_2,k_1,k_2}{\boldsymbol U}_{k_1,k_2}\\&=&\sum_{j=1}^{k}\sum_{i=1}^{m_j}\left[f''\left(u_{ij}^1\right)+f''\left(u_{ij}^2\right)\right]((\boldsymbol a_i^j)^\top \boldsymbol U \boldsymbol a_i^j )^2\\
	&\leq& 2\sum_{j=1}^{k}\sum_{i=1}^{m_j}((\boldsymbol a_i^j)^\top \boldsymbol U \boldsymbol a_i^j)^2.
	%\\&\leq&2 \lambda_{\max}^2\left(\boldsymbol U\right)\sum_{j=1}^{k}\sum_{i=1}^{m_j}\left\| \boldsymbol a_i^j \right\|_2^4\\&=&2 \lambda_{\max}^2\left(\boldsymbol U\right)\sum_{i=1}^{m}\left\| \boldsymbol a_i^j \right\|_2^4,
\end{eqnarray*}
Now, for any $\boldsymbol a \in \mathbb{R}^{n}$, we have
\begin{equation*}
\left(\boldsymbol a^\top \boldsymbol U \boldsymbol a\right)^2\leq \lambda_{\max}^2\left(\boldsymbol U\right) \left\| \boldsymbol a \right\|_2^4,  
\end{equation*}
where $\lambda_{\max}\left(\boldsymbol U\right)$ is the spectral norm or the largest eigenvalue of $\boldsymbol U$. Denoting $C_0=2\displaystyle\sum_{i=1}^{m}\left\| \boldsymbol a_i \right\|_2^4$ and using the property that the spectral norm is dominated by the Frobenius norm, we have $\lambda_{\max}^2\left(\boldsymbol U\right)\leq \left\| \boldsymbol U\right\|_{\textsc{F}}^2$, and therefore
\begin{eqnarray*}
	&&\sum_{j_1,j_2=1}^{n}\sum_{k_1,k_2=1}^{n}{\boldsymbol U}_{j_1,j_2}\mathbb{H}_{j_1,j_2,k_1,k_2}{\boldsymbol U}_{k_1,k_2}\\&\leq& 2\left\| \boldsymbol U\right\|_{\textsc{F}}^2\sum_{j=1}^{k}\sum_{i=1}^{m_j}\| \boldsymbol a_i^j\|_2^4= C_0 \left\| \boldsymbol U\right\|_{\textsc{F}}^2,
\end{eqnarray*}
thereby establishing that $\nabla F\left( \boldsymbol X\right)$ is Lipschitz-continuous.\hfill $\blacksquare$\\
Since the gradient $\nabla F\left( \boldsymbol X\right)$ of the optimization objective is Lipschitz-continuous, it is guaranteed that the PGD scheme for QPR does not increase the cost as the iterations progress. This descent property does not hold any longer when a momentum factor is incorporated in the PGD scheme, due to non-convexity of the rank-1 constraint. However, we have shown experimentally that the inclusion of a momentum term indeed improve the reconstruction SNR at a rate faster than the PGD scheme (cf. Section~\ref{QPR_vs_QPRA_sec} in the main manuscript), although we did not establish it analytically. This observation is also consistent with the remarks made my Cand\`es et al. in the context of PhaseLift (cf. Section 4.1 of \cite{PL2_candes}).

\ifCLASSOPTIONcaptionsoff
  \newpage
\fi


\begin{thebibliography}{99}

\bibitem{crystallography_chspr}
R. P. Millane, ``Phase retrieval in crystallography and optics," \emph{J. Opt. Soc. Amer. A}, vol. 7, no. 3, pp. 394--411, Mar. 1990.

\bibitem{holography_chspr}
 A. Szoke, ``Holographic microscopy with a complicated reference," \emph{J. Imag. Sci. Technol.}, vol. 41, pp. 332--341, 1997.
 
\bibitem{microscopy_chspr} 
A. J. J. Drenth, A. Huiser, and H. Ferwerda, ``The problem of phase retrieval in light and electron microscopy of strong objects," \emph{Optica Acta}, vol. 22, pp. 615--628, 1975.

\bibitem{fienup_main1}
J. R. Fienup, ``Phase retrieval algorithms: A comparison," \emph{Appl. Opt.}, vol. 21, pp. 2758--2769, 1982.

\bibitem{fienup_main2}
J. R. Fienup, ``Phase retrieval algorithms: A personal tour [invited]," \emph{Appl. Opt.}, vol. 52, pp. 45--56, 2013.

\bibitem{gs_pr}
R. W. Gerchberg and W. O. Saxton, ``A practical algorithm for the determination of phase from image and diffraction plane pictures," \emph{Optik}, vol. 35, pp. 237--246, 1972.

\bibitem{bauschke_pr}
H. H. Bauschke, P. L. Combettes, and D. Luke, ``Phase retrieval, error reduction algorithm, and Fienup variants: A view from convex optimization," \emph{J. Opt. Soc. Amer. A}, vol. 19, pp. 1334--1345, 2002.

\bibitem{quatieri_pr}
T. F. Quatieri and A. V. Oppenheim, ``Iterative techniques for minimum-phase signal reconstruction from phase or magnitude," \emph{IEEE Trans. Acoust., Speech, Signal Process.}, vol. 29, pp. 1187--1193, 1981.

\bibitem{yegna_pr}
B. Yegnanarayana and A. Dhayalan, ``Noniterative techniques for minimum phase signal reconstruction from phase or magnitude," in \emph{Proc. IEEE Intl. Conf. Acoust., Speech, and Signal Process.}, vol. 8, pp. 639--642, Apr. 1983.

\bibitem{css_holography}
C. S. Seelamantula, N. Pavillon, C. Depeursinge, and M. Unser, ``Exact complex-wave reconstruction in digital holography," \emph{J. Opt. Soc. Amer. A}, vol. 28, no. 6, pp. 983--992, Jun. 2011.

\bibitem{param2D_icip}
B. A. Shenoy, S. Mukherjee, and C. S. Seelamantula, ``Phase retrieval for a class of 2-D signals characterized by first-order difference equations," in \emph{Proc. IEEE Intl. Conf. on Image Process.}, pp. 325--329, 2013.

\bibitem{param2D_tip}
B. A. Shenoy and C. S. Seelamantula, ``Exact phase retrieval for a class of 2-D parametric signals," \emph{IEEE Trans. Signal Process.}, vol. 63, no. 1, pp. 90--103, 2015.

\bibitem{pr_shift_invar}
B. A. Shenoy, S. Mulleti, and C. S. Seelamantula, ``Exact phase retrieval in principal shift-invariant spaces," \emph{IEEE Trans. Signal Process.}, vol. 64, no. 2, pp. 406--416, 2016.

\bibitem{css_2d_hilbert}
B. A. Shenoy, S. Mulleti, and C. S. Seelamantula, ``On 2-D Hilbert integral equations, generalized minimum-phase signals, and phase retrieval," To appear in \emph{IEEE Trans. Signal Process.}

\bibitem{moravec_cpr}
M. L. Moravec, J. K. Romberg, and R. G. Baraniuk, ``Compressive phase retrieval," in \emph{Proc. Wavelets XII SPIE Int. Symp. Opt. Sci. Technol.}, Aug. 2007.

\bibitem{yu_vettreli_ssf}
Y. M. Yu and M. Vetterli, ``Sparse spectral factorization: Unicity and reconstruction algorithms," in \emph{Proc. IEEE Intl. Conf. Acoust. Speech, Signal Process.}, pp. 5976--5979, 2011.

\bibitem{gespar}
Y. Shechtman, A. Beck, and Y. C. Eldar, ``GESPAR: Efficient phase retrieval of sparse signals," \emph{IEEE Trans. Signal Process.}, vol. 62, no. 4, pp. 928--938, Feb. 2014. 

\bibitem{netrapalli_altminpr}
P. Netrapalli, P. Jain, and S. Sanghavi, ``Phase retrieval using alternating minimization," \emph{IEEE Trans. Signal Process.}, vol. 63, no. 18, pp. 4814--4826, Sep. 2015.

\bibitem{low_rank_PR_vaswani}
N. Vaswani, S. Nayer, and Y. C. Eldar, ``Low rank phase retrieval," \emph{arXiv:1608.04141v1}, Aug. 2016.

\bibitem{dolphin}
A. M. Tillmann, Y. C. Eldar, and J. Mairal, ``DOLPHIn--Dictionary learning for phase retrieval," \emph{arXiv:1602.02263v1}, Feb. 2016.

\bibitem{tfocs}
S. Becker, E. J. Cand\`es, and M. Grant, ``Templates for convex cone problems with applications to sparse signal recovery," \emph{Technical Report}, Department of Statistics, Stanford University, 2010.

\bibitem{schniter_gamp}
P. Schniter and S. Rangan, ``Compressive phase retrieval via generalized approximate massage passing," \emph{IEEE Trans. Signal Process.}, vol. 63, no. 4, pp. 1043--1055, Feb. 2015.

\bibitem{bspr_peng}
W. Peng and H. Wang, ``Binary sparse phase retrieval via simulated annealing," \emph{Math. Probl. in Engg.}, Article ID 8257612, May 2016.

\bibitem{palomar_mm}
T. Qiu and D. P. Palomar, ``Undersampled sparse phase retrieval via majorization-minimization," \emph{IEEE Trans. Signal Process.}, vol. 65, no. 22, pp. 5957--5969, Nov. 2017.

\bibitem{fogel_pr_imaging}
F. Fogel, I. Waldspurger, and A. d'Aspremont, ``Phase retrieval for imaging problems," \emph{arXiv:1304.7735v2}, Apr. 2014.

\bibitem{PL1_candes}
E. J. Cand\`es, T. Strohmer, and V. Voroninski, ``PhaseLift: Exact and stable signal recovery from magnitude measurements via convex programming," \emph{Comm. on Pure and Appl. Math.}, vol. 66, issue 8, pp. 1241--1274, Aug. 2013.

\bibitem{PL2_candes}
E. J. Cand\`es, Y. C. Eldar, T. Strohmer, and V. Voroninski, ``Phase retrieval via matrix completion," \emph{SIAM J. Imag. Sci.}, vol. 6, issue 1, pp. 199--224, Feb. 2013.  

\bibitem{cpr_pl}
H. Ohlsson, A. Y. Yang, R. Dong, and S. S. Sastry, ``Compressive phase retrieval from squared output measurements via semidefinite programming," \emph{arXiv:1111.6323v3}, Mar. 2012.

\bibitem{qcs_eldar}
Y. Shechtman, Y. C. Eldar, A. Szameit, and M. Segev, ``Sparsity based sub-wavelength imaging with partially incoherent light via quadratic compressed sensing," \emph{Opt. Exp.}, vol. 19, no. 16, pp. 14807--14822, 2011.

\bibitem{power_it}
G. H. Golub and C. H. Van Loan, \emph{Matrix Computations}. Fourth edition, JHU Press, Oct. 1996.

\bibitem{candes_wf_it}
E. J. Cand\`es, X. Li, and M. Soltanolkotabi, ``Phase retrieval via Wirtinger flow: Theory and algorithms," \emph{IEEE Trans. Info. Theory}, vol. 61, no. 4, pp. 1985--2007, Apr. 2015.

\bibitem{candes_twf}
Y. Chen and E. J. Cand\`es, ``Solving random quadratic systems of equations is nearly as easy as solving linear systems," in \emph{Proc. Advances in Neural Info. Process. Systems 28}, 2015. 

\bibitem{phasecut}
I. Waldspurger, A. dÕAspremont, and S. Mallat, ``Phase recovery, max-cut and complex semidefinite
programming," \emph{Math. Program.}, vol. 149, no. 1, pp. 47--81, 2015.
\bibitem{zymnis_qcs}
A. Zymnis, S. Boyd, and E. J. Cand\`es, ``Compressed sensing with quantized measurements," \emph{IEEE Signal Process. Lett.}, vol. 17, no. 2, pp. 149--152, Feb. 2010. 

\bibitem{laska1}
J. Laska, P. Boufounos, M. Davenport, and R. Baraniuk, ``Democracy in action: Quantization, saturation, and compressive sensing," \emph{Applied and Computational Harmonic Analysis}, vol. 31(3), pp. 429--443, Nov. 2011.

\bibitem{qcs_bit_precision}
E. Ardestanizadeh, M. Cheraghchi, and A. Shokrollahi, ``Bit precision analysis for compressed sensing," in \emph{Proc. IEEE Intl. Symp. on Info. Theory}, pp. 1--5, 2009.

\bibitem{binary_imaging1}
A Bourquard, F. Aguet, and M. Unser, ``Optical imaging using binary sensors," \emph{Opt. Exp.}, vol. 18, no. 5, Mar. 2010.

\bibitem{binary_imaging2}
A Bourquard and M. Unser, ``Binary compressed imaging," \emph{IEEE Trans. Image Process.}, vol. 22, no. 3, pp. 1042--1055, Mar. 2013.

\bibitem{baraniuk1}
P. T. Boufounos and R. G. Baraniuk, ``1-bit compressive sensing," in \emph{Proc. Conf. Info. Science and Systems}, Princeton, NJ, Mar. 2008.

\bibitem{plan1}
Y. Plan and R. Vershynin, ``One-bit compressed sensing by linear programming," \emph{arXiv:1109.4299v5}, Mar. 2012.

\bibitem{msp}
P. T. Boufounos, ``Greedy sparse signal reconstruction from sign measurements," in \emph{Proc. Asilomar Conf.  Signals, Systems, and Computation}, Asilomar, CA, Nov. 2009.


%\bibitem {luke_pr_diff_image}
%D.~R.~Luke, ``Relaxed averaged alternating reflections for diffraction imaging," {\it Inverse Problems}, vol.~21, pp.~37--50, 2005.

%\bibitem{oppenheim_dsp_book}
%A. V. Oppenheim, R. W. Schaffer, and J. R. Buck, \emph{Discrete-Time Signal Processing}. Prentice Hall, 1998. 

\bibitem{css_oct_2008} 
C.~S.~Seelamantula, M.~L.~Villiger, R.~A.~Leitgeb, and M.~Unser, ``Exact and efficient signal reconstruction in frequency-domain optical coherence tomography," \emph{J. Opt. Soc. Am. (A),} vol.~25, no.~7, pp.~1762--1771, Jul.~2008.

%\bibitem{ozcan_oct_2006}
%A. Ozcan, M. J. F. Digonnet, and G. S. Kino, ``Minimum-phase-function-based processing in frequency-domain optical-coherence tomography systems," \emph{J. Opt. Soc. Am. A}, vol. 23, pp. 1669--1677, 2006.

%\bibitem{css_iisc_journal}
%C. S. Seelamantula and T. Lasser, ``Hilbert transform relations in frequency-domain optical-coherence tomographic imaging," Invited article, \emph{Journal of the Indian Institute of Science, Special issue on Imaging and Microscopy}, vol. 93, no. 1, pp. 139--148, Mar. 2013.
%
\bibitem{css_sparsePR}
S. Mukherjee and C. S. Seelamantula, ``Fienup algorithm with sparsity constraints: Application to frequency-domain optical-coherence tomography," \emph{IEEE Trans. Signal Process.}, vol. 62, no. 18, pp. 4659--4672, Sep.15, 2014.

\bibitem{css_binaryPR}
S. Mukherjee and C. S. Seelamantula, ``Phase retrieval from binary measurements," \emph{IEEE Signal Process. Lett.}, vol. 25, no. 3, pp. 348--352, Mar. 2018.

\bibitem{vargamma_ref}
D. B. Madan and E. Seneta, ``The variance gamma (V.G.) model for share market returns," \emph{J. Business}, vol. 63, no. 4, pp. 511--524, Oct. 1990.

\bibitem{jayant_noll}
N. S. Jayant and P. Noll, \textit{Digital Coding of Waveforms: Principles and Applications to Speech and Video}, Prentice-Hall, USA, 1984.

\bibitem{bstj_paper}
B. Smith, ``Instantaneous companding of quantized signals," \emph{Bell System Tech. J.}, vol. 36, pp. 653--709, May 1957.


\bibitem{svt_candes}
J.-F. Cai, E. J. Cand\`es, and Z. Shen, ``A singular value thresholding algorithm for matrix completion," \emph{SIAM J. Optimization}, vol. 20, no. 4, pp. 1956--1982, Mar. 2010.

\bibitem{nesterov_momentum1}
Y. E. Nesterov, \emph{Introductory Lectures on Convex Optimization: A Basic Course}, Kluwer Academic Publishers, London, 2004.

\bibitem{geng_yang}
J. Geng, X. Yang, X. Wang, and L. Wang, ``An accelerated iterative hard-thresholding method for matrix completion," \emph{Intl. J. Signal Process., Image Process., and Pattern Recognition}, vol. 8, no. 7, pp. 141--150, 2015.

\bibitem{grant_cvx} 
M. Grant and S. Boyd, ``CVX: Matlab software for disciplined convex programming," version 2.0 beta, Sep. 2013.

\bibitem{ssim_paper_ref}
Z. Wang, A. C. Bovik, H. R. Sheikh, and E. P. Simoncelli, ``Image quality assessment: From error visibility to structural similarity," \emph{IEEE Trans. Image Process.}, vol. 13, no. 4, pp. 600--612, Apr. 2004.	

\bibitem{balan1_qpr}
R. Balan, ``Reconstruction of signals from magnitudes of redundant representations: The complex case," \emph{Found. Comp. Math.}, pp. 1--45, 2013.

\bibitem{balan2_qpr}
R. Balan, ``The Fisher information matrix and the CRLB in a non-AWGN model for the phase retrieval problem," in \emph{Proc. Intl. Conf. Sampl. Theory and Applications}, pp. 178--182, 2015. 

\bibitem{pr_crb_fpp_qpr}
C. Qian, N. D. Sidiropoulos, K. Huang, L. Huang, and H. C. So, ``Phase retrieval using feasible point pursuit: Algorithms and Cram\'er-Rao bound," \emph{IEEE Trans. Signal Process.}, vol. 64, no. 20, pp. 5282--5296, Oct. 2016.

\bibitem{bandeira_qpr}
A. S. Bandeira, J. Cahill, D. G. Mixon, and A. A. Nelson, ``Saving phase: Injectivity and stability for phase retrieval," \emph{Appl. and Comp. Harmonic Anal.}, vol. 37, no. 1, pp. 106--125, 2014.

\bibitem{cederquist_qpr}
J. N. Cederquist and C. C. Wackerman, ``Phase-retrieval error: A lower bound," \emph{J. Opt. Soc. Amer. A}, vol. 4, no. 9, pp. 1788--1792, 1987.

\end{thebibliography}
\end{document}